\documentclass[11pt]{article}

\usepackage[margin=1in]{geometry}
\usepackage{amsmath,amssymb}
\usepackage{graphicx}
\usepackage{subcaption}
\usepackage{multirow}
\usepackage[section]{placeins} % \FloatBarrier at each \section: floats cannot drift to the document end
\usepackage{tikz}
\usetikzlibrary{arrows}
\graphicspath{{./}{figures/}}
\usepackage{xurl}
\usepackage[numbers,super,sort&compress]{natbib} % Nature-style numbered citations
\usepackage{authblk}

\usepackage[colorlinks=true,linkcolor=blue,citecolor=blue,urlcolor=blue]{hyperref}
\newcommand{\code}[1]{\texttt{#1}}
\newcommand{\mcode}[1]{\text{\texttt{#1}}}
\newcommand{\CamelyonCromaMin}{$-0.44$}
\newcommand{\CamelyonCromaMax}{$0.20$}

\newcommand{\CamelyonCromaVsRiRho}{$0.95$}
\newcommand{\CamelyonCromaVsMariRho}{$0.94$}
\newcommand{\CamelyonRiVsMariRho}{$0.99$}
\newcommand{\CamelyonMariRiMaxAbsDelta}{$0.07$}
\newcommand{\CamelyonRankedNModels}{20}

\newcommand{\CamelyonProbeRiRho}{$-0.95$}
\newcommand{\CamelyonProbeMariRho}{$-0.94$}
\newcommand{\CamelyonProbeCromaRho}{$-0.94$}

\newcommand{\CamelyonBioProbeCromaRho}{$0.56$}
\newcommand{\CamelyonProbeNModels}{20}

\newcommand{\TcgaFourByFourCromaSpan}{$[-0.01, 0.40]$}

\newcommand{\TcgaFourByFourCromaMax}{$0.40$}

\newcommand{\TcgaFourByFourCromaConfounderDominantWord}{one}

\newcommand{\TcgaFourByFourCromaVsRiRho}{$0.95$}
\newcommand{\TcgaFourByFourCromaVsMariRho}{$0.95$}
\newcommand{\TcgaFourByFourRiVsMariRho}{$0.99$}
\newcommand{\TcgaFourByFourMariRiMaxAbsDelta}{$0.04$}
\newcommand{\TcgaFourByFourRankedNModels}{20}

\newcommand{\ProvenanceModel}{Midnight-12k}

\newcommand{\ProvenanceRunnerUpModel}{GenBio-PathFM}
\newcommand{\ProvenanceRunnerUpCroma}{$0.16$}

\newcommand{\TolkachCromaSpan}{$[0.11, 0.58]$}

\newcommand{\TolkachCromaMax}{$0.58$}

\newcommand{\TolkachCromaConfounderDominantWord}{none}

\newcommand{\TolkachCromaVsRiRho}{$0.93$}
\newcommand{\TolkachCromaVsMariRho}{$0.92$}
\newcommand{\TolkachRiVsMariRho}{$0.98$}
\newcommand{\TolkachMariRiMaxAbsDelta}{$0.04$}
\newcommand{\TolkachRankedNModels}{20}

\newcommand{\PandaCromaSpan}{$[-0.41, 0.26]$}

\newcommand{\PandaRankedNModels}{4}

\newcommand{\PandaBestCroma}{$0.26$}

\newcommand{\PandaIsupCromaSpan}{$[-0.47, -0.09]$}

\newcommand{\PandaIsupBestCroma}{$-0.09$}
\newcommand{\PandaIsupBestBioBacc}{$0.934$}

\newcommand{\CrossCohortRhoMean}{$0.90$}
\newcommand{\CrossCohortRhoRange}{$[0.88, 0.92]$}

\newcommand{\TileRankedNModels}{20}
\newcommand{\SlideNModels}{4}
\newcommand{\ProvenanceTolkachBoost}{$2.5\times$}

\newcommand{\ApdNModels}{20}

\newcommand{\NipdIdCromaCamelyon}{0.88}
\newcommand{\NipdIdCromaTcga}{0.93}
\newcommand{\NipdIdCromaTolkach}{0.93}
\newcommand{\NipdIdCromaPcabiop}{0.80}

\newcommand{\NipdIdRiCamelyon}{0.86}
\newcommand{\NipdIdRiTcga}{0.87}
\newcommand{\NipdIdRiTolkach}{0.92}

\newcommand{\NipdIdMariCamelyon}{0.85}
\newcommand{\NipdIdMariTcga}{0.89}
\newcommand{\NipdIdMariTolkach}{0.91}

\newcommand{\NipdOodCromaCamelyon}{0.64}
\newcommand{\NipdOodCromaTcga}{0.85}
\newcommand{\NipdOodCromaTolkach}{0.77}
\newcommand{\NipdOodCromaPcabiop}{0.60}

\newcommand{\NipdOodRiCamelyon}{0.65}
\newcommand{\NipdOodRiTcga}{0.84}
\newcommand{\NipdOodRiTolkach}{0.84}

\newcommand{\NipdOodMariCamelyon}{0.61}
\newcommand{\NipdOodMariTcga}{0.82}
\newcommand{\NipdOodMariTolkach}{0.78}

\newcommand{\ApdIdCromaCamelyon}{0.88}
\newcommand{\ApdIdCromaTcga}{0.93}
\newcommand{\ApdIdCromaTolkach}{0.93}

\newcommand{\ApdOodCromaCamelyon}{0.71}
\newcommand{\ApdOodCromaTcga}{0.87}
\newcommand{\ApdOodCromaTolkach}{0.87}

\newcommand{\PcabiopMoozyNipdId}{-6.5\%}
\newcommand{\PcabiopMoozyNipdOod}{4.8\%}
\newcommand{\PcabiopPrismNipdId}{-7.8\%}
\newcommand{\PcabiopPrismNipdOod}{-8.3\%}
\newcommand{\PcabiopProvGigaPathNipdId}{-24.5\%}
\newcommand{\PcabiopProvGigaPathNipdOod}{-10.7\%}
\newcommand{\PcabiopTitanNipdId}{-18.9\%}
\newcommand{\PcabiopTitanNipdOod}{-12.6\%}
\newcommand{\PcabiopProvGigaPathOodBaseline}{0.548}
\newcommand{\PcabiopProvGigaPathApdOod}{-1.2\%}
\title{\textbf{A distributional robustness margin for pathology foundation models}}

\author[1]{Clément Grisi\footnote{corresponding author: clement.grisi@radboudumc.nl}}
\author[1]{Jeroen van der Laak}
\author[1]{Geert Litjens}
\affil[1]{Department of Pathology, Radboud University Medical Center, Nijmegen, The Netherlands}
\date{}

\begin{document}
\maketitle

\begin{abstract}
Pathology foundation models encode non-biological variation introduced by tissue preparation, staining and scanning, enabling shortcut learning that undermines generalisation across institutions. The Robustness Index (\code{RI}) was proposed to assess whether local representation geometry is dominated by biological or non-biological variation. However, its construction suffers from structural limitations that make cross-model comparison unreliable, calling for a more principled metric. We introduce the Cross-confounder Robustness Margin (\code{CRoMa}), a signed, per-sample margin that measures whether samples sharing the same biology but different confounder lie closer than samples sharing the same confounder but different biology. It is defined for every sample, allowing models to be compared on the same cohort and robustness to be analysed as a distribution rather than reduced to a single pooled score. We evaluated \code{CRoMa} across \TileRankedNModels{} tile-level encoders on three benchmarks. Rankings by median \code{CRoMa} were highly consistent across benchmarks (Spearman $\rho \approx$ \CrossCohortRhoMean{}), yet every encoder retained confounder-dominated samples, whose prevalence and severity varied markedly. Similar patterns emerged for four slide-level encoders evaluated on a separate benchmark, extending the analysis beyond tile-level representations. Higher median \code{CRoMa} was associated with smaller shortcut-induced performance losses in downstream linear probes, supporting its use as a representation-level indicator of shortcut susceptibility.
\end{abstract}

% Nature Communications order: Introduction, Results, Discussion, Methods
\section{Introduction}

Advances in self-supervised learning have ushered in the era of foundation models. By pretraining on large collections of unlabelled data, these models shift much of the learning burden to the representation-learning stage, producing general-purpose embeddings that can be adapted to downstream tasks with substantially fewer labels \cite{SSLsurvey2021,caron2021,bommasani2022,oquab2024}. This is especially compelling in medical domains, where expert annotation is costly and disease prevalence is often long-tailed, leaving many clinically important tasks without enough labelled data for supervised training. In computational pathology, this promise is beginning to materialise: models built on foundation-model representations have started to match or surpass the narrower supervised systems that preceded them \cite{Chen2024-dy,Xu2024-gj,Vorontsov2024-te,campanella2025clinical,Neidlinger2026-xw}.\\
\\
The rapid development of pathology foundation models has, however, outpaced our ability to assess their limitations rigorously \cite{mahmoodBenchmarkingCrisis2025a}. A central vulnerability is their sensitivity to batch effects introduced by tissue preparation, sectioning, staining, and whole-slide scanning, which are strongly encoded in their feature spaces \cite{kömen2024batcheffects,dejong2025unrobust}. This vulnerability follows from how these models are pretrained: self-supervised learning rewards representations that capture recurring structure in the data, allowing technical artefacts that alter the appearance of pathology images to be encoded alongside biology. Such co-encoding becomes problematic when technical variation aligns with clinical labels. If, for example, one institution contributes mostly malignant cases and another mostly benign cases, a downstream model can exploit institution-specific staining or scanner signatures as shortcuts. In that setting, predictions may reflect acquisition artefacts rather than morphology, undermining generalisability across cohorts and institutions \cite{Geirhos2020-wa,Howard2021-wx,Dehkharghanian2023-oy}. Robustness to these confounders is therefore a core requirement for developing models that can support reliable clinical deployment.\\
\\
Several lines of work address this problem. On the mitigation side, stain normalisation and colour augmentation are established remedies from the supervised era \cite{tellez2019}, a rationale that now extends into self-supervised pretraining itself, where augmentation policies designed for natural images are being adapted to pathology \cite{zimmermann2024}. On the measurement side, Kömen et al. showed that batch effects nonetheless persist in the representations of current foundation models, whose embeddings predict the contributing medical centre with high accuracy \cite{kömen2024batcheffects}, and de Jong et al. showed that centre differences can dominate local representation geometry, introducing the Robustness Index (\code{RI}) to quantify this effect \cite{dejong2025unrobust}. Filiot et al. address both, assessing robustness to staining and scanner variation on the PLISM dataset and showing that distillation can improve it \cite{filiot2025plism}. Related measures have a longer history in single-cell genomics, where mixing statistics such as kBET and LISI quantify how well cells from different batches interleave in embedding space \cite{buttner2019kbet,korsunsky2019harmony,luecken2022scib}. Like \code{RI}, these summarise the label composition of local neighbourhoods rather than distance margins, and are typically reported as pooled dataset-level scores.\\
\\
Kömen et al. recently released PathoROB, a public benchmark of pathology foundation model robustness to non-biological variation that spans several datasets and downstream measures~\cite{pathorob}. Its central metric is \code{RI}, which asks whether local neighbourhoods in representation space are organised primarily by biology or by confounder. For each anchor, that is, each sample whose local neighbourhood is being scored, \code{RI} compares cross-confounder biological matches with same-confounder biological distractors, yielding an interpretable count-based readout of local organisation. However, this design has three structural limitations. First, because \code{RI} is count-based, it discards geometry: two models can have identical neighbour counts while placing biological matches and confounder-driven distractors at very different distances from the anchor (Figure \ref{fig:concept-ri-mari}, left panel). Second, \code{RI} exists only as a pooled aggregate: no per-sample score is defined, so the distribution of robustness across the cohort, and any vulnerable subset within it, cannot be examined. Third, \code{RI} is defined only for anchors whose fixed-$k$ neighbourhood contains the typed neighbours needed to form the biological-versus-confounder contrast. Anchors lacking such neighbours are silently excluded from the pooled score: different models can be evaluated on different effective subsets of the data, complicating cross-model comparisons (Figure \ref{fig:concept-undefined}).\\
\\
To address these limitations, we introduce the Cross-confounder Robustness Margin (\code{CRoMa}), a geometry-aware measure of model robustness in representation space. Rather than counting neighbours within a fixed neighbourhood, \code{CRoMa} measures, for each sample, the distance margin between cross-confounder biological matches and same-confounder distractors. This formulation requires no per-model neighbourhood tuning and is defined on every sample, enabling finer-grained analyses that can reveal confounder-dominated subgroups obscured by pooled averages. We evaluate \code{CRoMa} across \TileRankedNModels{} tile-level pathology foundation models on three benchmarks, finding consistent model rankings across datasets ($\rho\approx$ \CrossCohortRhoMean{}). To extend robustness evaluation beyond tile encoders, we further introduce PCaBiop, a purpose-built prostate biopsy benchmark on which we evaluate \SlideNModels{} slide-level foundation models. Across these evaluations, \code{CRoMa} closely tracks downstream shortcut learning, linking representation-level robustness to downstream model behaviour.

\section{Results}
\label{sec:results}

\begin{figure}[!htbp]
\centering
\begin{subfigure}{\linewidth}
  \centering
  \includegraphics[width=\linewidth]{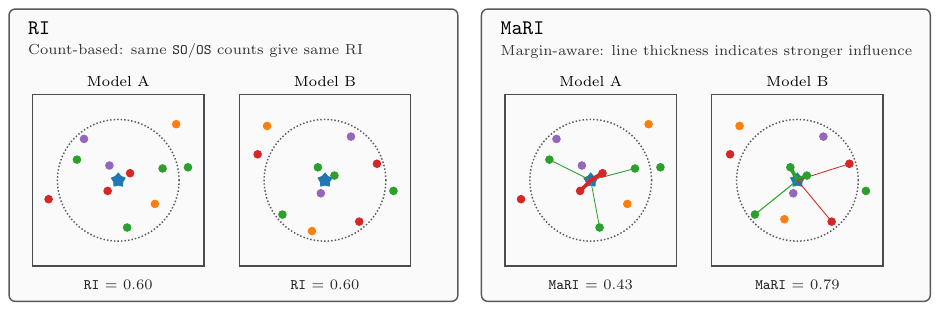}
  \caption{Count-identical neighbourhoods can differ in margin.}
  \label{fig:concept-ri-mari}
\end{subfigure}

\vspace{0.9em}

\begin{subfigure}[b]{0.36\linewidth}
  \centering
  \includegraphics[height=5.6cm]{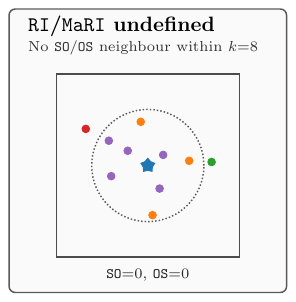}
  \caption{Fixed-$k$ metrics can be undefined.}
  \label{fig:concept-undefined}
\end{subfigure}
\hfill
\begin{subfigure}[b]{0.60\linewidth}
  \centering
  \includegraphics[height=5.6cm]{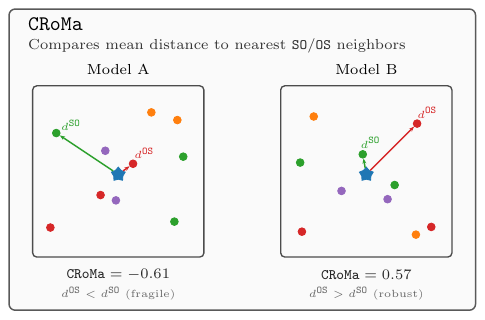}
  \caption{\code{CRoMa} compares nearest typed distances.}
  \label{fig:concept-croma}
\end{subfigure}

\vspace{0.7em}
\includegraphics[width=0.8\linewidth]{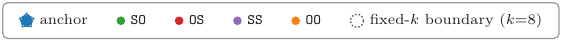}

\caption{\textbf{Count- and distance-based robustness metrics.} \textbf{SO} neighbours share biology but differ in confounder, whereas \textbf{OS} neighbours differ in biology but share the confounder. \textbf{a}, \code{RI} counts typed neighbours within a fixed-$k$ neighbourhood and gives both models the same score. \code{MaRI} weights the same evidence by distance (edge width denotes weight), yielding different scores for the two models. \textbf{b}, If the fixed neighbourhood contains no typed neighbours, \code{RI} and \code{MaRI} are undefined for that sample, which does not contribute to the pooled score. \textbf{c}, \code{CRoMa} compares the mean distance to the nearest typed neighbours, yielding a value for every sample by construction. Positive values denote biology-dominant geometry and negative values confounder-dominant geometry.}
\label{fig:concept-metrics}
\end{figure}
 % fig:concept-metrics (Figure 1)

\noindent
For each sample (the \emph{anchor}), we identify its typed neighbours in representation space: \textbf{SO} neighbours share its biology but differ in the confounder, whereas \textbf{OS} neighbours differ in biology but share the confounder (Figure~\ref{fig:concept-metrics}). The Robustness Index (\code{RI}) counts these typed neighbours within a fixed neighbourhood of size $k$ and pools the counts over the cohort into a single score (Methods). We first examine what this construction can and cannot detect.\\
\\
We use Camelyon as the primary benchmark and then test whether the findings hold across datasets and representation scales. Camelyon is the largest tile-level cohort ($20{,}400$ tiles versus $9{,}000$ for Tolkach-ESCA and $5{,}760$ for TCGA-4x4; Table~\ref{tab:dataset-summary}), spans the widest range of robustness, from clearly biology-dominant to clearly confounder-dominant encoders (Section~\ref{sec:cross-benchmark}), and is largely absent from disclosed pretraining corpora (only \code{GPFM} reports it\footnote{Some pretraining corpora are only partially disclosed, so residual overlap cannot be excluded.}; Table~\ref{tab:model-summary}). It is also the benchmark on which fixed-$k$ scores are least supported: only $10$--$46\%$ of samples contribute to \code{RI}, against $67$--$100\%$ on Tolkach-ESCA and $99$--$100\%$ on TCGA-4x4.

\subsection{Distance weighting exposes the limits of fixed-$k$ robustness scores}
\label{sec:mari-fixed-k}

Because \code{RI} is count-based, it captures \emph{how many} typed neighbours appear within the selected neighbourhood, but not \emph{where} they appear. \code{MaRI} is the minimal geometric correction: it keeps \code{RI}'s fixed-$k$, pooled construction but weights each typed neighbour by its distance to the anchor, so that identical counts can yield different scores when the neighbours occupy different positions (Methods). Empirically, this correction is modest (Table \ref{tab:main-results}). On Camelyon, $\Delta=\code{MaRI}-\code{RI}$ remains within $\pm$\CamelyonMariRiMaxAbsDelta{} for every encoder, and model rankings are nearly unchanged (Spearman $\rho=$\CamelyonRiVsMariRho; Supplementary Table~\ref{tab:rank-agreement}). The sign of $\Delta$, however, is informative: it is positive when cross-confounder biological matches are systematically closer than same-confounder distractors (\code{UNI2-h}, $+0.033$; \code{Virchow2}, $+0.017$), and negative when distractors are closer (\code{Midnight-12k}, $-0.070$; \code{Virchow}, $-0.043$).\\
\\
Two properties of the fixed-$k$, pooled design explain why \code{MaRI} cannot depart far from \code{RI}. First, a sample contributes to either score only if its neighbourhood contains at least one \textbf{SO} or \textbf{OS} neighbour; otherwise it is silently excluded. On Camelyon this exclusion is severe: at the operating $k=11$, fewer than half of all samples contribute for every encoder, and fewer than a quarter for more than half of them (Table~\ref{tab:main-results}), so the effective evaluation cohort is model-dependent. The cause is a mismatch between the selected neighbourhood and the rank at which typed neighbours appear: the first \textbf{SO} or \textbf{OS} neighbour occurs at a median rank of $\approx149$ among $20{,}400$ candidates (Supplementary Table~\ref{tab:typed-neighbour-coverage}), because $k$ is chosen to maximise biological $k$-NN accuracy, a criterion dominated by the same-biology, same-confounder (\textbf{SS}) neighbours that \code{RI} and \code{MaRI} discard (Supplementary Section~\ref{supp:k-selection}). Second, among contributing samples the two scores can differ only in neighbourhoods containing both \textbf{SO} and \textbf{OS} neighbours, which account for a median of $4.6\%$ of contributing samples and at most $10.6\%$ for any encoder (Supplementary Table~\ref{tab:typed-neighbour-coverage}). Within these, distance corrections point in either direction and cancel when pooled, so substantial sample-level differences reduce to a small change in the pooled score.\\
\\
These observations motivate a per-sample margin that does not depend on a pre-specified neighbourhood (Section~\ref{sec:croma}).

\subsection{\code{CRoMa} defines a typed margin beyond fixed neighbourhoods}
\label{sec:croma}

\code{CRoMa} recasts robustness as a direct distance comparison: for each sample, it asks whether cross-confounder biological matches (\textbf{SO}) lie closer than same-confounder biological distractors (\textbf{OS}). Rather than restricting the contrast to typed neighbours that happen to fall inside a fixed-$k$ window, \code{CRoMa} searches outward to the $m$ nearest neighbours of each type and compares their mean distances (Methods). This construction yields a signed margin in $(-1,1)$, positive when biological matches across confounders are closer than confounder-matched distractors, and negative otherwise. Whether a sample receives a score depends only on the benchmark's label composition---at least $m$ neighbours of each type must exist---and never on the model, so every encoder is evaluated on exactly the same population. On all benchmarks used here, that population is the full evaluation cohort.\\
\\
All encoders strongly encode both biological class and acquisition centre. We quantify this with two $k$-NN probes applied directly to the frozen embeddings: each tile is classified by majority vote among its nearest neighbours, once to predict its biological label (tumour versus normal) and once to predict its acquisition centre, and each probe is summarised by its balanced accuracy (Methods). On Camelyon, the biological probe is near-ceiling ($0.93$--$0.99$), while the centre probe is almost perfect ($0.91$--$1.00$) across the \CamelyonRankedNModels{} pathology encoders (Table~\ref{tab:main-results}). What separates a robust representation from a shortcut-prone one is therefore not whether each signal is present, but which of the two more strongly structures the representation when they compete. We quantify that competition with the median per-sample \code{CRoMa} margin, which asks whether the typical sample lies closer to cross-confounder biological matches or to same-confounder biological distractors. Model-level median \code{CRoMa} values ranged from \CamelyonCromaMin{} to \CamelyonCromaMax{} (Table~\ref{tab:main-results}). Thirteen of the \CamelyonRankedNModels{} encoders had positive medians, indicating biology-dominant geometry for the typical sample, whereas seven had negative medians. The six highest medians ranged from $0.16$ to \CamelyonCromaMax{}, with \code{Virchow2} and \code{CONCH} highest (both \CamelyonCromaMax{}). Values crossed zero gradually: \code{Prov-GigaPath} ($0.01$) and \code{UNI} ($-0.03$) lay close in magnitude despite falling on opposite sides of the boundary. The most negative medians were observed for \code{Prost40M} ($-0.32$) and \code{Hibou-L} (\CamelyonCromaMin{}).\\
\\
Although \code{CRoMa} evaluates the full cohort rather than a model-dependent subset of anchors, its model ranking broadly agrees with the fixed-$k$ scores where they are defined ($\rho=$\CamelyonCromaVsRiRho{} versus \code{RI}; $\rho=$\CamelyonCromaVsMariRho{} versus \code{MaRI}). The metrics are nevertheless not interchangeable. \code{H-optimus-1} and \code{Virchow} are close under \code{MaRI} ($0.677$ versus $0.708$) yet separate clearly under \code{CRoMa} ($0.08$ versus $0.16$; Table~\ref{tab:main-results}), whereas \code{UNI2-h} and \code{Prov-GigaPath} have nearly identical \code{CRoMa} medians ($0.04$ versus $0.01$) but differ substantially under \code{MaRI} ($0.548$ versus $0.369$). Such differences are expected: the fixed-$k$ scores for these encoders rest on $13$--$26\%$ of anchors, whereas \code{CRoMa} scores every anchor at the rank where its typed neighbours actually occur.\\
\\
Whether robust or not, almost every encoder recovers the biological label (tumour versus normal) near-ceiling, so biological $k$-NN accuracy explains the spread of \code{CRoMa} values only weakly ($\rho=$\CamelyonBioProbeCromaRho{}; Table~\ref{tab:main-results}). \code{CONCH}, \code{CONCHv1.5}, \code{Hibou-B} and \code{Hibou-L}, for example, recover biology near-identically ($k$-NN accuracy $\approx0.97$), yet span almost the entire \code{CRoMa} range. What separates the encoders instead is how decodable the acquisition centre is. Across the \CamelyonProbeNModels{} pathology encoders, the balanced accuracy with which a $k$-NN probe recovers the centre rank-predicts the median \code{CRoMa} almost perfectly ($\rho=$\CamelyonProbeCromaRho{}; Table~\ref{tab:main-results}): the more decodable the centre, the more confounder-dominant the margin. The same probe predicts \code{RI} and \code{MaRI} at least as well ($\rho=$\CamelyonProbeRiRho{} and \CamelyonProbeMariRho{}, respectively), so as model-level ranking statistics, all three robustness summaries largely recapitulate centre decodability. The advantage of \code{CRoMa} lies in its per-sample construction, which yields a distribution of signed margins that can reveal confounder-dominated samples hidden beneath an otherwise favourable aggregate score.

\begin{table}[!htbp]
\centering
\small
\setlength{\tabcolsep}{4pt}
\begin{tabular}{lccccccccc}
\hline
Model & bio bacc & conf bacc & \code{RI} & \code{MaRI} & $\Delta$ & \code{CRoMa} & $F(0)$ & $\mcode{LTM}_{10}$ & support \\
\hline
Virchow2 & \textbf{0.988} & 0.957 & 0.806 & 0.823 & $+0.017$ & \textbf{0.20} & 0.129 & -0.11 & 31.4\% \\
CONCH & 0.971 & 0.954 & 0.662 & 0.626 & $-0.037$ & 0.20 & 0.225 & -0.20 & 35.9\% \\
GenBio-PathFM & 0.983 & 0.927 & \textbf{0.842} & \textbf{0.850} & $+0.008$ & 0.19 & \textbf{0.092} & \textbf{-0.07} & 38.3\% \\
CONCHv1.5 & 0.971 & 0.913 & 0.774 & 0.763 & $-0.012$ & 0.19 & 0.174 & -0.14 & \textbf{46.3\%} \\
H0-mini & 0.969 & 0.926 & 0.741 & 0.718 & $-0.023$ & 0.17 & 0.180 & -0.16 & 38.7\% \\
Virchow & 0.980 & 0.944 & 0.751 & 0.708 & $-0.043$ & 0.16 & 0.221 & -0.18 & 26.4\% \\
Midnight-12k & 0.976 & 0.977 & 0.478 & 0.408 & $-0.070$ & 0.11 & 0.354 & -0.35 & 19.8\% \\
H-optimus-1 & 0.986 & 0.978 & 0.664 & 0.677 & $+0.013$ & 0.08 & 0.219 & -0.14 & 17.2\% \\
H-optimus-0 & 0.982 & 0.965 & 0.659 & 0.652 & $-0.007$ & 0.05 & 0.315 & -0.15 & 23.7\% \\
UNI2-h & 0.986 & 0.986 & 0.515 & 0.548 & $+0.033$ & 0.04 & 0.370 & -0.21 & 13.0\% \\
MUSK & 0.958 & 0.983 & 0.366 & 0.297 & $-0.069$ & 0.04 & 0.403 & -0.22 & 28.0\% \\
mSTAR & 0.979 & 0.984 & 0.460 & 0.434 & $-0.026$ & 0.02 & 0.418 & -0.18 & 18.0\% \\
Prov-GigaPath & 0.979 & 0.990 & 0.375 & 0.369 & $-0.007$ & 0.01 & 0.470 & -0.19 & 14.4\% \\
UNI & 0.982 & 0.999 & 0.108 & 0.092 & $-0.015$ & -0.03 & 0.651 & -0.22 & 9.6\% \\
Hibou-B & 0.973 & 0.999 & 0.057 & 0.041 & $-0.017$ & -0.09 & 0.737 & -0.36 & 13.4\% \\
GPFM & 0.955 & 0.999 & 0.034 & 0.017 & $-0.017$ & -0.10 & 0.753 & -0.36 & 20.9\% \\
Phikon & 0.955 & 1.000 & 0.009 & 0.004 & $-0.005$ & -0.20 & 0.905 & -0.48 & 16.6\% \\
Phikon-v2 & 0.954 & 0.999 & 0.019 & 0.008 & $-0.011$ & -0.21 & 0.932 & -0.50 & 16.9\% \\
Prost40M & 0.926 & 0.999 & 0.015 & 0.002 & $-0.012$ & -0.32 & 0.922 & -0.64 & 27.2\% \\
Hibou-L & 0.971 & 1.000 & 0.013 & 0.001 & $-0.011$ & -0.44 & 0.993 & -0.66 & 12.1\% \\
\hline
DINOv2-B & 0.919 & 0.912 & 0.561 & 0.507 & $-0.053$ & 0.05 & 0.345 & -0.18 & 68.0\% \\
\hline
\end{tabular}
\caption{\textbf{Representation robustness on Camelyon.} Results for 20 tile-level pathology foundation models, ordered by median \code{CRoMa} ($m{=}5$), with the natural-image control \code{DINOv2-B} shown separately. All models are evaluated at the shared operating point $k{=}11$, the dataset median of the per-model biological $k^\star$. Columns: biological and confounder $k$-NN balanced accuracy; pooled \code{RI} and \code{MaRI}; $\Delta{=}\code{MaRI}-\code{RI}$; median \code{CRoMa}; $F(0)$, the fraction with $\mcode{CRoMa}\leq0$; $\mcode{LTM}_{10}$, the mean of the lowest decile; and support, the fraction of samples effectively contributing to \code{RI}/\code{MaRI}. Bold denotes the best value in each score column (conf bacc and $\Delta$ are diagnostics).}
\label{tab:main-results}
\end{table}
 % tab:main-results (Camelyon)

\subsection{Per-sample margins expose hidden fragility}
\label{sec:tail}

\begin{figure}[!htbp]
\centering
\includegraphics[width=0.78\linewidth]{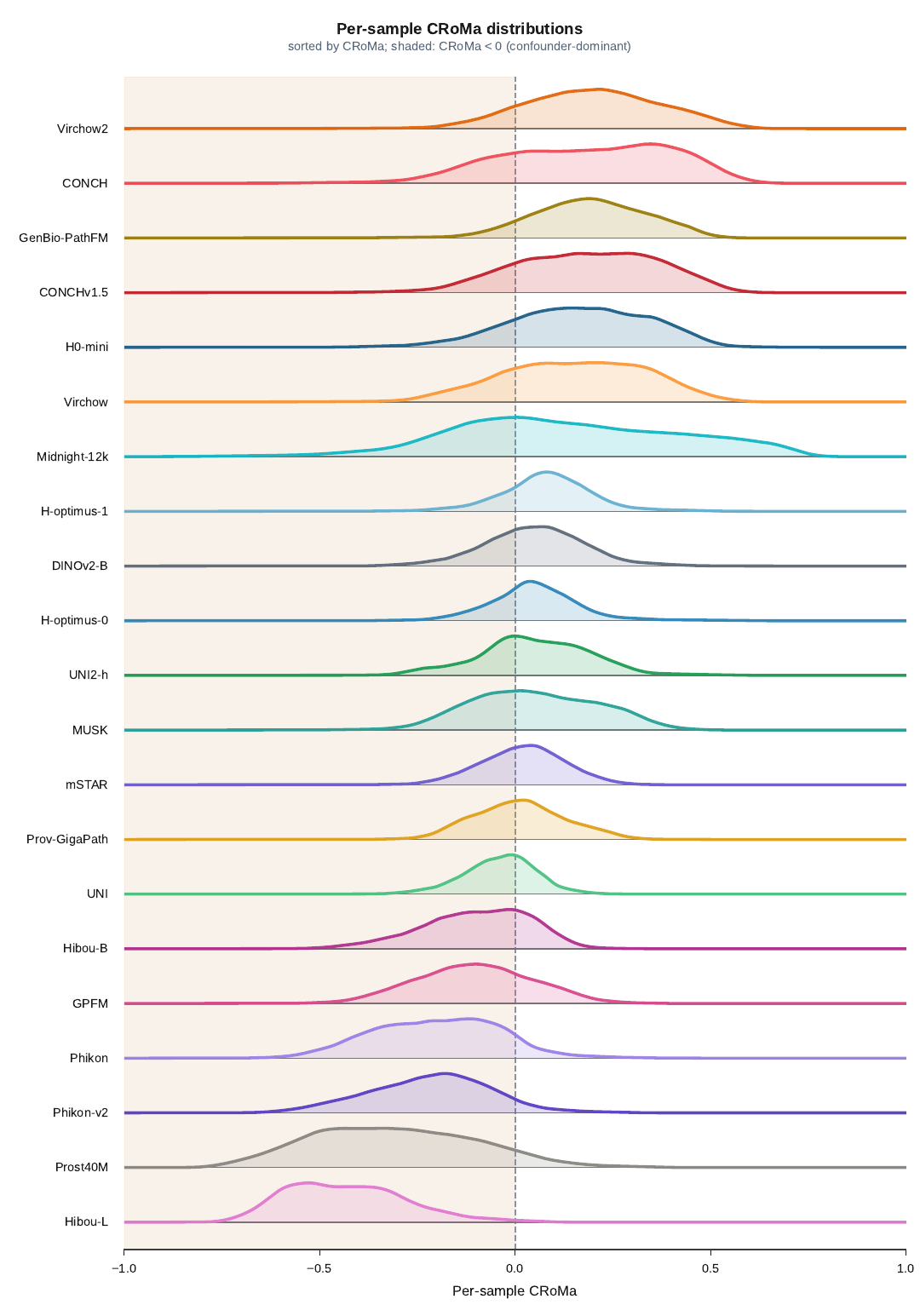}
\caption{\textbf{Per-sample \code{CRoMa} distributions on Camelyon.} Ridgeline distributions for the \CamelyonRankedNModels{} pathology encoders and the natural-image control (\code{DINOv2-B}), ordered by median \code{CRoMa}. The dashed line denotes $\mcode{CRoMa}=0$, while shading denotes the confounder-dominant region ($\mcode{CRoMa}<0$).}
\label{fig:croma-distribution}
\end{figure}

\begin{figure}[!htbp]
\centering
\includegraphics[width=0.7\linewidth]{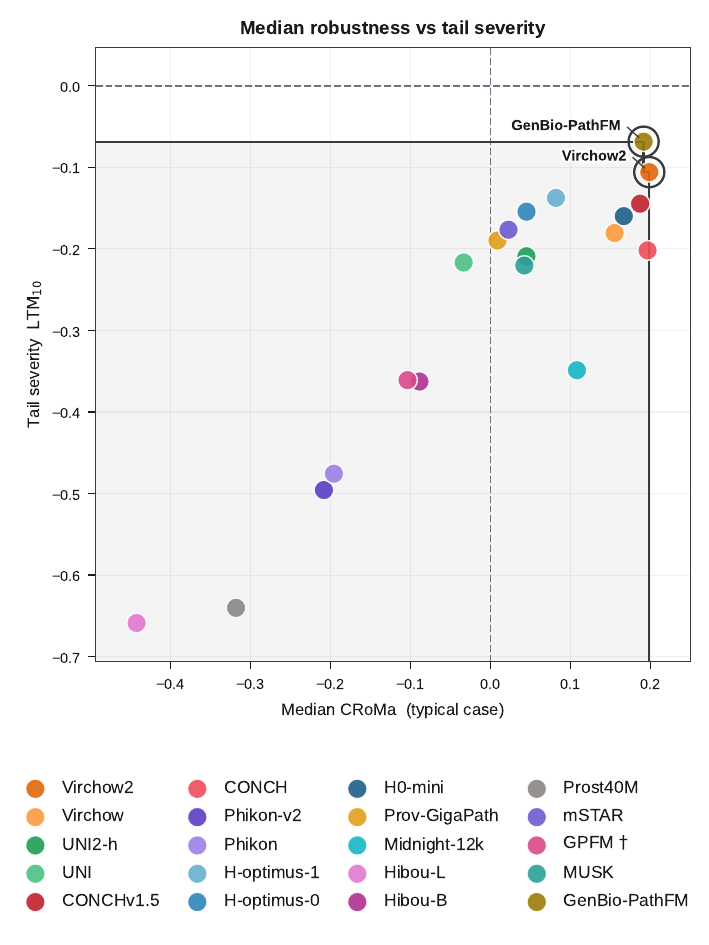}
\caption{\textbf{Median and lower-tail \code{CRoMa} on Camelyon.} Median \code{CRoMa} is plotted against worst-decile mean $\mcode{LTM}_{10}$ for the \CamelyonRankedNModels{} pathology encoders evaluated in this study. Higher values are preferable on both axes. Ringed points form the upper-right Pareto frontier, while shaded points are dominated on both axes. $\dagger$ marks the Camelyon-exposed encoder (\code{GPFM}).}
\label{fig:croma-pareto}
\end{figure}

Figure \ref{fig:croma-distribution} shows the full distribution of per-sample \code{CRoMa} margins for each encoder on Camelyon. The median \code{CRoMa} summarises the robustness of the typical sample, but it does not capture \emph{how much} of the distribution extends below zero, nor \emph{how far} into the confounder-dominant regime that lower tail reaches. Models with similar medians can consequently differ substantially in the prevalence and severity of vulnerable samples. We quantify this lower-tail structure using two complementary quantities: the fraction of non-positive margins, $F(0)$, and the mean of the worst decile, $\mcode{LTM}_{10}$ (Methods).\\
\\
On Camelyon, no representation is uniformly robust. Every one of the \CamelyonRankedNModels{} pathology encoders retains a confounder-dominated lower tail: all have $\mcode{LTM}_{10}<0$ (Table~\ref{tab:main-results}). The fraction of non-positive margins varies widely, from $12.9\%$ for \code{Virchow2}, the leading encoder by median margin, to $99.3\%$ for \code{Hibou-L}. Importantly, median performance, the non-positive fraction and tail severity capture distinct aspects of robustness. Although \code{CONCH} matches \code{Virchow2} at the median (\CamelyonCromaMax{}), it has more non-positive margins ($F(0)=22.5\%$ versus $12.9\%$) and a more severe lower tail ($\mcode{LTM}_{10}=-0.20$ versus $-0.11$). Conversely, \code{CONCH} has fewer non-positive margins than \code{H-optimus-0} ($22.5\%$ versus $31.5\%$), but a more severe lower tail ($-0.20$ versus $-0.15$). A robust encoder should combine a biology-dominant median with a mild lower tail. Plotting encoders along these two axes makes the trade-off explicit (Figure~\ref{fig:croma-pareto}). Only two of the \CamelyonRankedNModels{} pathology encoders lie on the resulting Pareto frontier: \code{Virchow2} and \code{GenBio-PathFM}. For every other encoder, at least one alternative performs no worse on either axis and better on at least one.\\
\\
These scalars summarise, but do not replace, the full margin distribution. \code{Midnight-12k} and \code{H-optimus-1}, for example, have similar medians ($0.11$ and $0.08$) but very different shapes: the margins of \code{H-optimus-1} concentrate around the median, whereas those of \code{Midnight-12k} span nearly the full range (Figure~\ref{fig:croma-distribution}).

% AUTO-GENERATED by scripts/repro/generate_distribution_floats.py -- do not edit by hand.
% The PDF is drawn by scripts/repro/figures/croma_distribution_figure.py, into
% output/metrics/median-k/pathorob-camelyon/studies/plots/pdf/croma_distribution.pdf.
% It is NOT staged: copy that file to paper/figures/results/pathorob-camelyon/pdf/croma_distribution.pdf to resolve the
% \includegraphics below (\graphicspath reaches paper/figures/, not output/).

% AUTO-GENERATED by scripts/repro/generate_pareto_float.py -- do not edit by hand.
% The PDF is drawn by scripts/repro/figures/croma_pareto_figure.py, into
% output/metrics/median-k/pathorob-camelyon/studies/plots/pdf/croma_pareto.pdf.
% It is NOT staged: copy that file to paper/figures/results/pathorob-camelyon/pdf/croma_pareto.pdf to resolve the
% \includegraphics below (\graphicspath reaches paper/figures/, not output/).

\subsection{Robustness rankings are consistent across benchmarks}
\label{sec:cross-benchmark}

\begin{figure}[!htbp]
\centering
\includegraphics[width=0.7\linewidth]{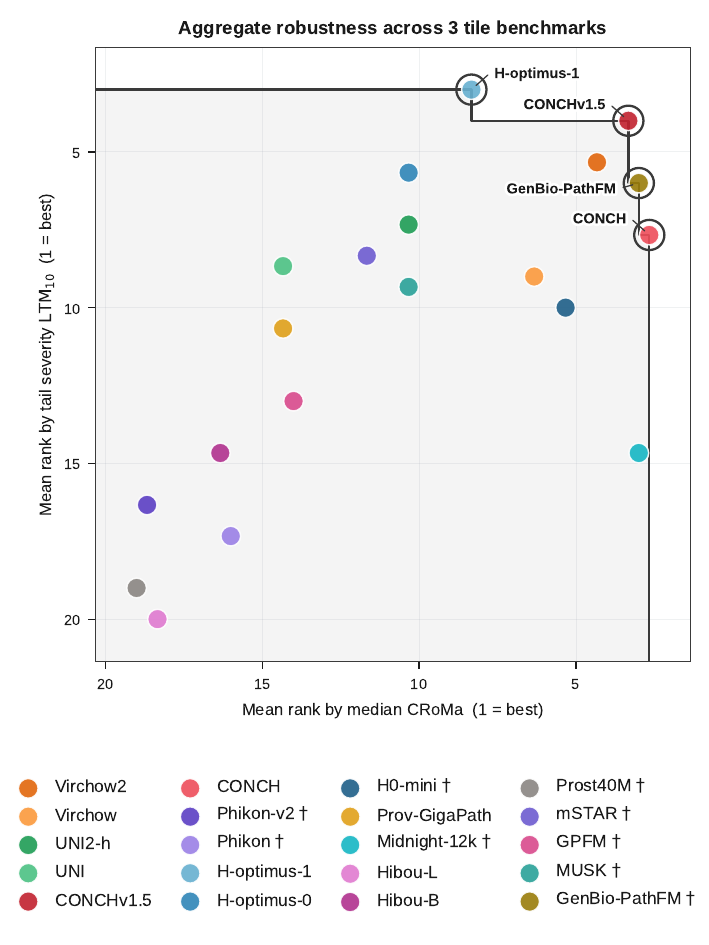}
\caption{\textbf{Median and lower-tail \code{CRoMa} ranks across tile benchmarks.} Each of the 3 tile benchmarks ranks the 20 pathology encoders by median \code{CRoMa} (rank $1$ = highest median) and by worst-decile mean $\mcode{LTM}_{10}$ (rank $1$ = mildest tail). The axes show the mean rank across benchmarks, with lower (=better) ranks plotted toward the upper-right corner. Ringed points form the Pareto frontier, while shaded points are dominated on both axes. $\dagger$ marks the 9 TCGA-exposed encoders.}
\label{fig:croma-pareto-rank}
\end{figure}

We next asked whether the Camelyon findings hold on TCGA-4x4 (Supplementary Table~\ref{tab:main-results-tcga4x4}) and Tolkach-ESCA (Supplementary Table~\ref{tab:main-results-tolkach}); PathoROB's paired TCGA-2x2 configuration is reported in Supplementary Section~\ref{supp:tcga}.\\
\\
Both datasets provide substantially higher \code{RI}/\code{MaRI} support: $99$--$100\%$ on TCGA-4x4 and $67$--$100\%$ on Tolkach-ESCA (against $10$--$46\%$ on Camelyon). This follows from the larger operating points these cohorts select ($k{=}71$ and $61$ respectively, against $k{=}11$ on Camelyon), since a wider fixed neighbourhood is more likely to contain a typed \textbf{SO}/\textbf{OS} contrast. Near-complete support removes the exclusion effect that dominated on Camelyon, yet \code{MaRI} still barely departs from \code{RI}. The correction $\Delta$ stays within $\pm$\TcgaFourByFourMariRiMaxAbsDelta{} on TCGA-4x4 and $\pm$\TolkachMariRiMaxAbsDelta{} on Tolkach-ESCA across all \TcgaFourByFourRankedNModels{} pathology encoders, and rankings remain nearly identical (Spearman $\rho=$\TcgaFourByFourRiVsMariRho{} and \TolkachRiVsMariRho{}). The near-equivalence is therefore intrinsic to the pooled design rather than a by-product of thin support. Where fixed-$k$ scores are defined, \code{CRoMa} again broadly tracks their rankings ($\rho=$\TcgaFourByFourCromaVsRiRho{} and \TolkachCromaVsRiRho{} versus \code{RI}, \TcgaFourByFourCromaVsMariRho{} and \TolkachCromaVsMariRho{} versus \code{MaRI} on TCGA-4x4 and Tolkach-ESCA), matching the agreement seen on Camelyon (Supplementary Table~\ref{tab:rank-agreement}).\\
\\
While several encoders had negative or near-zero median margins on Camelyon, median \code{CRoMa} values were generally higher on TCGA-4x4 and Tolkach-ESCA, spanning \TcgaFourByFourCromaSpan{} and \TolkachCromaSpan{}, respectively. Only \TcgaFourByFourCromaConfounderDominantWord{} of the \TcgaFourByFourRankedNModels{} encoders is confounder-dominant on TCGA-4x4, and \TolkachCromaConfounderDominantWord{} on Tolkach-ESCA. However, this upward shift did not eliminate the failure modes observed on Camelyon: even when the typical sample is robust, the worst decile remains confounder-dominant for every encoder ($\mcode{LTM}_{10} < 0$; Supplementary Tables~\ref{tab:main-results-tcga4x4} and~\ref{tab:main-results-tolkach}). The per-sample distributions illustrate this: on both benchmarks every encoder keeps mass on the confounder-dominant side of the boundary (Supplementary Figures \ref{fig:croma-distribution-tcga4x4} and \ref{fig:croma-distribution-tolkach}). Model rankings were largely stable across datasets (pairwise Spearman $\rho\in$\CrossCohortRhoRange{}). Encoders ranked among the most robust on Camelyon generally remained competitive on TCGA-4x4 and Tolkach-ESCA, whereas low-margin models tended to remain lower ranked. The \code{CONCH} family and \code{GenBio-PathFM} were consistently among the top five, with \code{H0-mini} close behind on all three and \code{Virchow2} leading Camelyon but slipping to fifth and seventh elsewhere. \code{Prost40M}, \code{Hibou-B} and the \code{Phikon} models remained near the bottom.\\
\\
The main exception was \code{Midnight-12k}, which ranked seventh on Camelyon (\code{CRoMa}$=0.11$) but first on both TCGA-4x4 (\TcgaFourByFourCromaMax{}) and Tolkach-ESCA (\TolkachCromaMax{}). Because \code{\ProvenanceModel} was pretrained exclusively on TCGA, its performance on TCGA-4x4 may partly reflect pretraining–benchmark overlap, and a within-cohort provenance analysis supports this interpretation (Supplementary Section~\ref{supp:pretraining-overlap}). Overlap alone does not explain the result, however: other TCGA-pretrained models showed no comparable advantage on TCGA-4x4, \code{Phikon} remaining weak and \code{Phikon-v2} showing no rank shift.\\
\\
Because median performance and tail severity capture distinct robustness properties, cross-benchmark consistency is best assessed in the median–tail plane rather than through a single ranking. Supplementary Figures~\ref{fig:croma-pareto-tcga4x4} and~\ref{fig:croma-pareto-tolkach} show that the pattern observed on Camelyon extends to TCGA-4x4 and Tolkach-ESCA: the encoder with the highest median margin does not have the least severe worst decile, no encoder is best on both axes, and only a small subset lies on the Pareto frontier. We aggregate results for the three tile-level benchmarks by averaging model rankings: Figure~\ref{fig:croma-pareto-rank} plots mean median-\code{CRoMa} rank against mean $\mcode{LTM}_{10}$ rank. Across benchmarks, only four of the \TileRankedNModels{} encoders remain undominated -- \code{CONCH}, \code{GenBio-PathFM}, \code{CONCHv1.5} and \code{H-optimus-1} -- and none leads on both axes.

% AUTO-GENERATED by scripts/repro/generate_rank_pareto_float.py -- do not edit by hand.
% The PDF is drawn by scripts/repro/figures/rank_pareto_figure.py, into
% output/studies/rank-pareto/plots/pdf/rank_pareto.pdf.
% It is NOT staged: copy that file to paper/figures/rank_pareto.pdf to resolve the
% \includegraphics below (\graphicspath reaches paper/figures/, not output/).

\subsection{Confounder dominance extends to whole-slide representations}
\label{sec:slide-level}

Slide-level foundation models are attractive for clinical prediction tasks because they provide a pretrained aggregator, yet their robustness has so far not been measured. We therefore characterised four widely used slide-level encoders on PCaBiop, a benchmark of $1{,}000$ prostate biopsies from two centres (Supplementary Section~\ref{supp:panda}). The contributing centre remained near-perfectly decodable for all four encoders, median margins spanned \PandaCromaSpan{}, and only \code{PRISM} was biology-dominant (\code{CRoMa}$=$\PandaBestCroma{}). \code{MOOZY} attained the highest biological accuracy yet a slightly negative median margin, and \code{PRISM} and \code{MOOZY} differed sharply in median margin and non-positive fraction while having near-identical worst-decile severity ($\mcode{LTM}_{10}=-0.39$ versus $-0.41$).\\
\\
With a finer biological label, every encoder became confounder-dominant. On PCaBiop-ISUP, which expands the cohort to $3{,}000$ biopsies balanced across six ISUP grade groups and both centres, median margins spanned \PandaIsupCromaSpan{}, although differences in cohort composition and evaluation design preclude attributing this shift to label granularity alone.

\subsection{\code{CRoMa} predicts downstream shortcut susceptibility}
\label{sec:nipd}

\begin{figure}[!htbp]
\centering
\includegraphics[width=0.92\linewidth]{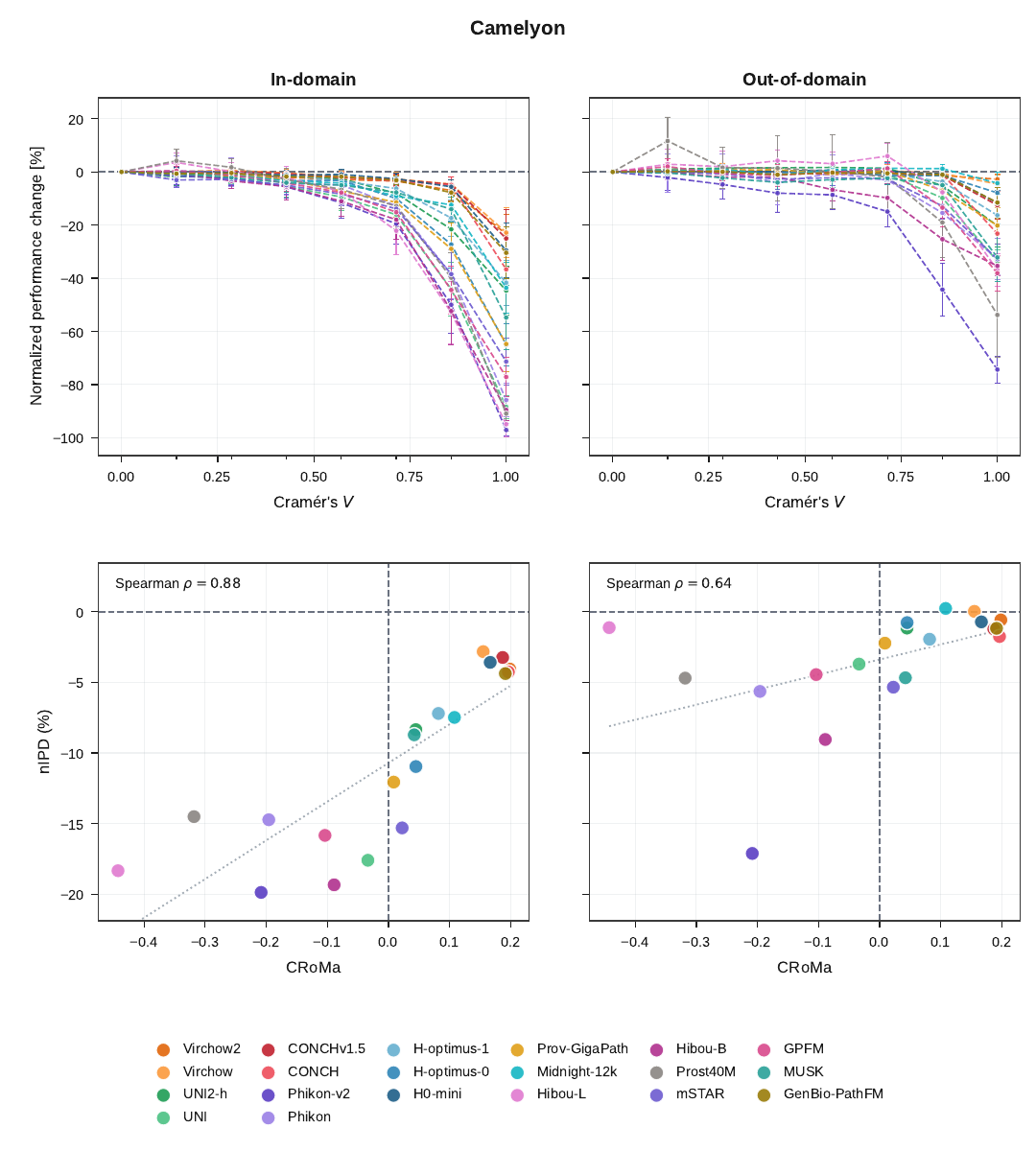}
\caption{\textbf{Representation robustness and downstream shortcut susceptibility on Camelyon.} \code{CRoMa} measures robustness in frozen representation space: higher values indicate more biology-dominant geometry. \code{nIPD} summarises downstream linear-probe degradation (Methods): lower values indicate greater susceptibility. Columns show evaluation on acquisition groups represented during training (left, in domain) and unseen acquisition groups (right, out of domain). \textbf{Top row:} normalised performance change $g(V)$ relative to the above-chance performance at $V=0$ as the training-set confounder--biology correlation increases to $V{=}1$. Curves show the $20$ tile encoders and error bars show 95\% Student's $t$ confidence intervals for the mean normalised change across 20 resampling runs. \textbf{Bottom row:} median \code{CRoMa}$(m{=}5)$ versus \code{nIPD}, one point per encoder. Dashed lines mark $\mcode{CRoMa}=0$ and $\mcode{nIPD}=0$.}
\label{fig:croma-nipd-camelyon}
\end{figure}

To test whether frozen-embedding margins anticipate shortcut learning by a supervised probe, we applied PathoROB's confounder-biased linear-probe sweep \cite{pathorob} and summarised degradation using normalised integrated performance degradation (\code{nIPD}), which expresses the loss relative to the above-chance performance available before a shortcut is introduced (Methods). Because OOD evaluation requires held-out acquisition domains, the downstream analysis includes samples that are not used to compute the representation metrics (Table~\ref{tab:downstream-rowset}).\\
\\
Across the \ApdNModels{} tile-level pathology encoders, higher median \code{CRoMa} was associated with smaller shortcut-induced degradation in every benchmark (Figure~\ref{fig:croma-nipd-camelyon}; Supplementary Figures~\ref{fig:croma-nipd-tcga4x4} and~\ref{fig:croma-nipd-tolkach}). The association was strongest in domain, where the controlled test set most directly isolates susceptibility to the engineered shortcut (Spearman $\rho=\NipdIdCromaCamelyon{}$, $\NipdIdCromaTcga{}$ and $\NipdIdCromaTolkach{}$ for Camelyon, TCGA-4x4 and Tolkach-ESCA, respectively; Supplementary Table \ref{tab:apd-correlation}). Thus, models whose representations were more biology-dominant retained a larger fraction of their baseline predictive performance as the confounder--biology correlation increased. Where defined, \code{RI} and \code{MaRI} showed comparable in-domain associations ($\rho=\NipdIdRiCamelyon{}$--$\NipdIdRiTolkach{}$ and $\NipdIdMariCamelyon{}$--$\NipdIdMariTolkach{}$; Supplementary Table~\ref{tab:apd-correlation}), consistent with the close agreement of the three model-level rankings. The association remained positive, but was weaker, on held-out acquisition domains ($\rho=\NipdOodCromaCamelyon{}$, $\NipdOodCromaTcga{}$ and $\NipdOodCromaTolkach{}$, respectively). This attenuation is expected because OOD degradation is not a pure shortcut endpoint: it combines reliance on the engineered shortcut with transfer to unseen centres or cohorts. \code{APD}, retained for continuity with PathoROB, yielded the same qualitative conclusion (Supplementary Table~\ref{tab:apd-correlation}).\\
\\
Similar conclusions hold when evaluating downstream degradation at slide level using PCaBiop (Supplementary Figure~\ref{fig:croma-nipd-pcabiop}). \code{MOOZY} and \code{PRISM} underwent less degradation than \code{TITAN} and \code{Prov-GigaPath}, both in domain (\code{nIPD}: \PcabiopMoozyNipdId{}, \PcabiopPrismNipdId{}, \PcabiopTitanNipdId{} and \PcabiopProvGigaPathNipdId{}, respectively) and out-of-domain (\PcabiopMoozyNipdOod{}, \PcabiopPrismNipdOod{}, \PcabiopTitanNipdOod{} and \PcabiopProvGigaPathNipdOod{}). \code{Prov-GigaPath} also illustrates the baseline-headroom limitation of \code{APD}: its OOD balanced accuracy at $V=0$ was \PcabiopProvGigaPathOodBaseline{}, leaving little above-chance performance, so \code{APD} recorded a decrease of only \PcabiopProvGigaPathApdOod{} whereas \code{nIPD} recorded \PcabiopProvGigaPathNipdOod{}. Together, these findings identify \code{CRoMa} as a representation-level indicator of shortcut susceptibility that is available before task-specific adaptation.

% Main-text figure floats. fig:croma-distribution is not wired here: a float can only be
% placed at or after its \input point, so results.tex \inputs it inline to keep it in Sec 3.3.
% AUTO-GENERATED by scripts/repro/generate_apd_floats.py -- do not edit by hand.
% PDFs are rendered under output/studies/apd/plots/pdf/croma_nipd_<benchmark>.pdf.

\section{Discussion}
\label{sec:discussion}

A useful pathology foundation model must encode biologically meaningful signal without letting batch effects from tissue preparation, staining or scanning dominate the geometry of its representation space. We showed that existing fixed-neighbourhood robustness scores capture part of this property but characterise neither its geometry nor its distribution across samples. The typed contrast introduced by \code{RI} is conceptually well founded: cross-confounder biological matches signal biological structure, whereas same-confounder distractors indicate that non-biological cues shape local geometry. Yet \code{MaRI}'s close agreement with \code{RI} shows that the limiting factor is not the absence of distance information but the fixed-$k$, pooled design itself, which excludes samples in a model-dependent manner and lets opposing sample-level asymmetries cancel. \code{CRoMa} addresses both by defining a signed distance margin for every sample and treating aggregate scores as summaries of a distribution rather than as the estimand itself.\\
\\
Treating robustness as a distribution reveals what no pooled aggregate can: no foundation model is uniformly robust. Even models with biology-dominant median margins retain confounder-dominated samples, and every evaluated tile encoder exhibits a negative lower tail. The resulting picture is not a division between robust and non-robust encoders, but a spectrum of robustness profiles in which central tendency, the non-positive fraction and lower-tail severity vary partly independently. Model selection is therefore better framed as a Pareto problem balancing central tendency against lower-tail severity than driven by a single score. After aggregating ranks across the three tile benchmarks, only four encoders remain undominated in the median–tail plane, and none leads on both axes.\\
\\
One encoder nevertheless stood out for its consistency. \code{GenBio-PathFM} was the only model on the median–tail Pareto frontier in all three tile benchmarks. Its JEDI recipe adds JEPA-based refinement to a DINO-family pretraining stage~\cite{genbiopathfm}, departing from the predominantly DINO/DINOv2-only recipes of contemporary pathology encoders, but differences in model scale, data curation, corpus composition and pretraining–evaluation overlap prevent attributing this consistency to JEPA alone. The result nonetheless motivates systematic investigation of hybrid and alternative pretraining objectives for developing more robust pathology foundation models.\\
\\
The broad agreement of \code{CRoMa} rankings across benchmarks suggests that it captures a reproducible property of learned representations, even as the distributions themselves shift with the biological task, confounder and cohort. Robustness is thus neither purely model-intrinsic nor wholly dataset-specific: some encoders consistently privilege biological over technical structure, but the extent and location of their failures remain context-dependent. The slide-level analysis broadened the evaluation beyond tile encoders: among four whole-slide models assessed on PCaBiop, margins ranged from biology-dominant to strongly confounder-dominant.\\
\\
\code{CRoMa} also anticipated the downstream failure mode it is designed to expose: encoders with more biology-dominant margins sustained smaller shortcut-induced performance losses. Because \code{CRoMa} is computed from frozen embeddings, this predictive signal is available before any task-specific predictor is fitted. Comprehensive robustness evaluation across institutions and population strata typically requires large, labelled and harmonised cohorts, together with repeated downstream training --- resources that are often unavailable or prohibitively costly. \code{CRoMa} does not replace external validation, but when evaluated on a compact, deliberately structured multi-institutional cohort, it can help identify representations that are more susceptible to learning shortcuts during task-specific adaptation.\\
\\
Beyond model comparison, sample-level margins could support failure analysis within individual models. A model that appears robust on average may still be unsuitable for deployment if its failures concentrate in clinically important subsets. Low-margin samples can be interrogated for enrichment of particular biological subclasses, institutions or patient characteristics, revealing structured failure modes concealed by an otherwise acceptable aggregate score. Identifying such subsets could guide targeted data collection, data augmentation and model pretraining, while motivating evaluation protocols that deliberately stress the conditions under which robustness is weakest.\\
\\
Several limitations remain. First, \code{CRoMa} requires known confounder labels: it measures the geometry induced by a labelled factor such as the acquisition centre, but cannot identify confounders that are not specified a priori. Second, our experiments treat the centre as the sole confounder. The construction accepts any labelled factor --- scanner, staining batch, fixation protocol --- but multi-factor and continuous confounders remain untested, and centre labels in practice bundle several of these sources into one. Finally, this study uses the margin distribution to compare models. The proposed diagnostic use for a single model remains to be demonstrated, and is the most direct extension of this work.\\
\\
Taken together, these findings support a more stringent standard for robustness evaluation in pathology foundation models: one that asks not only whether a model is robust on average, but where, and by how much, confounders override biological similarity. Making this boundary explicit links representation geometry to shortcut susceptibility and directs validation towards the samples and settings in which transfer is most likely to fail. More broadly, detecting such vulnerabilities before downstream adaptation is an important step towards building foundation models that generalise reliably across clinical settings.

\section{Methods}

We consider the Robustness Index (\code{RI}) from prior work, a distance-weighted control built on it (\code{MaRI}), and the Cross-confounder Robustness Margin (\code{CRoMa}) introduced in this work. All three metrics are built from the same typed-neighbour evidence. For each anchor sample, they compare cross-confounder biological matches, which share the anchor’s biological label but differ in confounder, with same-confounder biological distractors, which share the anchor’s confounder but differ in biological label. We first define this common setup, then show how each metric aggregates the resulting evidence: \code{RI} as a pooled count of typed neighbours, \code{MaRI} as a pooled distance-weighted analogue, and \code{CRoMa} as a per-sample signed margin between the two neighbour types. Figure~\ref{fig:concept-metrics} illustrates the three metrics on a single worked neighbourhood.

\subsection{Setup and notation}

Each sample $i$ is represented by a feature vector $z_i$ extracted using a foundation model, a biological label $y_i$, a confounder label $c_i$ (for example, its acquisition centre), and a grouping identifier $g_i$ denoting the physical unit from which the sample was derived (a whole-slide image for tile-level samples, a case for slide-level samples). We refer to the sample being scored as the \emph{anchor}. Features are L2-normalised, and candidate neighbours are ranked by cosine distance $d_{ij}=1-\cos(z_i,z_j)$. Before scoring, samples from the same physical unit as the anchor ($g_j=g_i$) are excluded from its candidate neighbourhood, preventing any metric from benefiting from near-duplicate samples, such as tiles from the same slide.\\
\\
For an anchor $i$ and neighbour $j$, the pair is \emph{typed} according to whether biology and confounder match, following the four-way scheme introduced by~\cite{dejong2025unrobust} and adopted in PathoROB~\cite{pathorob}. Two pair types provide the relevant contrast. \textbf{SO} pairs have the \textbf{S}ame biological label and an \textbf{O}ther confounder ($y_j=y_i, c_j\neq c_i$). These are cross-confounder biological matches: their proximity indicates biology-organised geometry. \textbf{OS} pairs have \textbf{O}ther biology and the \textbf{S}ame confounder ($y_j\neq y_i, c_j=c_i$). These are same-confounder distractors: their proximity indicates confounder-organised geometry. The remaining pair types are uninformative and are therefore excluded from scoring. For both \textbf{SS} and \textbf{OO} neighbours, their proximity to the anchor cannot be attributed specifically to either biological or confounder signal: \textbf{SS} pairs share both factors, whereas \textbf{OO} pairs share neither. Accordingly, all three metrics compare only \textbf{SO} and \textbf{OS} neighbours.

\subsection{Robustness Index}

We follow the PathoROB definition of the Robustness Index (\code{RI}) \cite{pathorob}. For a fixed neighbourhood size $k$, let $\mathcal{N}_k(i)$ denote the $k$ nearest neighbours of anchor $i$, after excluding samples from the same physical unit. Within this neighbourhood, \code{RI} counts the two informative neighbour types:

\begin{align}
	\mcode{SO}_i &= \sum_{j\in\mathcal{N}_k(i)}\mathbf{1}[y_j=y_i \land c_j\neq c_i] \\
	\mcode{OS}_i &= \sum_{j\in\mathcal{N}_k(i)}\mathbf{1}[y_j\neq y_i \land c_j=c_i]
\end{align}

\noindent
A model is then summarised by the pooled proportion of informative neighbours that are cross-confounder biological matches:

\begin{equation}
	\mcode{RI}=\frac{\sum_i \mcode{SO}_i}{\sum_i \left(\mcode{SO}_i+\mcode{OS}_i\right)}
\end{equation}

\noindent
The score ranges from $0$, indicating neighbourhoods dominated by same-confounder distractors, to $1$, indicating neighbourhoods dominated by cross-confounder biological matches. Unless stated otherwise, and to stay consistent with the original PathoROB protocol, we adopt its two-step procedure for selecting $k$. First, for each model, we choose the value of $k$ that maximises biological $k$-nearest-neighbour balanced accuracy, sweeping PathoROB's candidate grid ($k\in\{1,3,5,7,9\}$, then steps of $10$ up to $91$). We then evaluate all models using the median of these model-specific optimal values. This choice provides a common evaluation scale, although no single $k$ is both model-specific and directly comparable across models. The slide-level panel is an explicit exception and is evaluated at per-model $k^\star$, as detailed below and in Supplementary Section~\ref{supp:panda}.

\subsection{A distance-weighted control}

Because \code{RI} is count-based, it treats all neighbours within the fixed-$k$ set equally, regardless of their distance from the anchor. Two models can therefore obtain identical \code{RI} scores while inducing opposite local geometries: in one, cross-confounder biological matches lie much closer to the anchor than same-confounder distractors, whereas in the other, the distractors lie closer than the biological matches (Figure~\ref{fig:concept-ri-mari}). To test whether this insensitivity to distance is \code{RI}'s main deficiency, we construct the Margin-aware Robustness Index (\code{MaRI}) as a minimal distance-weighted control. \code{MaRI} preserves the same fixed-$k$, proportion-based formulation but weights each typed neighbour by its distance to the anchor:

\begin{equation}
	w_{ij}=\exp\!\left(-\frac{d_{ij}}{\tau}\right)
\end{equation}

\noindent
where $\tau>0$ is a temperature parameter. The weighted typed evidence becomes:

\begin{align}
	\mcode{SO}_{w}(i) &= \sum_{j\in\mathcal{N}_k(i)} w_{ij}\,\mathbf{1}[y_j=y_i \land c_j\neq c_i] \\
	\mcode{OS}_{w}(i) &= \sum_{j\in\mathcal{N}_k(i)} w_{ij}\,\mathbf{1}[y_j\neq y_i \land c_j=c_i]
\end{align}

\noindent
resulting in the pooled index:

\begin{equation}
	\mcode{MaRI}=\frac{\sum_i \mcode{SO}_{w}(i)}{\sum_i \left(\mcode{SO}_{w}(i)+\mcode{OS}_{w}(i)\right)}
\end{equation}

\noindent
Smaller values of $\tau$ concentrate the score on the closest typed neighbours. As $\tau$ grows the weights flatten and \code{MaRI} approaches \code{RI}'s unweighted counts. To keep the temperature on the scale of the distances being weighted, we set $\tau$ separately for each model to the median distance among typed \textbf{SO}/\textbf{OS} neighbours.\\
\\
Distance weighting changes how typed neighbours contribute to the score, but not which anchors contribute: like \code{RI}, \code{MaRI} includes an anchor only if its fixed-$k$ neighbourhood contains at least one typed neighbour ($\mcode{SO}_i+\mcode{OS}_i>0$; Figure~\ref{fig:concept-undefined}). We call the fraction of anchors satisfying this condition a model's \emph{support}. Because support varies across models---a direct consequence of how $k$ is selected (Supplementary Section~\ref{supp:k-selection})---pooled \code{RI} and \code{MaRI} summarise model-dependent subsets of the evaluation cohort.

\subsection{Cross-confounder Robustness Margin}
\label{sec:methods-croma}

\code{CRoMa} keeps the distance sensitivity introduced by \code{MaRI} while assigning a score to every sample. Rather than requiring informative neighbours to appear within a pre-specified $k$-neighbourhood, it locates the nearest informative neighbours of each type and measures their distance separation directly (Figure~\ref{fig:concept-croma}). For each sample $i$, let $d^{\mcode{SO}}_m(i)$ and $d^{\mcode{OS}}_m(i)$ denote the mean cosine distances to its $m$ nearest \textbf{SO} and $m$ nearest \textbf{OS} neighbours:

\begin{align}
	d^{\mcode{SO}}_m(i) &= \frac{1}{m}\sum_{j=1}^{m} d_{i,\,\sigma^{\mcode{SO}}_j(i)} \\
	d^{\mcode{OS}}_m(i) &= \frac{1}{m}\sum_{j=1}^{m} d_{i,\,\sigma^{\mcode{OS}}_j(i)}
\end{align}

\noindent
where $\sigma^{\mcode{SO}}_j(i)$ and $\sigma^{\mcode{OS}}_j(i)$ index the $j$-th nearest \textbf{SO} and \textbf{OS} neighbours, respectively.\\
\\
The per-sample robustness margin is defined as:

\begin{equation}
	\mcode{CRoMa}_m(i) = \frac{d^{\mcode{OS}}_m(i)-d^{\mcode{SO}}_m(i)}{d^{\mcode{OS}}_m(i)+d^{\mcode{SO}}_m(i)}
\end{equation}

\vspace{3mm}

\noindent
This yields a signed, scale-free margin in $(-1,1)$. Positive values indicate that same-confounder biological distractors are farther from the anchor than cross-confounder biological matches, consistent with a biology-dominant local geometry. Negative values indicate a confounder-dominant geometry. Averaging over $m$ neighbours per type reduces sensitivity to single-neighbour outliers. In contrast to the neighbourhood size $k$ used by \code{RI} and \code{MaRI}, which can vary between models and benchmarks, $m$ is fixed a priori and used unchanged across all models and datasets. We set $m=5$, ensuring that no individual neighbour contributes more than one fifth of a typed mean. Our conclusions are stable across a sweep of $m$ values (Supplementary Section~\ref{supp:m-sweep}).\\
\\
Because \code{CRoMa} depends on the distance ratio $r_i = d^{\mcode{OS}}_m(i)/d^{\mcode{SO}}_m(i)$ rather than on absolute distances\footnote{$\mcode{CRoMa}_m(i)=(r_i-1)/(r_i+1)$}, it is invariant to any global multiplicative rescaling of distances within a model. Thus, a model is not favoured simply because its embedding distances are uniformly larger or smaller: only the relative separation between cross-confounder biological matches and same-confounder biological distractors matters. This gives \code{CRoMa} a common interpretation across models and datasets, while retaining a direct geometric meaning. For example, a median \code{CRoMa} of $0.7$ implies that, for the typical anchor, the nearest same-confounder biological distractors are approximately $5.7\times$ farther away than the nearest cross-confounder biological matches\footnote{At $\mcode{CRoMa}_m(i)=0.7$, $r_i=(1+\mcode{CRoMa}_m(i))/(1-\mcode{CRoMa}_m(i))=1.7/0.3\approx5.7$}. A complete geometric interpretation is provided in Supplementary Section~\ref{supp:geometry}.\\
\\
By construction, \code{CRoMa} assigns a score to every sample, provided that at least $m$ \mcode{SO} and $m$ \mcode{OS} neighbours exist in the evaluation set, a condition satisfied by all benchmarks used in this study for $m=5$. The resulting margins ${\mcode{CRoMa}_m(i)}$ therefore define an empirical robustness distribution over the full evaluation cohort, rather than over the model-dependent support set on which \code{RI} and \code{MaRI} are effectively computed. This distribution is the primary robustness readout, because it preserves sample-level heterogeneity that a pooled score necessarily compresses. For compact model comparison, we summarise it along three complementary axes: central tendency, the prevalence of non-positive margins, and the severity of robustness failures.\\
\\
As a robust measure of central tendency, we report the median robustness margin:

\begin{equation}
	\mcode{CRoMa}=\mathrm{median}_i \,\mcode{CRoMa}_m(i)
\end{equation}

\noindent
The sign of the median indicates whether the typical sample occupies a biology-dominant or confounder-dominant local geometry. The median alone does not describe the lower end of the distribution, which we characterise along two axes: \emph{prevalence} and \emph{severity}.\\
\\
Prevalence is measured at the decision boundary. Let $F$ denote the empirical cumulative distribution function of ${\mcode{CRoMa}_m(i)}$. We report:

\begin{equation}
	F(0)=\mathbb{P}\left[\mcode{CRoMa}_m(i)\leq 0\right],
\end{equation}

\noindent
Since zero is the boundary between biology-dominant and confounder-dominant geometry, $F(0)$ is the fraction of samples for which biology does not dominate locally: negative margins indicate confounder-dominant geometry, whereas zero denotes a tie. Severity is measured in the lower tail of the margin distribution. Let $Q_{\alpha}$ denote the $\alpha$-quantile of ${\mcode{CRoMa}_m(i)}$. We define the lower-tail mean as:

\begin{equation}
	\mcode{LTM}_{\alpha}=\mathbb{E}\!\left[\mcode{CRoMa}_m(i) \mid \mcode{CRoMa}_m(i) \le Q_{\alpha}\right],
\end{equation}

\noindent
$\mcode{LTM}_{\alpha}$ measures the average robustness margin among the worst $\alpha$ fraction of samples. We report $\mcode{LTM}_{\alpha}$ at $\alpha=0.10$, corresponding to the mean of the worst decile. Our conclusions are stable across $\alpha\in\{0.05,0.10,0.20\}$ (Supplementary Section~\ref{supp:alpha-sweep}).

\subsection{Downstream shortcut susceptibility}
\label{sec:methods-nipd}

Having defined representation-level robustness, we next asked whether it anticipates shortcut learning after supervised adaptation. For the tile-level benchmarks, we reproduced the confounder-biased linear-probe protocol introduced in PathoROB~\cite{pathorob}. Linear probes predicted the biological label from frozen embeddings across training splits in which the confounder--biology correlation was progressively increased from Cramér's $V=0$ (independence) to $V=1$ (complete dependence). Total training-set size and the marginal frequencies of biological classes and confounder groups were held fixed, only their joint composition changed. Each probe was evaluated on balanced in-domain (ID) data from acquisition groups represented during training and out-of-domain (OOD) data from held-out groups. ID performance provides the primary measure of shortcut susceptibility, whereas OOD performance additionally reflects transfer to an external acquisition domain.\\
\\
Let $a^{d}(V_s)$ denote balanced accuracy in evaluation setting $d\in\{\mathrm{ID},\mathrm{OOD}\}$ at Cramér's $V_s$, with $V_0=0$ and $V_S=1$. PathoROB summarises the resulting trajectory using average performance drop (\code{APD}):

\begin{equation}
    \mcode{APD}^{d}
    =
    \frac{1}{S}\sum_{s=1}^{S}
    \left(\frac{a^{d}(V_s)}{a^{d}(V_0)}-1\right)
\end{equation}

\noindent
However, \code{APD} can treat failures of very different severity as equivalent. In a binary task, a decline from $0.60$ to chance ($0.50$) and a decline from $0.95$ to $0.792$ both give $\code{APD}\approx-16.7\%$. Yet the first removes all performance above chance, whereas the second removes only $35.2\%$. \code{APD} obscures this distinction because it normalises by total baseline performance, including the chance-level component that cannot be lost. We therefore introduce normalised integrated performance degradation (\code{nIPD}). Denoting chance balanced accuracy by $\pi$, the normalised performance-change curve of a model is:

\begin{equation}
    g^{d}(V_s)
    =
    \frac{a^{d}(V_s)-a^{d}(V_0)}
         {a^{d}(V_0)-\pi}
\end{equation}

\noindent
For the two examples above, $g(1)=-1.00$ and $-0.352$, respectively, directly distinguishing complete from partial loss of above-chance performance. \code{nIPD} summarises this curve across the full confounding range as its signed area:

\begin{equation}
    \mcode{nIPD}^{d}
    =
    \int_{0}^{1} g^{d}(V)\,\mathrm{d}V
\end{equation}

\noindent
Because $g(V)$ is observed only at the sampled Cramér's-$V$ values, we approximate this integral using the trapezoidal rule:

\begin{equation}
    \mcode{nIPD}^{d}
    \approx
    \sum_{s=1}^{S}
    (V_s-V_{s-1})
    \frac{g^{d}(V_{s-1})+g^{d}(V_s)}{2}
\end{equation}

\noindent
Values near zero indicate stable performance, whereas more negative values indicate greater loss relative to the above-chance performance available at $V=0$. We use \code{nIPD} as the primary reduction and retain \code{APD} for comparison with PathoROB.

\subsection{Benchmarks}

\begin{figure}[!htbp]
\centering
\includegraphics[width=\linewidth]{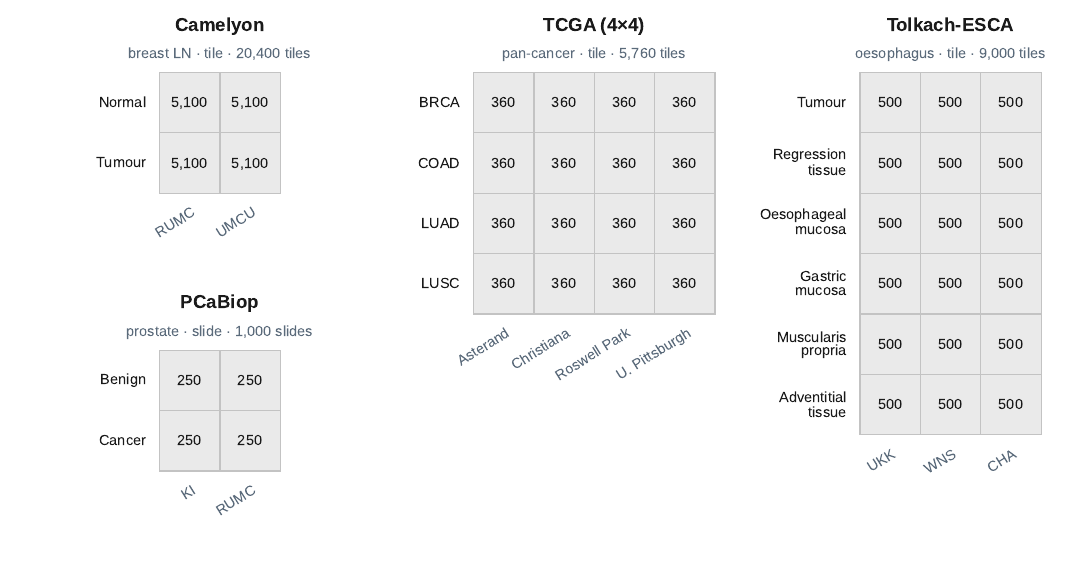}
\caption{\textbf{Each benchmark is balanced across biological class and confounder.} Cells show evaluated samples per biological-class$\times$confounder combination: $5{,}100$ tiles for Camelyon ($20{,}400$ total), $360$ for TCGA-4x4 ($5{,}760$), $500$ for Tolkach-ESCA ($9{,}000$) and $250$ slides for PCaBiop ($1{,}000$). Rows denote biological classes and columns acquisition centres. The paired TCGA-2x2 design is shown in Supplementary Figure~\ref{fig:tcga2x2-cardinality}.}
\label{fig:dataset-cardinality}
\end{figure}

We evaluate all metrics on benchmarks that pair a biological label with the acquisition site that produced the sample (Table~\ref{tab:dataset-summary}; per-cell cardinalities in Figure~\ref{fig:dataset-cardinality}). Three tile-level benchmarks are taken directly from PathoROB~\cite{pathorob}: Camelyon, comprising breast lymph-node tumour and normal tissue across two centres, TCGA-4x4, comprising four cancer types from four centres, and Tolkach-ESCA, comprising six oesophageal tissue classes across three cohorts. We refer to the original PathoROB study for detailed dataset descriptions. To assess whether the same biology--confounder tension extends beyond tile-level encoders, we additionally curate PCaBiop, a slide-level benchmark of H\&E prostate biopsies sourced from PANDA~\cite{panda}. Together, these benchmarks span multiple representation scales, organs, biological label structures, and confounding regimes. Representative samples from each benchmark are shown in Supplementary Figure~\ref{fig:dataset-examples}, illustrating the non-biological variation between centres.

\begin{table}[!htbp]
\centering
\small
\setlength{\tabcolsep}{5pt}
\begin{tabular}{lllll}
\hline
Benchmark & Level & Biological label (\#cls) & Confounder (\#) & Samples \\
\hline
Camelyon~\cite{pathorob}          & tile  & breast LN: tumour/normal (2) & medical centre (2) & $20{,}400$ tiles \\
TCGA-4x4~\cite{pathorob}  & tile  & cancer type (4)              & medical centre (4) & $5{,}760$ tiles  \\
Tolkach-ESCA~\cite{pathorob}       & tile  & oesophageal tissue (6)       & medical centre (3)         & $9{,}000$ tiles  \\
PCaBiop              & slide & prostate: benign/cancer (2)  & medical centre (2)  & $1{,}000$ slides \\
\hline
\end{tabular}
\caption{\textbf{Evaluation benchmarks span multiple organs and representation scales.} The four cohorts cover multiple organs and biological conditions, as well as tile- and slide-level representations. Confounders are non-biological acquisition artefacts from the medical centre that contributed the sample. Camelyon, TCGA-4x4 and Tolkach-ESCA are taken from PathoROB~\cite{pathorob}. PCaBiop is a collection of prostate biopsies sourced from PANDA~\cite{panda}.}
\label{tab:dataset-summary}
\end{table}

\subsection{Models}

All models are evaluated as frozen feature extractors, so the metrics probe the representations exposed after pretraining rather than any task-specific adaptation (Table~\ref{tab:model-summary}). For the tile-level benchmarks, we evaluate a common panel of \TileRankedNModels{} tile encoders spanning a broad range of architecture sizes, pretraining corpus scales, and self-supervised objectives, alongside a natural-image control (\code{DINOv2-B}) reported separately throughout. For the slide-level benchmark, we evaluate \SlideNModels{} whole-slide encoders that aggregate tile-level information into a single slide embedding. Because robustness estimates may be affected by overlap between a benchmark and a model's pretraining domain, we record each model's disclosed pretraining corpus, to the best of our ability, in Table~\ref{tab:model-summary}.

\begin{table}[!htbp]
\centering
\footnotesize
\setlength{\tabcolsep}{2pt} % tightened to absorb the per-model citations
\begin{tabular}{llcccl}
\hline
Model & Method & Params & WSIs / tiles & Dim & Pretraining corpus \\
\hline
\multicolumn{6}{l}{\emph{Tile-level encoders}} \\
\hline
Virchow2~\cite{zimmermann2024} & DINOv2 & $632$M & $3.1$M / $1.9$B & $2560$ & MSKCC (prop.) \\
Virchow~\cite{Vorontsov2024-te} & DINOv2 & $632$M & $1.5$M / $2$B & $2560$ & MSKCC (prop.) \\
UNI2-h~\cite{uni2h} & DINOv2 & $681$M & ${>}350$k / ${>}200$M & $1536$ & MGB (prop.) \\
UNI~\cite{Chen2024-dy} & DINOv2 & $307$M & $100$k / $100$M & $1024$ & MGB (prop.) \\
CONCHv1.5~\cite{titan} & vision--language & $307$M & $100$k / $100$M & $768$ & PMC-OA + educational \\
CONCH~\cite{conch} & vision--language & $86$M & $21.4$k / $16$M & $512$ & PMC-OA + educational \\
H-optimus-1~\cite{hoptimus1} & n/d & $1.1$B & ${>}1$M / n/d & $1536$ & Bioptimus (prop.) \\
H-optimus-0~\cite{hoptimus0} & DINOv2 & $1.1$B & $500$K / $273$M & $1536$ & Bioptimus (prop.) \\
H0-mini~\cite{filiot2025plism} & distilled (H-opt-0) & $86$M & $6{,}093$ / $43$M & $1536$ & \textbf{TCGA (distilled)} \\
Prov-GigaPath~\cite{Xu2024-gj} & DINOv2 & $1.1$B & $171$k / $1.3$B & $1536$ & Providence (prop.) \\
Midnight-12k~\cite{midnight} & DINOv2 (var.) & $1.1$B & $12$k / $384$M & $3072$ & \textbf{TCGA (public)} \\
Prost40M~\cite{grisi2026bcr} & DINO & $22$M & $2$k / ${\sim}40$M & $384$ & \textbf{TCGA-PRAD + LEOPARD} \\
Phikon~\cite{phikon} & iBOT & $86$M & $6$k / $43.3$M & $768$ & \textbf{TCGA (public)} \\
Phikon-v2~\cite{phikonv2} & DINOv2 & $307$M & ${\sim}58$k / ${\sim}456$M & $1024$ & \textbf{PanCancer-XL (incl.\ TCGA)} \\
Hibou-L~\cite{hibou} & DINOv2 & $307$M & ${>}1$M / $1.2$B & $1024$ & HistAI (prop.) \\
Hibou-B~\cite{hibou} & DINOv2 & $86$M & ${>}1$M / $512$M & $768$ & HistAI (prop.) \\
mSTAR~\cite{mstar} & distilled (UNI) & $307$M & $11{,}765$ / ${>}116$M & $1024$ & \textbf{TCGA (self-taught)} \\
GPFM~\cite{gpfm} & distilled (multi-expert) & $307$M & $72{,}280$ / $190$M & $1024$ & \textbf{33 public (TCGA, CAMELYON)} \\
MUSK~\cite{musk} & vision--language & $307$M & ${\sim}33$k / $50$M & $2048$ & \textbf{TCGA $+$ PMC-OA $+$ Quilt-1M} \\
GenBio-PathFM~\cite{genbiopathfm} & JEDI (DINO$+$JEPA) & $1.13$B & ${\sim}177$k / ${\sim}400$M & $4608$ & \textbf{HistAI $+$ TCGA $+$ GTEx $+$ REG} \\
DINOv2-B~\cite{oquab2024} & DINOv2 & $86$M & n/a & $768$ & LVD-142M (natural images) \\
\hline
\multicolumn{6}{l}{\emph{Slide-level encoders}} \\
\hline
PRISM~\cite{prism} & Perceiver & $99$M$^{\ast}$ & $587$k / n/d & $1280$ & Paige (prop.) \\
TITAN~\cite{titan} & CoCa & $48.5$M & $336$k / n/d & $768$ & MGB (prop.) \\
Prov-GigaPath~\cite{Xu2024-gj} & LongNet & $86.3$M$^{\ast}$ & $171$k / $1.3$B & $768$ & Providence (prop.) \\
MOOZY~\cite{moozy} & grid ViT & $85.8$M$^{\dagger}$ & $77$k$^{\dagger}$ / $1.7$B & $768$ & \textbf{56 public (TCGA, PANDA)} \\
\hline
\end{tabular}
\caption{\textbf{Foundation models evaluated in this study.} Tile- and slide-level encoders are summarised by model architecture or training method, parameter count, disclosed pretraining scale in whole-slide images (WSIs) and tiles, embedding dimension and pretraining corpus. \code{DINOv2-B} is included as a natural-image control. Bold indicates that the disclosed pretraining corpus includes a dataset contributing to at least one of the four evaluation benchmarks. Dim, embedding dimension; prop., proprietary; n/d, not disclosed; n/a, not applicable; k, thousand; M, million; B, billion. For \code{CONCHv1.5}, the WSI and tile counts refer to the UNI-initialised vision trunk; subsequent vision--language training used $1.26$M image--caption pairs. For \code{Midnight-12k}, $384$M denotes online tile draws rather than unique tiles. $^{\ast}$ Slide-encoder parameter count reported by Ding et al.~\cite{titan}. $^{\dagger}$ For \code{MOOZY}, $77$k denotes multiscale slide feature grids rather than distinct WSIs, and $85.8$M includes the frozen patch encoder.}
\label{tab:model-summary}
\end{table}

\subsection{Evaluation protocol}

On the tile-level benchmarks, \code{RI}, \code{MaRI} and both $k$-NN probes use the median-$k$ protocol described above. Because the slide-level panel contains only four encoders, the shared median collapses to $k=3$ and reduces mean support from $36.9\%$ to $27.0\%$, so slide-level models are instead evaluated at each encoder's biologically selected $k^\star$ (Supplementary Section~\ref{supp:panda}). \code{CRoMa} and its distributional summaries are $k$-independent and unaffected. \code{MaRI} additionally uses a per-model temperature $\tau$, and \code{CRoMa} uses $m=5$ neighbours per type for all models and benchmarks. For TCGA, PathoROB provides two native evaluation configurations. We focus on the TCGA-4x4 configuration -- four cancer types (BRCA, COAD, LUAD, LUSC) across four medical centres. For completeness, we additionally report results on PathoROB's paired TCGA-2x2 benchmark (Supplementary Section~\ref{supp:tcga}). Alongside \code{RI} and \code{MaRI}, we report each model's support. \code{CRoMa} is defined on the full evaluation cohort by construction.\\
\\
Each results table additionally reports two decodability probes computed directly on the frozen embeddings (bio bacc and conf bacc). Both are leave-one-out $k$-NN classifiers over the evaluation cohort: each sample is assigned the majority label among its $k$ nearest neighbours and performance is summarised as balanced accuracy. The biological probe predicts the biological label, the confounder probe predicts the confounder label, and both are evaluated at the same operating $k$ as \code{RI} and \code{MaRI}.\\
\\
The downstream tile-level analyses used the original PathoROB split schedules. For PCaBiop, the ID pool comprised $1{,}000$ prostate biopsy slides from PANDA, balanced across benign and cancer labels and the Radboud and Karolinska centres ($250$ slides per label--centre cell). The OOD pool comprised $324$ prostate biopsy slides from PAR~\cite{par}, balanced across benign and cancer labels. We sampled $11$ training compositions at $V\in\{0,0.1,\ldots,1\}$ while holding the training--validation pool at $400$ slides, with $10\%$ used for validation. Table~\ref{tab:downstream-rowset} summarises the samples entering the representation metrics and the downstream probes on each benchmark.

\begin{table}[!htbp]
\centering
\small
\setlength{\tabcolsep}{6pt}
\begin{tabular}{lcccc}
\hline
                       & Representation metrics
                       & \multicolumn{3}{c}{Downstream probe} \\
\cline{3-5}
Benchmark              & (\code{RI}/\code{MaRI}/\code{CRoMa})
                       & In domain      & Out of domain  & Full pool      \\
\hline
Camelyon               & $20{,}400$ (2) & $20{,}400$ (2) & $2{,}002$ (3)  & $22{,}402$ (5) \\
TCGA-4x4               & $5{,}760$ (4)  & $5{,}760$ (4)  & $2{,}400$ (4)  & $8{,}160$ (8)  \\
Tolkach-ESCA           & $9{,}000$ (3)  & $10{,}800$ (2) & $5{,}500$ (2)  & $16{,}300$ (4) \\
PCaBiop      & $1{,}000$ (2)  & $1{,}000$ (2)  & $324$ (1)      & $1{,}324$ (3)  \\
\hline
\end{tabular}
\caption{\textbf{Representation metrics and downstream probes use different evaluation samples.} Tile counts for the tile-level benchmarks and slide counts for PCaBiop. Parentheses give the number of contributing acquisition centres. \code{RI}, \code{MaRI} and \code{CRoMa} use each benchmark's representation-evaluation cohort. The downstream probe uses both an ID cohort and held-out OOD centres or cohorts. For Camelyon and TCGA-4x4, the representation and ID samples are identical. Tolkach-ESCA uses distinct representation and probe cohorts. The slide-level analysis uses $1{,}000$ prostate biopsy slides from PANDA, originating from two centres, and $324$ prostate biopsies from PAR, used as an external OOD cohort.}
\label{tab:downstream-rowset}
\end{table}

\subsection{Statistics}

Representation-level summaries (median, $F(0)$, $\mcode{LTM}_{10}$) are descriptive statistics over the evaluation cohort. Spearman rank correlations were computed across the 20 pathology tile encoders, with the natural-image control excluded, or across the four slide encoders for PCaBiop.\\
\\
Downstream probe trajectories were estimated over 20 resampling runs. In each run, the regularisation parameter of the logistic-regression probe was selected by validation balanced accuracy from 15 values logarithmically spaced between $10^{-8}$ and $10^{4}$. Performance was evaluated using balanced accuracy. At each Cramér's-$V$ value, accuracies were averaged across runs before calculating $g(V)$ and integrating \code{nIPD} by the trapezoidal rule. Figure error bars show two-sided 95\% Student's $t$ confidence intervals for the mean paired normalised change from $V=0$ across the 20 runs.

\section*{Data availability}
All benchmarks derive from publicly available data. The Camelyon, TCGA-4x4 and Tolkach-ESCA cohorts follow the PathoROB benchmark protocol~\cite{pathorob}. The slide-level benchmark uses prostate biopsy slides from the PANDA challenge~\cite{panda}, with the PAR cohort~\cite{par} as an external out-of-domain set, and is available at \url{https://huggingface.co/datasets/waticlems/pcabiop}.

\section*{Code availability}
Metric implementations and evaluation manifests are released as an open-source Python package, \code{croma}, available at \url{https://github.com/clemsgrs/croma}.

\bibliographystyle{naturemag} % Nature bibliography style (naturemag.bst, CTAN macros/latex/contrib/nature)
\bibliography{references}

@article{panda,
      title={Artificial intelligence for diagnosis and {Gleason} grading of prostate cancer: the {PANDA} challenge},
      author={Bulten, Wouter and Kartasalo, Kimmo and Chen, Po-Hsuan Cameron and others},
      journal={Nature Medicine},
      volume={28},
      number={1},
      pages={154--163},
      year={2022},
      doi={10.1038/s41591-021-01620-2},
}

@misc{grisi2026bcr,
      title = {Deep learning from routine histology improves risk stratification for biochemical recurrence in prostate cancer},
      author = {Grisi, Cl{\'e}ment and Faryna, Khrystyna and Uysal, Nefise and Agosti, Vittorio and Munari, Enrico and Kammerer-Jacquet, Sol{\`e}ne-Florence and de Oliveira Salles, Paulo Guilherme and Tolkach, Yuri and B{\"u}ttner, Reinhard and Semko, Sofiya and Pikul, Maksym and Heidenreich, Axel and van der Laak, Jeroen and Litjens, Geert},
      year = {2026},
      eprint = {2603.14187},
      note={arXiv:2603.14187},
      archivePrefix = {arXiv},
      primaryClass = {cs.CV},
      url = {https://arxiv.org/abs/2603.14187},
}

@article{conch,
      title={A visual-language foundation model for computational pathology},
      author={Lu, Ming Y. and Chen, Bowen and Williamson, Drew F. K. and Chen, Richard J. and Liang, Ivy and Ding, Tong and Jaume, Guillaume and Odintsov, Igor and Le, Long Phi and Gerber, Georg and Parwani, Anil V. and Zhang, Andrew and Mahmood, Faisal},
      journal={Nature Medicine},
      volume={30},
      number={3},
      pages={863--874},
      year={2024},
      doi={10.1038/s41591-024-02856-4},
}

@misc{midnight,
      title={Training state-of-the-art pathology foundation models with orders of magnitude less data},
      author={Mikhail Karasikov and Joost van Doorn and Nicolas K{\"a}nzig and Melis Erdal Cesur and Hugo Mark Horlings and Robert Berke and Fei Tang and Sebastian Otálora},
      year={2025},
      eprint={2504.05186},
      note={arXiv:2504.05186},
      archivePrefix={arXiv},
      primaryClass={cs.CV},
      url={https://arxiv.org/abs/2504.05186},
}

@misc{moozy,
      title={{MOOZY}: A Patient-First Foundation Model for Computational Pathology},
      author={Yousef Kotp and Vincent Quoc-Huy Trinh and Christopher Pal and Mahdi S. Hosseini},
      year={2026},
      eprint={2603.27048},
      note={arXiv:2603.27048},
      archivePrefix={arXiv},
      primaryClass={cs.CV},
      url={https://arxiv.org/abs/2603.27048},
}

@misc{zimmermann2024,
	title={Virchow2: Scaling Self-Supervised Mixed Magnification Models in Pathology}, 
	author={Eric Zimmermann and Eugene Vorontsov and Julian Viret and Adam Casson and Michal Zelechowski and George Shaikovski and Neil Tenenholtz and James Hall and David Klimstra and Razik Yousfi and Thomas Fuchs and Nicolo Fusi and Siqi Liu and Kristen Severson},
	year={2024},
	eprint={2408.00738},
	note={arXiv:2408.00738},
	archivePrefix={arXiv},
	primaryClass={cs.CV},
}

@misc{hoptimus0,
	author = {Saillard, Charlie and Jenatton, Rodolphe and Llinares-López, Felipe and Mariet, Zelda and Cahané, David and Durand, Eric and Vert, Jean-Philippe},
	title = {H-optimus-0},
	year = {2024},
	howpublished = {\url{https://github.com/bioptimus/releases/tree/main/models/h-optimus/v0}},
	url = {https://github.com/bioptimus/releases/tree/main/models/h-optimus/v0},
}

@article{hoptimus1,
	title = {Abstract {LB174}: {H-optimus-1}: a foundation model for computational histopathology},
	author = {Scalbert, Marin and Saillard, Charlie and Peeters, Thomas and Gonzalez, Liam and Valter, Dasha and Llinares-L{\'o}pez, Felipe and Mariet, Zelda E. and Jenatton, Rodolphe},
	journal = {Cancer Research},
	volume = {86},
	number = {8\_Supplement},
	pages = {LB174},
	year = {2026},
	publisher = {American Association for Cancer Research},
	doi = {10.1158/1538-7445.AM2026-LB174},
	note = {Model weights: \url{https://huggingface.co/bioptimus/H-optimus-1}},
}

@misc{uni2h,
	author = {{Mahmood Lab}},
	title = {{MahmoodLab/UNI2-h}},
	year = {2025},
	howpublished = {Hugging Face model repository, \url{https://huggingface.co/MahmoodLab/UNI2-h}},
	url = {https://huggingface.co/MahmoodLab/UNI2-h},
}

@article{pathorob,
	title={Towards robust foundation models for digital pathology},
	author={Jonah Kömen and Edwin D. de Jong and Julius Hense and Hannah Marienwald and Jonas Dippel and Philip Naumann and Eric Marcus and Lukas Ruff and Maximilian Alber and Jonas Teuwen and Frederick Klauschen and Klaus-Robert Müller},
	journal={Nature Communications},
	volume={17},
	number={1},
	pages={5218},
	year={2026},
	doi={10.1038/s41467-026-73923-2},
	url={https://doi.org/10.1038/s41467-026-73923-2},
}

@article{campanella2025clinical,
	title={A clinical benchmark of public self-supervised pathology foundation models},
	author={Campanella, Gabriele and Chen, Shengjia and Singh, Manbir and Verma, Ruchika and Muehlstedt, Silke and Zeng, Jennifer and Stock, Aryeh and Croken, Matt and Veremis, Brandon and Elmas, Abdulkadir and others},
	journal={Nature Communications},
	volume={16},
	number={1},
	pages={3640},
	year={2025},
	publisher={Nature Publishing Group UK London}
}

@misc{kömen2024batcheffects,
	title={Do Histopathological Foundation Models Eliminate Batch Effects? A Comparative Study}, 
	author={Jonah Kömen and Hannah Marienwald and Jonas Dippel and Julius Hense},
	year={2024},
	eprint={2411.05489},
	note={arXiv:2411.05489},
	archivePrefix={arXiv},
	primaryClass={cs.LG},
	url={https://arxiv.org/abs/2411.05489}, 
}

@misc{dejong2025unrobust,
	title={Current Pathology Foundation Models are unrobust to Medical Center Differences}, 
	author={Edwin D. de Jong and Eric Marcus and Jonas Teuwen},
	year={2025},
	eprint={2501.18055},
	note={arXiv:2501.18055},
	archivePrefix={arXiv},
	primaryClass={cs.LG},
	url={https://arxiv.org/abs/2501.18055}, 
}

@inproceedings{filiot2025plism,
	title={Distilling foundation models for robust and efficient models in digital pathology},
	author={Filiot, Alexandre and Dop, Nicolas and Tchita, Oussama and Riou, Auriane and Dubois, R{\'e}my and Peeters, Thomas and Valter, Daria and Scalbert, Marin and Saillard, Charlie and Robin, Genevi{\`e}ve and Olivier, Antoine},
	booktitle={Medical Image Computing and Computer Assisted Intervention -- {MICCAI} 2025},
	series={Lecture Notes in Computer Science},
	volume={15966},
	pages={162--172},
	publisher={Springer Nature Switzerland},
	year={2025},
	doi={10.1007/978-3-032-04981-0_16},
}

@ARTICLE{Vorontsov2024-te,
	title     = "A foundation model for clinical-grade computational pathology
	and rare cancers detection",
	author    = "Vorontsov, Eugene and Bozkurt, Alican and Casson, Adam and
	Shaikovski, George and Zelechowski, Michal and Severson, Kristen
	and Zimmermann, Eric and Hall, James and Tenenholtz, Neil and
	Fusi, Nicolo and Yang, Ellen and Mathieu, Philippe and van Eck,
	Alexander and Lee, Donghun and Viret, Julian and Robert, Eric
	and Wang, Yi Kan and Kunz, Jeremy D and Lee, Matthew C H and
	Bernhard, Jan H and Godrich, Ran A and Oakley, Gerard and
	Millar, Ewan and Hanna, Matthew and Wen, Hannah and Retamero,
	Juan A and Moye, William A and Yousfi, Razik and Kanan,
	Christopher and Klimstra, David S and Rothrock, Brandon and Liu,
	Siqi and Fuchs, Thomas J",
	journal   = "Nat. Med.",
	publisher = "Springer Science and Business Media LLC",
	volume    =  30,
	number    =  10,
	pages     = "2924--2935",
	month     =  oct,
	year      =  2024,
	copyright = "https://creativecommons.org/licenses/by/4.0",
	language  = "en"
}

@ARTICLE{Chen2024-dy,
	title    = "Towards a general-purpose foundation model for computational
	pathology",
	author   = "Chen, Richard J and Ding, Tong and Lu, Ming Y and Williamson,
	Drew F K and Jaume, Guillaume and Song, Andrew H and Chen, Bowen
	and Zhang, Andrew and Shao, Daniel and Shaban, Muhammad and
	Williams, Mane and Oldenburg, Lukas and Weishaupt, Luca L and
	Wang, Judy J and Vaidya, Anurag and Le, Long Phi and Gerber,
	Georg and Sahai, Sharifa and Williams, Walt and Mahmood, Faisal",
	journal  = "Nat. Med.",
	volume   =  30,
	number   =  3,
	pages    = "850--862",
	month    =  mar,
	year     =  2024,
	language = "en"
}

@ARTICLE{Xu2024-gj,
	title     = "A whole-slide foundation model for digital pathology from
	real-world data",
	author    = "Xu, Hanwen and Usuyama, Naoto and Bagga, Jaspreet and Zhang,
	Sheng and Rao, Rajesh and Naumann, Tristan and Wong, Cliff and
	Gero, Zelalem and Gonz{\'a}lez, Javier and Gu, Yu and Xu, Yanbo
	and Wei, Mu and Wang, Wenhui and Ma, Shuming and Wei, Furu and
	Yang, Jianwei and Li, Chunyuan and Gao, Jianfeng and Rosemon,
	Jaylen and Bower, Tucker and Lee, Soohee and Weerasinghe,
	Roshanthi and Wright, Bill J and Robicsek, Ari and Piening,
	Brian and Bifulco, Carlo and Wang, Sheng and Poon, Hoifung",
	journal   = "Nature",
	publisher = "Springer Science and Business Media LLC",
	volume    =  630,
	number    =  8015,
	pages     = "181--188",
	month     =  jun,
	year      =  2024,
	copyright = "https://creativecommons.org/licenses/by/4.0",
	language  = "en"
}

@ARTICLE{Neidlinger2026-xw,
	title     = "Benchmarking foundation models as feature extractors for weakly
	supervised computational pathology",
	author    = "Neidlinger, Peter and El Nahhas, Omar S M and Muti, Hannah
	Sophie and Lenz, Tim and Hoffmeister, Michael and Brenner,
	Hermann and van Treeck, Marko and Langer, Rupert and Dislich,
	Bastian and Behrens, Hans Michael and R{\"o}cken, Christoph and
	Foersch, Sebastian and Truhn, Daniel and Marra, Antonio and
	Saldanha, Oliver Lester and Kather, Jakob Nikolas",
	journal   = "Nat. Biomed. Eng.",
	publisher = "Springer Science and Business Media LLC",
	volume    =  10,
	number    =  6,
	pages     = "1113--1123",
	month     =  jun,
	year      =  2026,
	copyright = "https://creativecommons.org/licenses/by/4.0",
	language  = "en"
}

@ARTICLE{Howard2021-wx,
	title     = "The impact of site-specific digital histology signatures on deep
	learning model accuracy and bias",
	author    = "Howard, Frederick M and Dolezal, James and Kochanny, Sara and
	Schulte, Jefree and Chen, Heather and Heij, Lara and Huo,
	Dezheng and Nanda, Rita and Olopade, Olufunmilayo I and Kather,
	Jakob N and Cipriani, Nicole and Grossman, Robert L and Pearson,
	Alexander T",
	journal   = "Nat. Commun.",
	publisher = "Springer Science and Business Media LLC",
	volume    =  12,
	number    =  1,
	pages     = "4423",
	month     =  jul,
	year      =  2021,
	copyright = "https://creativecommons.org/licenses/by/4.0",
	language  = "en"
}

@ARTICLE{Geirhos2020-wa,
	title     = "Shortcut learning in deep neural networks",
	author    = "Geirhos, Robert and Jacobsen, J{\"o}rn-Henrik and Michaelis,
	Claudio and Zemel, Richard and Brendel, Wieland and Bethge,
	Matthias and Wichmann, Felix A",
	journal   = "Nat. Mach. Intell.",
	publisher = "Springer Science and Business Media LLC",
	volume    =  2,
	number    =  11,
	pages     = "665--673",
	month     =  nov,
	year      =  2020,
	copyright = "https://www.springernature.com/gp/researchers/text-and-data-mining",
	language  = "en"
}

@ARTICLE{Dehkharghanian2023-oy,
	title    = "Biased data, biased {AI}: deep networks predict the acquisition
	site of {TCGA} images",
	author   = "Dehkharghanian, Taher and Bidgoli, Azam Asilian and Riasatian,
	Abtin and Mazaheri, Pooria and Campbell, Clinton J V and
	Pantanowitz, Liron and Tizhoosh, H R and Rahnamayan, Shahryar",
	journal  = "Diagn. Pathol.",
	volume   =  18,
	number   =  1,
	pages    = "67",
	month    =  may,
	year     =  2023,
	language = "en"
}

@article{mahmoodBenchmarkingCrisis2025a,
	title = {A benchmarking crisis in biomedical machine learning},
	volume = {31},
	issn = {1546-170X},
	doi = {10.1038/s41591-025-03637-3},
	language = {eng},
	number = {4},
	journal = {Nature Medicine},
	author = {Mahmood, Faisal},
	month = apr,
	year = {2025},
	pmid = {40200055},
	pages = {1060},
}

@misc{bommasani2022,
	title={On the Opportunities and Risks of Foundation Models}, 
	author={Rishi Bommasani and Drew A. Hudson and Ehsan Adeli and Russ Altman and Simran Arora and Sydney von Arx and Michael S. Bernstein and Jeannette Bohg and Antoine Bosselut and Emma Brunskill and Erik Brynjolfsson and Shyamal Buch and Dallas Card and Rodrigo Castellon and Niladri Chatterji and Annie Chen and Kathleen Creel and Jared Quincy Davis and Dora Demszky and Chris Donahue and Moussa Doumbouya and Esin Durmus and Stefano Ermon and John Etchemendy and Kawin Ethayarajh and Li Fei-Fei and Chelsea Finn and Trevor Gale and Lauren Gillespie and Karan Goel and Noah Goodman and Shelby Grossman and Neel Guha and Tatsunori Hashimoto and Peter Henderson and John Hewitt and Daniel E. Ho and Jenny Hong and Kyle Hsu and Jing Huang and Thomas Icard and Saahil Jain and Dan Jurafsky and Pratyusha Kalluri and Siddharth Karamcheti and Geoff Keeling and Fereshte Khani and Omar Khattab and Pang Wei Koh and Mark Krass and Ranjay Krishna and Rohith Kuditipudi and Ananya Kumar and Faisal Ladhak and Mina Lee and Tony Lee and Jure Leskovec and Isabelle Levent and Xiang Lisa Li and Xuechen Li and Tengyu Ma and Ali Malik and Christopher D. Manning and Suvir Mirchandani and Eric Mitchell and Zanele Munyikwa and Suraj Nair and Avanika Narayan and Deepak Narayanan and Ben Newman and Allen Nie and Juan Carlos Niebles and Hamed Nilforoshan and Julian Nyarko and Giray Ogut and Laurel Orr and Isabel Papadimitriou and Joon Sung Park and Chris Piech and Eva Portelance and Christopher Potts and Aditi Raghunathan and Rob Reich and Hongyu Ren and Frieda Rong and Yusuf Roohani and Camilo Ruiz and Jack Ryan and Christopher Ré and Dorsa Sadigh and Shiori Sagawa and Keshav Santhanam and Andy Shih and Krishnan Srinivasan and Alex Tamkin and Rohan Taori and Armin W. Thomas and Florian Tramèr and Rose E. Wang and William Wang and Bohan Wu and Jiajun Wu and Yuhuai Wu and Sang Michael Xie and Michihiro Yasunaga and Jiaxuan You and Matei Zaharia and Michael Zhang and Tianyi Zhang and Xikun Zhang and Yuhui Zhang and Lucia Zheng and Kaitlyn Zhou and Percy Liang},
	year={2022},
	eprint={2108.07258},
	note={arXiv:2108.07258},
	archivePrefix={arXiv},
	primaryClass={cs.LG},
	url={https://arxiv.org/abs/2108.07258}, 
}

@ARTICLE{SSLsurvey2021,
	author={Jing, Longlong and Tian, Yingli},
	journal={ IEEE Transactions on Pattern Analysis \& Machine Intelligence },
	title={{ Self-Supervised Visual Feature Learning With Deep Neural Networks: A Survey }},
	year={2021},
	volume={43},
	number={11},
	ISSN={1939-3539},
	pages={4037-4058},
	doi={10.1109/TPAMI.2020.2992393},
	url = {https://doi.ieeecomputersociety.org/10.1109/TPAMI.2020.2992393},
	publisher={IEEE Computer Society},
	address={Los Alamitos, CA, USA},
	month=nov
}

@misc{caron2021,
	title={Emerging Properties in Self-Supervised Vision Transformers}, 
	author={Mathilde Caron and Hugo Touvron and Ishan Misra and Hervé Jégou and Julien Mairal and Piotr Bojanowski and Armand Joulin},
	year={2021},
	eprint={2104.14294},
	note={arXiv:2104.14294},
	archivePrefix={arXiv},
	primaryClass={cs.CV},
	url={https://arxiv.org/abs/2104.14294}, 
}

@article{phikon,
	title={Scaling self-supervised learning for histopathology with masked image modeling},
	author={Filiot, Alexandre and Ghermi, Ridouane and Olivier, Antoine and Jacob, Paul and Fidon, Lucas and Mac Kain, Alice and Saillard, Charlie and Schiratti, Jean-Baptiste},
	journal={medRxiv},
	year={2023},
	doi={10.1101/2023.07.21.23292757},
	url={https://www.medrxiv.org/content/10.1101/2023.07.21.23292757},
}

@misc{phikonv2,
	title={Phikon-v2, a large and public feature extractor for biomarker prediction},
	author={Filiot, Alexandre and Jacob, Paul and Mac Kain, Alice and Saillard, Charlie},
	year={2024},
	eprint={2409.09173},
	note={arXiv:2409.09173},
	archivePrefix={arXiv},
	primaryClass={eess.IV},
	url={https://arxiv.org/abs/2409.09173},
}

@misc{hibou,
	title={Hibou: a family of foundational vision transformers for pathology},
	author={Nechaev, Dmitry and Pchelnikov, Alexey and Ivanova, Ekaterina},
	year={2024},
	eprint={2406.05074},
	note={arXiv:2406.05074},
	archivePrefix={arXiv},
	primaryClass={eess.IV},
	url={https://arxiv.org/abs/2406.05074},
}

@article{mstar,
	title={A multimodal knowledge-enhanced whole-slide pathology foundation model},
	author={Xu, Yingxue and Wang, Yihui and Zhou, Fengtao and Ma, Jiabo and Jin, Cheng and Yang, Shu and Li, Jinbang and Zhang, Zhengyu and Zhao, Chenglong and Zhou, Huajun and Li, Zhenhui and Lin, Huangjing and Wang, Xin and Wang, Jiguang and Han, Anjia and Chan, Ronald Cheong Kin and Liang, Li and Zhang, Xiuming and Chen, Hao},
	journal={Nature Communications},
	volume={16},
	number={1},
	pages={11406},
	year={2025},
	doi={10.1038/s41467-025-66220-x},
}

@article{gpfm,
	title={A generalizable pathology foundation model using a unified knowledge distillation pretraining framework},
	author={Ma, Jiabo and Guo, Zhengrui and Zhou, Fengtao and Wang, Yihui and Xu, Yingxue and Li, Jinbang and Yan, Fang and Cai, Yu and Zhu, Zhengjie and Jin, Cheng and Lin, Yi and Jiang, Xinrui and Zhao, Chenglong and Li, Danyi and Han, Anjia and Li, Zhenhui and Chan, Ronald Cheong Kin and Wang, Jiguang and Fei, Peng and Cheng, Kwang-Ting and Zhang, Shaoting and Liang, Li and Chen, Hao},
	journal={Nature Biomedical Engineering},
	volume={10},
	number={3},
	pages={545--564},
	year={2026},
	doi={10.1038/s41551-025-01488-4},
}

@article{musk,
	title={A vision--language foundation model for precision oncology},
	author={Xiang, Jinxi and Wang, Xiyue and Zhang, Xiaoming and Xi, Yinghua and Eweje, Feyisope and Chen, Yijiang and Li, Yuchen and Bergstrom, Colin and Gopaulchan, Matthew and Kim, Ted and Yu, Kun-Hsing and Willens, Sierra and Olguin, Francesca Maria and Nirschl, Jeffrey J. and Neal, Joel and Diehn, Maximilian and Yang, Sen and Li, Ruijiang},
	journal={Nature},
	volume={638},
	number={8051},
	pages={769--778},
	year={2025},
	doi={10.1038/s41586-024-08378-w},
}

@article{genbiopathfm,
	title={{GenBio-PathFM}: a state-of-the-art foundation model for histopathology},
	author={Kapse, Saarthak and Ayg{\"u}n, Mehmet and Cole, Elijah and Lundberg, Emma and Song, Le and Xing, Eric P.},
	journal={bioRxiv},
	year={2026},
	doi={10.64898/2026.03.17.712534},
	url={https://doi.org/10.64898/2026.03.17.712534},
}

@article{prism,
	title={{PRISM}: a multi-modal generative foundation model for slide-level histopathology},
	author={Shaikovski, George and Casson, Adam and Severson, Kristen and Zimmermann, Eric and Wang, Yi Kan and Kunz, Jeremy D. and Retamero, Juan A. and Oakley, Gerard and Klimstra, David and Kanan, Christopher and Hanna, Matthew and Zelechowski, Michal and Viret, Julian and Tenenholtz, Neil and Hall, James and Fusi, Nicolo and Yousfi, Razik and Hamilton, Peter and Moye, William A. and Vorontsov, Eugene and Liu, Siqi and Fuchs, Thomas J.},
	journal={arXiv preprint arXiv:2405.10254},
	year={2024},
	doi={10.48550/arXiv.2405.10254},
	url={https://arxiv.org/abs/2405.10254},
}

@article{titan,
	title={A multimodal whole-slide foundation model for pathology},
	author={Ding, Tong and Wagner, Sophia J. and Song, Andrew H. and Chen, Richard J. and Lu, Ming Y. and Zhang, Andrew and Vaidya, Anurag J. and Jaume, Guillaume and Shaban, Muhammad and Kim, Ahrong and Williamson, Drew F. K. and Robertson, Harry and Chen, Bowen and Almagro-P{\'e}rez, Cristina and Doucet, Paul and Sahai, Sharifa and Chen, Chengkuan and Chen, Christina S. and Komura, Daisuke and Kawabe, Akihiro and Ochi, Mieko and Sato, Shinya and Yokose, Tomoyuki and Miyagi, Yohei and Ishikawa, Shumpei and Gerber, Georg and Peng, Tingying and Le, Long Phi and Mahmood, Faisal},
	journal={Nature Medicine},
	volume={31},
	number={11},
	pages={3749--3761},
	year={2025},
	doi={10.1038/s41591-025-03982-3},
}

@article{par,
	title={The {PAR} dataset: Prostate biopsy whole slide images from an underrepresented Middle Eastern population},
	author={{Muhammad Ali}, Peshawa J. and Vincent, Navin and Abdulla, Saman S. and {Mohammed Fadhl}, Han N. and Blilie, Anders and Szolnoky, Kelvin and Mielcarz, Julia Anna and Ji, Xiaoyi and Kartasalo, Kimmo and {Al-Talabani}, Abdulbasit K. and Mulliqi, Nita},
	journal={Scientific Data},
	volume={13},
	pages={1061},
	year={2026},
	doi={10.1038/s41597-026-07798-9},
}

@misc{oquab2024,
	title={{DINOv2}: Learning Robust Visual Features without Supervision},
	author={Maxime Oquab and Timothée Darcet and Théo Moutakanni and Huy Vo and Marc Szafraniec and Vasil Khalidov and Pierre Fernandez and Daniel Haziza and Francisco Massa and Alaaeldin El-Nouby and Mahmoud Assran and Nicolas Ballas and Wojciech Galuba and Russell Howes and Po-Yao Huang and Shang-Wen Li and Ishan Misra and Michael Rabbat and Vasu Sharma and Gabriel Synnaeve and Hu Xu and Hervé Jegou and Julien Mairal and Patrick Labatut and Armand Joulin and Piotr Bojanowski},
	year={2024},
	eprint={2304.07193},
	note={arXiv:2304.07193},
	archivePrefix={arXiv},
	primaryClass={cs.CV},
	url={https://arxiv.org/abs/2304.07193},
}

@article{tellez2019,
	title={Quantifying the effects of data augmentation and stain color normalization in convolutional neural networks for computational pathology},
	author={Tellez, David and Litjens, Geert and B{\'a}ndi, P{\'e}ter and Bulten, Wouter and Bokhorst, John-Melle and Ciompi, Francesco and van der Laak, Jeroen},
	journal={Medical Image Analysis},
	volume={58},
	pages={101544},
	year={2019},
	doi={10.1016/j.media.2019.101544},
}

@article{buttner2019kbet,
	title={A test metric for assessing single-cell {RNA}-seq batch correction},
	author={B{\"u}ttner, Maren and Miao, Zhichao and Wolf, F. Alexander and Teichmann, Sarah A. and Theis, Fabian J.},
	journal={Nature Methods},
	volume={16},
	number={1},
	pages={43--49},
	year={2019},
	doi={10.1038/s41592-018-0254-1},
}

@article{korsunsky2019harmony,
	title={Fast, sensitive and accurate integration of single-cell data with {Harmony}},
	author={Korsunsky, Ilya and Millard, Nghia and Fan, Jean and Slowikowski, Kamil and Zhang, Fan and Wei, Kevin and Baglaenko, Yuriy and Brenner, Michael and Loh, Po-ru and Raychaudhuri, Soumya},
	journal={Nature Methods},
	volume={16},
	number={12},
	pages={1289--1296},
	year={2019},
	doi={10.1038/s41592-019-0619-0},
}

@article{luecken2022scib,
	title={Benchmarking atlas-level data integration in single-cell genomics},
	author={Luecken, Malte D. and B{\"u}ttner, Maren and Chaichoompu, Kridsadakorn and Danese, Anna and Interlandi, Marta and Mueller, Michaela F. and Strobl, Daniel C. and Zappia, Luke and Dugas, Martin and Colom{\'e}-Tatch{\'e}, Maria and Theis, Fabian J.},
	journal={Nature Methods},
	volume={19},
	number={1},
	pages={41--50},
	year={2022},
	doi={10.1038/s41592-021-01336-8},
}

\section*{Acknowledgements}
This work was supported by the European Research Council under the European Union's Horizon Europe programme (ERC Starting Grant AIS-CaP, grant agreement no.~101041730, to G.L.). We acknowledge financial support by Mr.~H.~J.~M.~Roels through a donation to Oncode Institute.

\section*{Author contributions}
C.G. conceived the study, curated the data, performed the experiments and analyses, and wrote the manuscript. J.v.d.L. and G.L. supervised the study. All authors reviewed the manuscript.

\section*{Competing interests}
J.v.d.L. reports consulting fees from Philips, ContextVision and AbbVie, and grants from Philips, ContextVision and Sectra, outside the submitted work. G.L. reports grants from the Dutch Cancer Society and the Dutch Research Council (NWO), grants from Philips Digital Pathology Solutions and personal fees from Novartis, outside the submitted work. All other authors declare no competing interests.

\clearpage
\appendix
\setcounter{figure}{0}\renewcommand{\thefigure}{S\arabic{figure}}
\setcounter{table}{0}\renewcommand{\thetable}{S\arabic{table}}
\section{Supplementary material}

\subsection{Example samples}
\label{supp:example-patches}

\begin{figure}[!htbp]
\centering
\includegraphics[width=\linewidth]{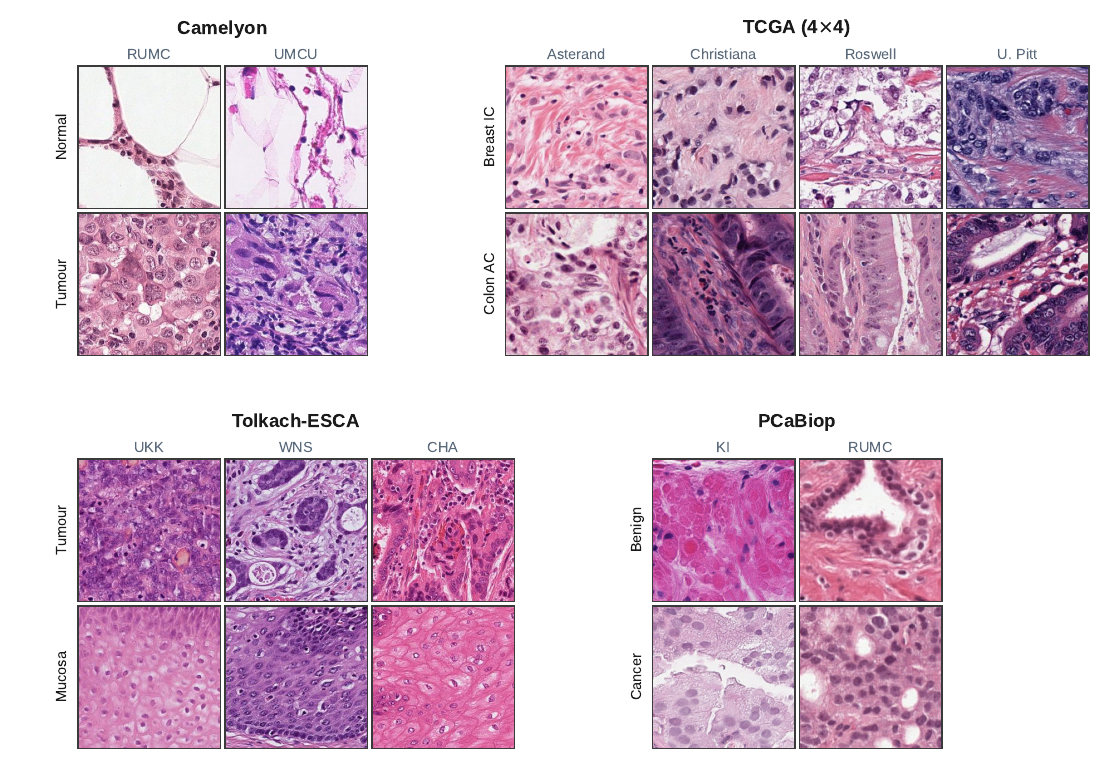}
\caption{\textbf{Biology-matched example samples across acquisition centres.} Each benchmark block shows two biological classes (rows) across all acquisition centres (columns), so within a row biology is fixed and only the centre changes. Because PCaBiop is a slide-level benchmark, its block shows representative tiles rather than whole slides, which make the visual differences between its centres easier to see. The resulting staining, tissue preparation and scanning differences are clearly visible: Camelyon's UMCU tiles are markedly more purple than RUMC's, and the three Tolkach-ESCA cohorts differ in overall hue.}
\label{fig:dataset-examples}
\end{figure}

\subsection{Sensitivity of \texorpdfstring{$\mcode{LTM}_{\alpha}$}{LTM} to the tail fraction \texorpdfstring{$\alpha$}{alpha}}
\label{supp:alpha-sweep}

The tail-severity summary $\mcode{LTM}_{\alpha}$ requires a tail fraction $\alpha$, fixed at $\alpha=0.10$ in the main text. Here we show that model comparison by $\mcode{LTM}_{\alpha}$ is insensitive to this choice over a plausible range. For each benchmark we recompute $\mcode{LTM}_{\alpha}$ per model at $\alpha\in\{0.05,0.10,0.20\}$ from the per-sample \code{CRoMa}$(m{=}5)$ values, rank the models, and measure the Spearman rank correlation between the rankings induced by different $\alpha$. Rankings are stable between adjacent tail fractions on every benchmark (Table~\ref{tab:alpha-stability}): the $0.05$-vs-$0.10$ and $0.10$-vs-$0.20$ correlations are $\ge 0.87$ throughout, and the $0.05$-vs-$0.10$ correlation exceeds $0.94$ on four of the five benchmarks. As expected, the widest comparison ($0.05$ vs $0.20$) is looser because the $5\%$ and $20\%$ tails probe increasingly different subpopulations. The default $\alpha=0.10$ therefore sits in a stable interior region, and the tail-analysis conclusions do not hinge on it.

\begin{table}[!htbp]
\centering
\small
\begin{tabular}{lccc}
\hline
Benchmark & $\rho(0.05,0.10)$ & $\rho(0.10,0.20)$ & $\rho(0.05,0.20)$ \\
\hline
Camelyon           & $0.98$ & $0.98$ & $0.94$ \\
TCGA-2x2  & $0.98$ & $0.91$ & $0.84$ \\
TCGA-4x4  & $0.99$ & $0.87$ & $0.81$ \\
Tolkach-ESCA       & $0.91$ & $0.92$ & $0.73$ \\
PCaBiop              & $1.00$ & $1.00$ & $1.00$ \\
\hline
\end{tabular}
\caption{\textbf{Rank correlations across lower-tail fractions.} Entries are Spearman correlations between per-model $\mcode{LTM}_{\alpha}$ rankings at $\alpha=0.05$, $0.10$ and $0.20$, computed from \code{CRoMa}$(m{=}5)$ within each benchmark. Each tile benchmark ranks \TileRankedNModels{} encoders. PCaBiop ranks \SlideNModels{}, so its correlations rest on a far smaller panel.}
\label{tab:alpha-stability}
\end{table}

\subsection{Robustness of \texorpdfstring{\code{CRoMa}}{CRoMa} model comparison to the averaging radius \texorpdfstring{$m$}{m}}
\label{supp:m-sweep}

\code{CRoMa} averages the distances to the $m$ nearest neighbours of each type, fixed at $m=5$ in the main text. Here we show that model comparison is insensitive to this choice over a wide range of $m$. For the three tile benchmarks (Camelyon, TCGA-4x4, and Tolkach-ESCA) we take the pooled $\mcode{CRoMa}$ per model at every integer $m\in[1,20]$ and track both the ranking and the number of confounder-dominant models ($\mcode{CRoMa}<0$). The comparison is essentially invariant across the sweep (Table~\ref{tab:m-sweep}). Model rankings are highly concordant for every benchmark (Spearman $\rho\ge0.98$ between the extremes $m=1$ and $m=20$), and the number of confounder-dominant models is constant across the full sweep. Increasing $m$ therefore affects only margin magnitudes -- and the lower-tail summaries derived from them -- without altering model order or the sign of the dominant signal. Thus, $m=5$ represents a practical operating point rather than a tuned parameter: averaging over five neighbours limits the contribution of any single neighbour to one-fifth of either type-specific mean.

% AUTO-GENERATED by scripts/repro/generate_m_sweep_table.py -- do not edit by hand.
\begin{table}[!htbp]
	\centering
	\begin{tabular}{lcccc}
		\hline
		Benchmark & $n$ & $\rho(m{=}1,m{=}5)$ & $\rho(m{=}5,m{=}20)$ & $\rho(m{=}1,m{=}20)$ \\
		\hline
		Camelyon & $20$ & $0.992$ & $0.989$ & $0.982$ \\
		TCGA-4x4 & $20$ & $0.992$ & $0.992$ & $0.983$ \\
		Tolkach-ESCA & $20$ & $0.994$ & $0.992$ & $0.988$ \\
		\hline
	\end{tabular}
	\caption{\textbf{Robustness of \code{CRoMa} model comparison to the averaging radius $m$.} For each of the three tile benchmarks, we report Spearman correlations between the median $\mcode{CRoMa}$ model rankings at the headline radius $m=5$ and each extreme ($m=1$, $m=20$), and between the two extremes.}
	\label{tab:m-sweep}
\end{table}

\subsection{Why the selected neighbourhood size undercuts fixed-\texorpdfstring{$k$}{k} coverage}
\label{supp:k-selection}

Fixed-$k$ metrics derive their neighbourhood size from a biological $k$-NN criterion: $k$ is chosen to maximise biological $k$-NN classification accuracy (Methods). A biological $k$-NN classifier votes on the biology label of \emph{every} neighbour in the window. Since same-biology, same-centre samples are naturally closer than other neighbour types, the classifier's accuracy is dominated by the dense \textbf{SS} pocket surrounding each anchor. On Camelyon, \textbf{SS} neighbours comprise $86$--$93\%$ of the selected window. Maximising that accuracy therefore drives the selected size small (median $k=11$ on Camelyon, per-model optima $5$--$61$), toward the very \textbf{SS} neighbours the robustness metrics discard as uninformative and rarely extending far enough to capture the typed \textbf{SO}/\textbf{OS} neighbours they score. The first \textbf{SO} or \textbf{OS} neighbour occurs only at a median rank of approximately $149$, far beyond the operated $k=11$. The coverage gap reported in the main text follows directly: a window selected by an \textbf{SS}-dominated classification objective is, almost by construction, too small to include the typed neighbours required by \code{RI}. Most anchors therefore contain no typed evidence and are silently excluded from the pooled score. \code{CRoMa} avoids this conflict by imposing no fixed neighbourhood at all, reading each typed neighbour at whatever rank it occurs (Section~\ref{sec:croma}).

\begin{table}[!htbp]
\centering
\footnotesize
\setlength{\tabcolsep}{3pt}
\begin{tabular}{lrrrrrrr}
\hline
& & \multicolumn{2}{c}{median rank}
& & \multicolumn{3}{c}{among supported samples (\%)} \\
\cline{3-4}\cline{6-8}
Model & \code{CRoMa} & \textbf{SO} & \textbf{OS} & support (\%) & pure-\textbf{SO} & pure-\textbf{OS} & contested \\
\hline
Virchow2      &  0.20 &  111 & 476 & 31.4 & 80.1 & 14.8 &  5.1 \\
CONCH         &  0.20 &   97 & 312 & 35.9 & 65.0 & 27.2 &  7.8 \\
GenBio-PathFM &  0.19 &  114 & 694 & 38.3 & 84.0 & 11.4 &  4.6 \\
CONCHv1.5     &  0.19 &   55 & 311 & 46.3 & 74.5 & 17.9 &  7.6 \\
H0-mini       &  0.17 &  134 & 393 & 38.7 & 73.7 & 19.0 &  7.3 \\
Virchow       &  0.16 &  194 & 450 & 26.4 & 70.3 & 25.0 &  4.6 \\
Midnight-12k  &  0.11 &  241 & 338 & 19.8 & 44.4 & 50.6 &  4.9 \\
H-optimus-1   &  0.08 &  276 & 550 & 17.2 & 65.8 & 28.7 &  5.5 \\
H-optimus-0   &  0.05 &  261 & 367 & 23.7 & 61.2 & 32.3 &  6.5 \\
UNI2-h        &  0.04 &  379 & 570 & 13.0 & 51.2 & 45.7 &  3.1 \\
MUSK          &  0.04 &  195 & 197 & 28.0 & 37.0 & 52.5 & 10.6 \\
mSTAR         &  0.02 &  329 & 338 & 18.0 & 43.8 & 50.3 &  5.9 \\
Prov-GigaPath &  0.01 &  445 & 376 & 14.4 & 37.9 & 57.5 &  4.6 \\
UNI           & -0.03 &  933 & 414 &  9.6 & 13.2 & 85.2 &  1.6 \\
Hibou-B       & -0.09 & 1211 & 304 & 13.4 &  6.5 & 91.2 &  2.3 \\
GPFM          & -0.10 &  785 & 193 & 20.9 &  4.5 & 94.4 &  1.1 \\
Phikon        & -0.20 & 2216 & 313 & 16.6 &  1.6 & 98.4 &  0.1 \\
Phikon-v2     & -0.21 & 2909 & 269 & 16.9 &  3.0 & 96.9 &  0.2 \\
Prost40M      & -0.32 & 1817 & 124 & 27.2 &  1.9 & 97.8 &  0.3 \\
Hibou-L       & -0.44 & 9089 & 327 & 12.1 &  1.3 & 98.4 &  0.2 \\
\hline
DINOv2-B      &  0.05 &   32 &  50 & 68.0 & 46.3 & 30.4 & 23.3 \\
\hline
\end{tabular}
\caption{\textbf{Typed-neighbour ranks and fixed-$k$ neighbourhood composition on Camelyon.} Pathology encoders are ordered by median \code{CRoMa} ($m{=}5$). Positive values indicate biology-dominant geometry. For each encoder, the \textbf{SO} and \textbf{OS} columns report the median rank across samples of the nearest cross-confounder biological match and same-confounder biological distractor, respectively. At the shared operating point $k{=}11$, support is the percentage of all samples whose neighbourhood contains at least one \textbf{SO} or \textbf{OS} neighbour. Pure-\textbf{SO}, pure-\textbf{OS} and contested neighbourhoods partition these supported samples and are reported as percentages of support. Only contested neighbourhoods contain both evidence types and can yield different \code{RI} and \code{MaRI} values. Across the \CamelyonRankedNModels{} pathology encoders, the nearest typed neighbour occurs at a median rank of approximately $149$ in a cohort of $20{,}400$ tiles, well beyond the fixed $k{=}11$ window. The natural-image control \code{DINOv2-B} is shown separately.}
\label{tab:typed-neighbour-coverage}
\end{table}

\subsection{A geometric interpretation of the \texorpdfstring{\code{CRoMa}}{CRoMa} margin}
\label{supp:geometry}

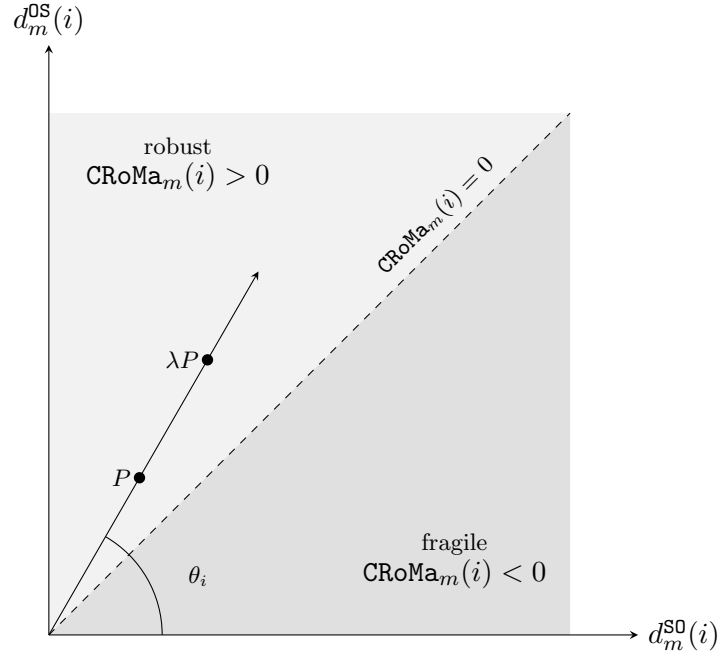
\begin{figure}[!htbp]
\centering
\begin{tikzpicture}[scale=3.0,>=stealth,line join=round]
  % regions
  \fill[black!5]  (0,0) -- (2.3,2.3) -- (0,2.3) -- cycle;   % above diagonal: robust
  \fill[black!12] (0,0) -- (2.3,2.3) -- (2.3,0) -- cycle;   % below diagonal: fragile
  % axes
  \draw[->] (0,0) -- (2.6,0) node[right] {$d^{\mcode{SO}}_m(i)$};
  \draw[->] (0,0) -- (0,2.6) node[above] {$d^{\mcode{OS}}_m(i)$};
  % diagonal (CRoMa = 0)
  \draw[dashed] (0,0) -- (2.3,2.3);
  \node[rotate=45,anchor=south east] at (2.05,2.05) {\footnotesize $\mcode{CRoMa}_m(i)=0$};
  % region labels, tucked into the far corner of their own region
  \node[align=center,anchor=north west] at (0.12,2.24) {\footnotesize robust\\[-2pt]$\mcode{CRoMa}_m(i)>0$};
  \node[align=center,anchor=south east] at (2.24,0.16) {\footnotesize fragile\\[-2pt]$\mcode{CRoMa}_m(i)<0$};
  % ray + two points (scale invariance)
  \draw[->] (0,0) -- (60:1.85);
  \fill (60:0.80) circle (0.025) node[left,inner sep=3pt] {\footnotesize $P$};
  \fill (60:1.40) circle (0.025) node[left,inner sep=3pt] {\footnotesize $\lambda P$};
  % angle arc
  \draw (0.5,0) arc (0:60:0.5);
  \node at (0.66,0.26) {\footnotesize $\theta_i$};
\end{tikzpicture}
\caption{\textbf{\code{CRoMa} is the angular offset from the diagonal.} In the $(d^{\mcode{SO}}_m,d^{\mcode{OS}}_m)$ plane, $\mcode{CRoMa}_m(i)=\tan(\theta_i-\pi/4)$: points above the diagonal have more distant \textbf{OS} distractors and positive margins, points below have negative margins. $P$ and $\lambda P$ lie on one ray, so they share $\theta_i$ and therefore the same margin.}
\label{fig:croma-geometry}
\end{figure}

Because $\mcode{CRoMa}_m(i)$ depends only on the ratio $d^{\mcode{OS}}_m/d^{\mcode{SO}}_m$, it has a simple reading in the plane of typed mean distances $(d^{\mcode{SO}}_m,d^{\mcode{OS}}_m)$ (Figure~\ref{fig:croma-geometry}). Both typed distances are nonnegative, so the pair lies in the first quadrant, where its polar angle is simply:

$$
\theta_i=\arctan\!\left(\frac{d^{\mcode{OS}}_m}{d^{\mcode{SO}}_m}\right)=\arctan\left(r_i\right),
$$
\noindent
with $r_i$ the distance ratio of Section~\ref{sec:methods-croma}. This gives:

$$
\mcode{CRoMa}_m(i)=\tan\!\left(\theta_i-\frac{\pi}{4}\right).
$$
\noindent
Thus, $\mcode{CRoMa}_m(i)$ measures the signed angular deviation from the $45^\circ$ diagonal. Equal typed distances lie on this diagonal, $d^{\mcode{SO}}_m=d^{\mcode{OS}}_m$, and yield $\mcode{CRoMa}_m(i)=0$. Points above the diagonal, where the distractor is farther away, are robust ($\mcode{CRoMa}_m(i)>0$), and points below it are fragile ($\mcode{CRoMa}_m(i)<0$). The score is independent of radial distance: moving a point along a ray from the origin scales both typed distances by a common factor $\lambda>0$, which cancels in the ratio $d^{\mcode{OS}}_m/d^{\mcode{SO}}_m$ and leaves $\theta_i$---and hence $\mcode{CRoMa}_m(i)$---unchanged, so all points on a ray share the same margin. This cancellation is precisely the scale-free property noted in Section~\ref{sec:methods-croma}: only the \emph{direction} of the typed-distance pair carries signal, not its length.\\
\\

\begin{figure}[!htbp]
\centering
\includegraphics[width=0.54\linewidth]{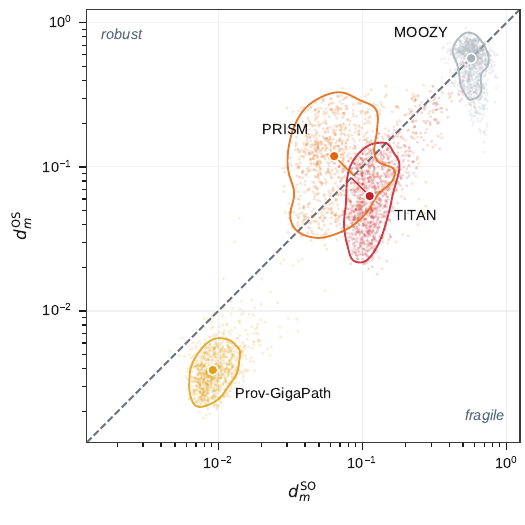}
\caption{\textbf{The typed-distance plane measured on PCaBiop.} One point per slide for the \SlideNModels{} slide-level encoders ($m=5$). Outlines enclose $75\%$ of an encoder's slides and filled markers are per-encoder medians. Axes are logarithmic because median typed-distance radii span a factor of $80$, from $0.010$ (\code{Prov-GigaPath}) to $0.80$ (\code{MOOZY}). The log map turns a common rescaling into a shift along the diagonal, so position along the diagonal is scale and offset from it is robustness: the angular offset $\theta_i-\pi/4$ of Figure~\ref{fig:croma-geometry} becomes a perpendicular displacement from the diagonal of length $\lvert\log r\rvert/\sqrt{2}$, where $r=d^{\mcode{OS}}_m/d^{\mcode{SO}}_m$, the side of the diagonal carrying its sign. \code{PRISM} and \code{TITAN} illustrate this best: both models sit at nearly the same radius ($0.14$ versus $0.13$) but on opposite sides of the diagonal (\code{CRoMa} $=$ \PandaBestCroma{} versus $-0.30$). The plane also separates failure modes the sign alone conflates: \code{Prov-GigaPath} is fragile with all typed distances collapsed toward zero, whereas \code{TITAN} is fragile at an ordinary distance scale.}
\label{fig:croma-geometry-planview}
\end{figure}

\noindent
This scale invariance is what lets \code{CRoMa} margins be compared across settings that hold one factor fixed and vary the other. The absolute value of $d^{\mcode{SO}}_m$ and $d^{\mcode{OS}}_m$ is a nuisance quantity: it reflects the encoder (embedding norm, output temperature, feature dimension) and the benchmark (biological classes, cohort composition) rather than robustness itself. With the benchmark fixed and the model varied, an encoder that merely spreads all of its distances wider is not credited as more robust, because the shared factor $\lambda$ cancels and only the relative separation of cross- and same-confounder neighbours---the angle $\theta_i$---remains. With the model fixed and the benchmark varied, datasets that sit at different typical distance magnitudes are placed on one common scale, so a model's margin on Camelyon reads on the same footing as its margin on TCGA-4x4.\\
\\
The nuisance is large in practice, not hypothetical. Measuring the plane directly on PCaBiop (Figure~\ref{fig:croma-geometry-planview}) places the \SlideNModels{} slide-level encoders at median typed-distance radii spanning a factor of $80$, so their point clouds sit at four widely separated positions along the diagonal. Their robustness is unrelated to that position. \code{PRISM} and \code{TITAN} make the point sharply: they occupy nearly the same radius yet fall on opposite sides of the diagonal. Reading robustness off the raw typed distances would therefore rank encoders largely by embedding geometry. Our ratio removes that dependence by construction.

\subsection{Metric rank agreement within each benchmark}
\label{supp:rank-agreement}

Table~\ref{tab:rank-agreement} collects the nine within-benchmark rank correlations among \code{RI}, \code{MaRI} and \code{CRoMa} referenced in the main text.

\begin{table}[!htbp]
\centering
\small
\begin{tabular}{lccc}
\hline
Benchmark & \code{RI} vs \code{MaRI} & \code{CRoMa} vs \code{RI} & \code{CRoMa} vs \code{MaRI} \\
\hline
Camelyon          & \CamelyonRiVsMariRho       & \CamelyonCromaVsRiRho       & \CamelyonCromaVsMariRho \\
TCGA 4x4 & \TcgaFourByFourRiVsMariRho & \TcgaFourByFourCromaVsRiRho & \TcgaFourByFourCromaVsMariRho \\
Tolkach-ESCA      & \TolkachRiVsMariRho        & \TolkachCromaVsRiRho        & \TolkachCromaVsMariRho \\
\hline
\end{tabular}
\caption{\textbf{Pairwise rank correlations among \code{RI}, \code{MaRI} and \code{CRoMa}.} Spearman correlations across the \CamelyonRankedNModels{} pathology encoders shared by the three tile benchmarks. The three metrics induce nearly identical model orderings on every tile benchmark: \code{RI} and \code{MaRI} are almost interchangeable as rankers, and \code{CRoMa} tracks both wherever the fixed-$k$ scores are defined, despite scoring a different and larger population of anchors. A shared ranking is not interchangeability, however: encoders with matched \code{MaRI} can separate sharply under \code{CRoMa} (Section~\ref{sec:croma}), and only the per-sample construction exposes the confounder-dominated tail (Section~\ref{sec:tail}).}
\label{tab:rank-agreement}
\end{table}

\subsection{Pretraining provenance}
\label{supp:pretraining-overlap}

% HARDCODED: 1.21x (DINOv2-B), 1.36x/1.26x (median boost, TCGA-exposed vs rest), 1.35x
% (Phikon) are medians over the Boost column of tab:pretraining-overlap, computed from the
% rounded 2-dp values shown there (exposed n=9, rest n=11; DINOv2-B excluded from both).

\code{\ProvenanceModel} was pretrained exclusively on TCGA, so its wide lead on TCGA-4x4---\code{CRoMa}$=$\TcgaFourByFourCromaMax{} versus \ProvenanceRunnerUpCroma{} for the runner-up, \code{\ProvenanceRunnerUpModel}---could partly reflect pretraining overlap. To test this, we re-scored a larger Tolkach-ESCA cohort that adds the held-out TCGA cases and measured the resulting robustness gains (Supplementary Table~\ref{tab:pretraining-overlap}). \code{\ProvenanceModel} shows the largest TCGA-cohort boost of any encoder (\ProvenanceTolkachBoost{} in typed-distance odds), but exposure alone does not explain it: every encoder gains on this cohort, including the natural-image control ($1.21\times$), and the $9$ TCGA-exposed encoders are barely separated from the rest (median $1.36\times$ versus $1.26\times$). \code{Phikon}, also pretrained exclusively on public TCGA, gains only $1.35\times$. \code{\ProvenanceModel} still leads once the TCGA cases are removed. Pretraining overlap therefore amplifies, rather than creates, the observed advantage.

\begin{table}[!htbp]
\centering
\small
\begin{tabular}{lccc}
\hline
Model & Non-TCGA cohorts & TCGA cohort & Boost \\
\hline
Midnight-12k$^{\dagger}$ & 0.60 & 0.82 & 2.50$\times$ \\
H0-mini$^{\dagger}$ & 0.38 & 0.53 & 1.47$\times$ \\
Virchow & 0.37 & 0.52 & 1.45$\times$ \\
Virchow2 & 0.37 & 0.51 & 1.44$\times$ \\
Hibou-L & 0.12 & 0.28 & 1.41$\times$ \\
UNI2-h & 0.23 & 0.38 & 1.39$\times$ \\
GenBio-PathFM$^{\dagger}$ & 0.41 & 0.53 & 1.37$\times$ \\
GPFM$^{\dagger}$ & 0.24 & 0.38 & 1.37$\times$ \\
Prost40M$^{\dagger}$ & 0.14 & 0.28 & 1.36$\times$ \\
Phikon$^{\dagger}$ & 0.17 & 0.31 & 1.35$\times$ \\
H-optimus-0 & 0.24 & 0.37 & 1.32$\times$ \\
MUSK$^{\dagger}$ & 0.30 & 0.42 & 1.32$\times$ \\
Phikon-v2$^{\dagger}$ & 0.13 & 0.26 & 1.31$\times$ \\
mSTAR$^{\dagger}$ & 0.20 & 0.32 & 1.28$\times$ \\
Prov-GigaPath & 0.13 & 0.24 & 1.26$\times$ \\
H-optimus-1 & 0.27 & 0.38 & 1.26$\times$ \\
CONCH & 0.44 & 0.52 & 1.24$\times$ \\
CONCHv1.5 & 0.41 & 0.49 & 1.22$\times$ \\
Hibou-B & 0.14 & 0.23 & 1.22$\times$ \\
UNI & 0.18 & 0.25 & 1.17$\times$ \\
\hline
DINOv2-B & 0.17 & 0.27 & 1.21$\times$ \\
\hline
\end{tabular}
\caption{\textbf{Median \code{CRoMa} across the full Tolkach-ESCA tileset.} Median per-sample $\mcode{CRoMa}(m{=}5)$ for the \TolkachRankedNModels{} tile-level pathology encoders, computed over all $13{,}800$ tiles from the three non-TCGA cohorts and separately over the $2{,}500$ tiles from the TCGA cohort. These row sets differ from the balanced $9{,}000$-tile evaluation cohort used in Table~\ref{tab:main-results-tolkach}; the corresponding values therefore need not be identical. The \emph{boost} measures how much further the nearest same-confounder biological distractors sit (relative to the nearest cross-confounder biological matches) on the TCGA cases than on the non-TCGA cohorts. It is the between-cohort ratio of $r=d^{\mcode{OS}}_m/d^{\mcode{SO}}_m$, the typed-distance ratio on which \code{CRoMa} is defined (Section~\ref{sec:methods-croma}). $\dagger$ marks the $9$ TCGA-exposed encoders (Table~\ref{tab:model-summary}). The natural-image control \code{DINOv2-B} is shown separately.}
\label{tab:pretraining-overlap}
\end{table}

\clearpage

\subsection{Per-model results on TCGA-4x4 and Tolkach-ESCA}
\label{supp:tile-tables}

Supplementary Tables~\ref{tab:main-results-tcga4x4} and~\ref{tab:main-results-tolkach} report the full per-model results on TCGA-4x4 and Tolkach-ESCA, in the same format as the Camelyon results in Table~\ref{tab:main-results}. The cross-benchmark synthesis in Section~\ref{sec:cross-benchmark} draws on these values.

\begin{table}[!htbp]
\centering
\small
\setlength{\tabcolsep}{4pt}
\begin{tabular}{lccccccccc}
\hline
Model & bio bacc & conf bacc & \code{RI} & \code{MaRI} & $\Delta$ & \code{CRoMa} & $F(0)$ & $\mcode{LTM}_{10}$ & support \\
\hline
Midnight-12k$^{\dagger}$ & \textbf{0.882} & 0.543 & \textbf{0.898} & \textbf{0.936} & $+0.037$ & \textbf{0.40} & \textbf{0.140} & -0.21 & 99.4\% \\
GenBio-PathFM$^{\dagger}$ & 0.851 & 0.695 & 0.757 & 0.780 & $+0.023$ & 0.16 & 0.242 & -0.19 & 99.9\% \\
CONCHv1.5 & 0.811 & 0.482 & 0.828 & 0.851 & $+0.024$ & 0.15 & 0.193 & -0.13 & 100.0\% \\
CONCH & 0.790 & 0.480 & 0.801 & 0.825 & $+0.024$ & 0.15 & 0.216 & -0.15 & \textbf{100.0\%} \\
Virchow2 & 0.825 & 0.589 & 0.771 & 0.795 & $+0.025$ & 0.13 & 0.239 & -0.17 & \textbf{100.0\%} \\
H0-mini$^{\dagger}$ & 0.820 & 0.613 & 0.737 & 0.754 & $+0.017$ & 0.12 & 0.257 & -0.19 & 99.9\% \\
Virchow & 0.789 & 0.653 & 0.697 & 0.706 & $+0.009$ & 0.09 & 0.301 & -0.18 & \textbf{100.0\%} \\
H-optimus-1 & 0.878 & 0.672 & 0.775 & 0.786 & $+0.012$ & 0.09 & 0.220 & \textbf{-0.10} & 99.8\% \\
UNI2-h & 0.847 & 0.734 & 0.711 & 0.725 & $+0.013$ & 0.07 & 0.265 & -0.12 & 99.3\% \\
mSTAR$^{\dagger}$ & 0.817 & 0.687 & 0.696 & 0.703 & $+0.007$ & 0.06 & 0.298 & -0.12 & 99.9\% \\
H-optimus-0 & 0.846 & 0.691 & 0.707 & 0.711 & $+0.004$ & 0.05 & 0.287 & -0.11 & 99.9\% \\
MUSK$^{\dagger}$ & 0.724 & 0.587 & 0.674 & 0.680 & $+0.006$ & 0.05 & 0.332 & -0.14 & \textbf{100.0\%} \\
Prov-GigaPath & 0.831 & 0.666 & 0.683 & 0.679 & $-0.005$ & 0.05 & 0.308 & -0.15 & 99.9\% \\
UNI & 0.807 & 0.707 & 0.673 & 0.678 & $+0.005$ & 0.05 & 0.320 & -0.12 & 100.0\% \\
Hibou-L & 0.759 & 0.737 & 0.577 & 0.541 & $-0.037$ & 0.04 & 0.405 & -0.25 & 99.7\% \\
GPFM$^{\dagger}$ & 0.759 & 0.718 & 0.612 & 0.604 & $-0.008$ & 0.04 & 0.387 & -0.15 & 100.0\% \\
Phikon$^{\dagger}$ & 0.792 & 0.786 & 0.577 & 0.577 & $-0.000$ & 0.04 & 0.396 & -0.20 & 99.2\% \\
Hibou-B & 0.768 & 0.719 & 0.600 & 0.590 & $-0.010$ & 0.04 & 0.387 & -0.18 & 99.9\% \\
Phikon-v2$^{\dagger}$ & 0.790 & 0.779 & 0.550 & 0.544 & $-0.006$ & 0.03 & 0.409 & -0.19 & 99.4\% \\
Prost40M$^{\dagger}$ & 0.635 & 0.714 & 0.464 & 0.459 & $-0.005$ & -0.01 & 0.528 & -0.25 & \textbf{100.0\%} \\
\hline
DINOv2-B & 0.607 & 0.573 & 0.540 & 0.513 & $-0.028$ & 0.01 & 0.468 & -0.12 & 100.0\% \\
\hline
\end{tabular}
\caption{\textbf{Representation robustness on TCGA-4x4.} Results for the 20 tile-level pathology foundation models, ordered by median \code{CRoMa} ($m{=}5$), with the natural-image control \code{DINOv2-B} shown separately. All models are evaluated at the shared operating point $k{=}71$, the dataset median of the per-model biological $k^\star$. Columns: biological and confounder $k$-NN balanced accuracy; pooled \code{RI} and \code{MaRI}; $\Delta{=}\code{MaRI}-\code{RI}$; median \code{CRoMa}; $F(0)$, the fraction with $\mcode{CRoMa}\leq0$; $\mcode{LTM}_{10}$, the mean of the lowest decile; and support, the fraction of samples effectively contributing to \code{RI}/\code{MaRI}. Bold denotes the best value in each score column (conf bacc and $\Delta$ are diagnostics). $\dagger$ marks the $9$ TCGA-exposed encoders (Table~\ref{tab:model-summary}).}
\label{tab:main-results-tcga4x4}
\end{table}

\begin{table}[!htbp]
\centering
\small
\setlength{\tabcolsep}{4pt}
\begin{tabular}{lccccccccc}
\hline
Model & bio bacc & conf bacc & \code{RI} & \code{MaRI} & $\Delta$ & \code{CRoMa} & $F(0)$ & $\mcode{LTM}_{10}$ & support \\
\hline
Midnight-12k & 0.976 & 0.681 & 0.943 & 0.941 & $-0.002$ & \textbf{0.58} & 0.051 & -0.08 & 99.0\% \\
CONCH & 0.973 & 0.595 & 0.951 & 0.957 & $+0.006$ & 0.44 & 0.045 & -0.04 & 99.8\% \\
CONCHv1.5 & 0.973 & 0.597 & 0.952 & 0.964 & $+0.012$ & 0.39 & 0.043 & -0.03 & \textbf{99.9\%} \\
GenBio-PathFM & \textbf{0.981} & 0.535 & \textbf{0.960} & \textbf{0.964} & $+0.004$ & 0.39 & \textbf{0.038} & \textbf{-0.02} & \textbf{99.9\%} \\
H0-mini & 0.967 & 0.594 & 0.935 & 0.946 & $+0.011$ & 0.38 & 0.058 & -0.07 & 99.7\% \\
Virchow & 0.970 & 0.654 & 0.935 & 0.943 & $+0.008$ & 0.37 & 0.053 & -0.05 & 99.6\% \\
Virchow2 & 0.978 & 0.538 & 0.954 & 0.957 & $+0.002$ & 0.35 & 0.040 & -0.04 & 99.6\% \\
MUSK & 0.969 & 0.692 & 0.924 & 0.931 & $+0.008$ & 0.29 & 0.063 & -0.07 & 99.2\% \\
H-optimus-1 & 0.977 & 0.631 & 0.940 & 0.949 & $+0.008$ & 0.26 & 0.049 & -0.04 & 99.1\% \\
GPFM & 0.969 & 0.805 & 0.883 & 0.902 & $+0.018$ & 0.24 & 0.095 & -0.10 & 97.6\% \\
H-optimus-0 & 0.971 & 0.689 & 0.911 & 0.919 & $+0.007$ & 0.23 & 0.076 & -0.08 & 98.5\% \\
UNI2-h & 0.976 & 0.729 & 0.916 & 0.927 & $+0.012$ & 0.22 & 0.064 & -0.06 & 97.7\% \\
mSTAR & 0.970 & 0.803 & 0.881 & 0.898 & $+0.017$ & 0.19 & 0.087 & -0.09 & 97.7\% \\
Phikon & 0.962 & 0.877 & 0.772 & 0.782 & $+0.010$ & 0.17 & 0.179 & -0.19 & 81.5\% \\
UNI & 0.973 & 0.817 & 0.880 & 0.891 & $+0.011$ & 0.17 & 0.086 & -0.08 & 95.3\% \\
Hibou-B & 0.967 & 0.901 & 0.774 & 0.775 & $+0.000$ & 0.13 & 0.188 & -0.17 & 85.7\% \\
Prov-GigaPath & 0.962 & 0.914 & 0.707 & 0.746 & $+0.039$ & 0.13 & 0.236 & -0.16 & 79.1\% \\
Prost40M & 0.911 & 0.800 & 0.701 & 0.721 & $+0.019$ & 0.13 & 0.277 & -0.24 & 96.2\% \\
Phikon-v2 & 0.956 & 0.870 & 0.741 & 0.746 & $+0.005$ & 0.12 & 0.225 & -0.17 & 83.3\% \\
Hibou-L & 0.955 & 0.941 & 0.624 & 0.586 & $-0.038$ & 0.11 & 0.315 & -0.29 & 67.0\% \\
\hline
DINOv2-B & 0.905 & 0.492 & 0.876 & 0.867 & $-0.008$ & 0.18 & 0.099 & -0.07 & 100.0\% \\
\hline
\end{tabular}
\caption{\textbf{Representation robustness on Tolkach-ESCA.} Results for the 20 tile-level pathology foundation models, ordered by median \code{CRoMa} ($m{=}5$), with the natural-image control \code{DINOv2-B} shown separately. All models are evaluated at the shared operating point $k{=}61$, the dataset median of the per-model biological $k^\star$. Columns: biological and confounder $k$-NN balanced accuracy; pooled \code{RI} and \code{MaRI}; $\Delta{=}\code{MaRI}-\code{RI}$; median \code{CRoMa}; $F(0)$, the fraction with $\mcode{CRoMa}\leq0$; $\mcode{LTM}_{10}$, the mean of the lowest decile; and support, the fraction of samples effectively contributing to \code{RI}/\code{MaRI}. Bold denotes the best value in each score column (conf bacc and $\Delta$ are diagnostics).}
\label{tab:main-results-tolkach}
\end{table}

\clearpage

\subsection{\code{CRoMa} distributions}
\label{supp:distributions}

% AUTO-GENERATED by scripts/repro/generate_distribution_floats.py -- do not edit by hand.
% The PDF is drawn by scripts/repro/figures/croma_distribution_figure.py, into
% output/metrics/median-k/pathorob-tcga-4x4/studies/plots/pdf/croma_distribution.pdf.
% It is NOT staged: copy that file to paper/figures/results/pathorob-tcga-4x4/pdf/croma_distribution.pdf to resolve the
% \includegraphics below (\graphicspath reaches paper/figures/, not output/).
\begin{figure}[!htbp]
\centering
\includegraphics[width=0.78\linewidth]{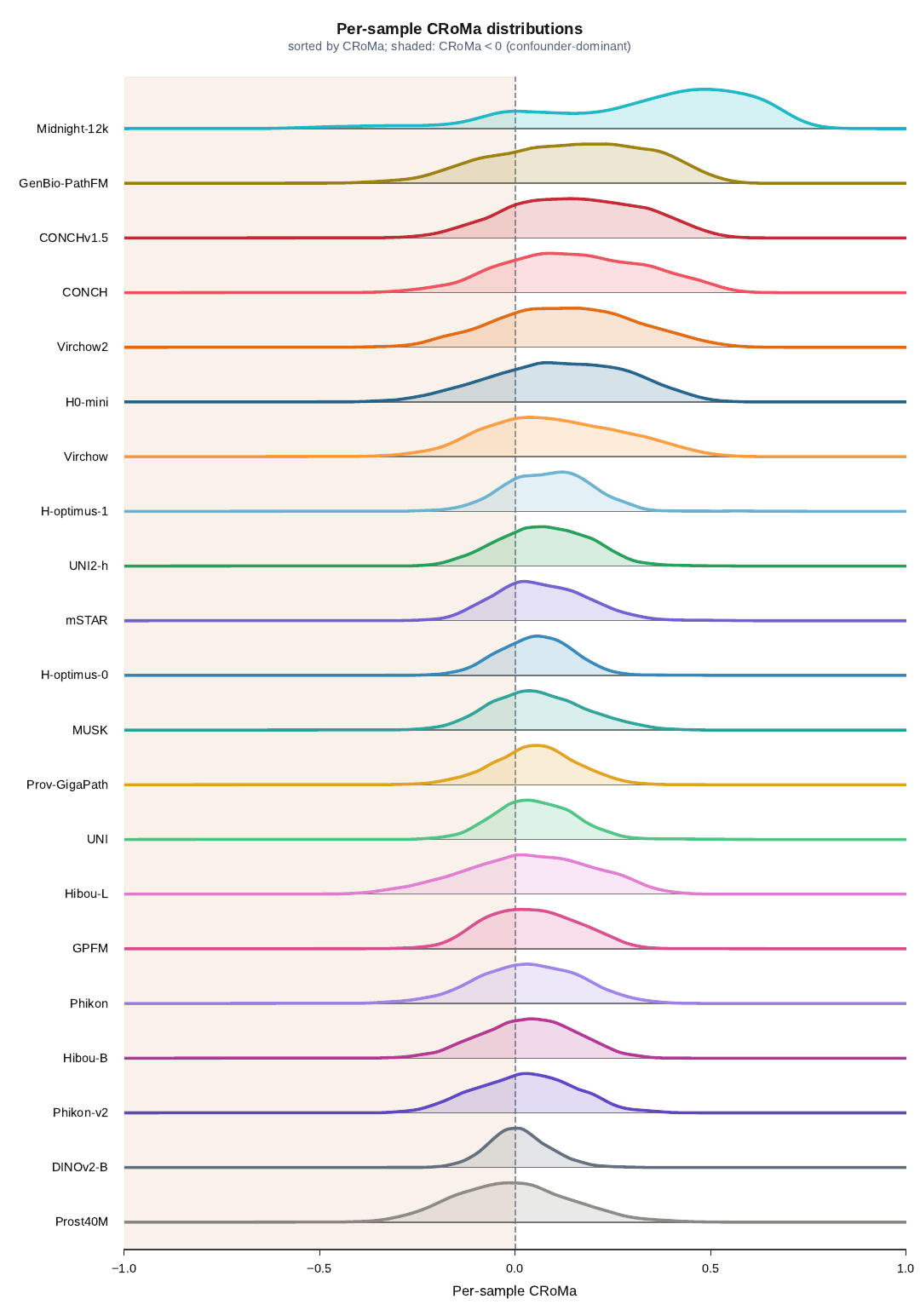}
\caption{\textbf{Per-sample \code{CRoMa} distributions on TCGA-4x4.} Ridgeline distributions for \TcgaFourByFourRankedNModels{} pathology encoders and the natural-image control (\code{DINOv2-B}), ordered by pooled median. The dashed line denotes $\mcode{CRoMa}=0$; shading denotes $\mcode{CRoMa}<0$. Per-model $F(0)$ and $\mcode{LTM}_{10}$ are reported in Table~\ref{tab:main-results-tcga4x4}.}
\label{fig:croma-distribution-tcga4x4}
\end{figure}
% AUTO-GENERATED by scripts/repro/generate_distribution_floats.py -- do not edit by hand.
% The PDF is drawn by scripts/repro/figures/croma_distribution_figure.py, into
% output/metrics/median-k/pathorob-tolkach-esca/studies/plots/pdf/croma_distribution.pdf.
% It is NOT staged: copy that file to paper/figures/results/pathorob-tolkach-esca/pdf/croma_distribution.pdf to resolve the
% \includegraphics below (\graphicspath reaches paper/figures/, not output/).
\begin{figure}[!htbp]
\centering
\includegraphics[width=0.78\linewidth]{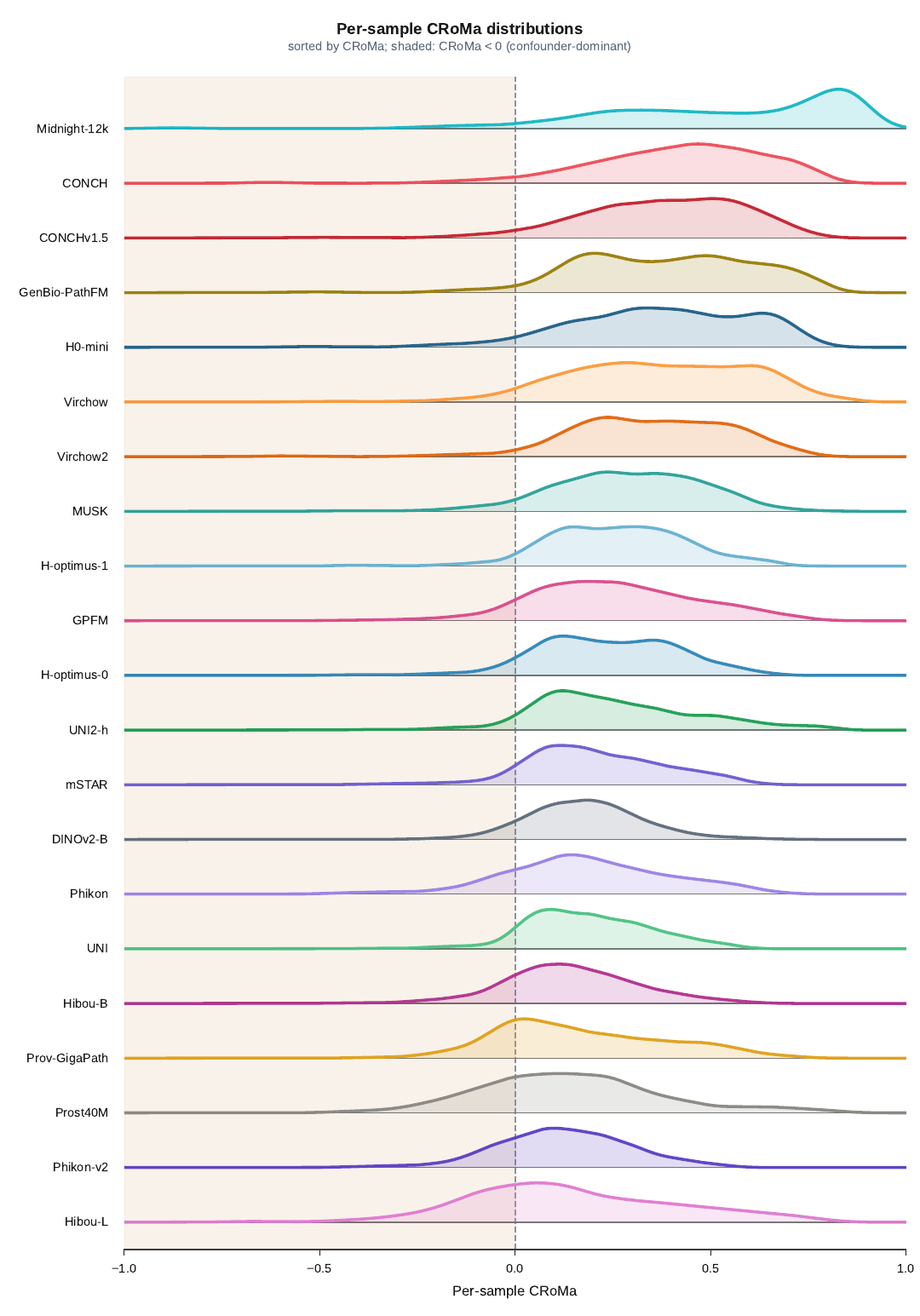}
\caption{\textbf{Per-sample \code{CRoMa} distributions on Tolkach-ESCA.} Ridgeline distributions for \TolkachRankedNModels{} pathology encoders and the natural-image control (\code{DINOv2-B}), ordered by pooled median. The dashed line denotes $\mcode{CRoMa}=0$; shading denotes $\mcode{CRoMa}<0$. Per-model $F(0)$ and $\mcode{LTM}_{10}$ are reported in Table~\ref{tab:main-results-tolkach}.}
\label{fig:croma-distribution-tolkach}
\end{figure}

\clearpage

\subsection{Downstream shortcut susceptibility}
\label{supp:apd-benchmarks}

% AUTO-GENERATED by scripts/repro/generate_apd_floats.py -- do not edit by hand.
% PDFs are rendered under output/studies/apd/plots/pdf/croma_nipd_<benchmark>.pdf.
\begin{figure}[!htbp]
\centering
\includegraphics[width=0.92\linewidth]{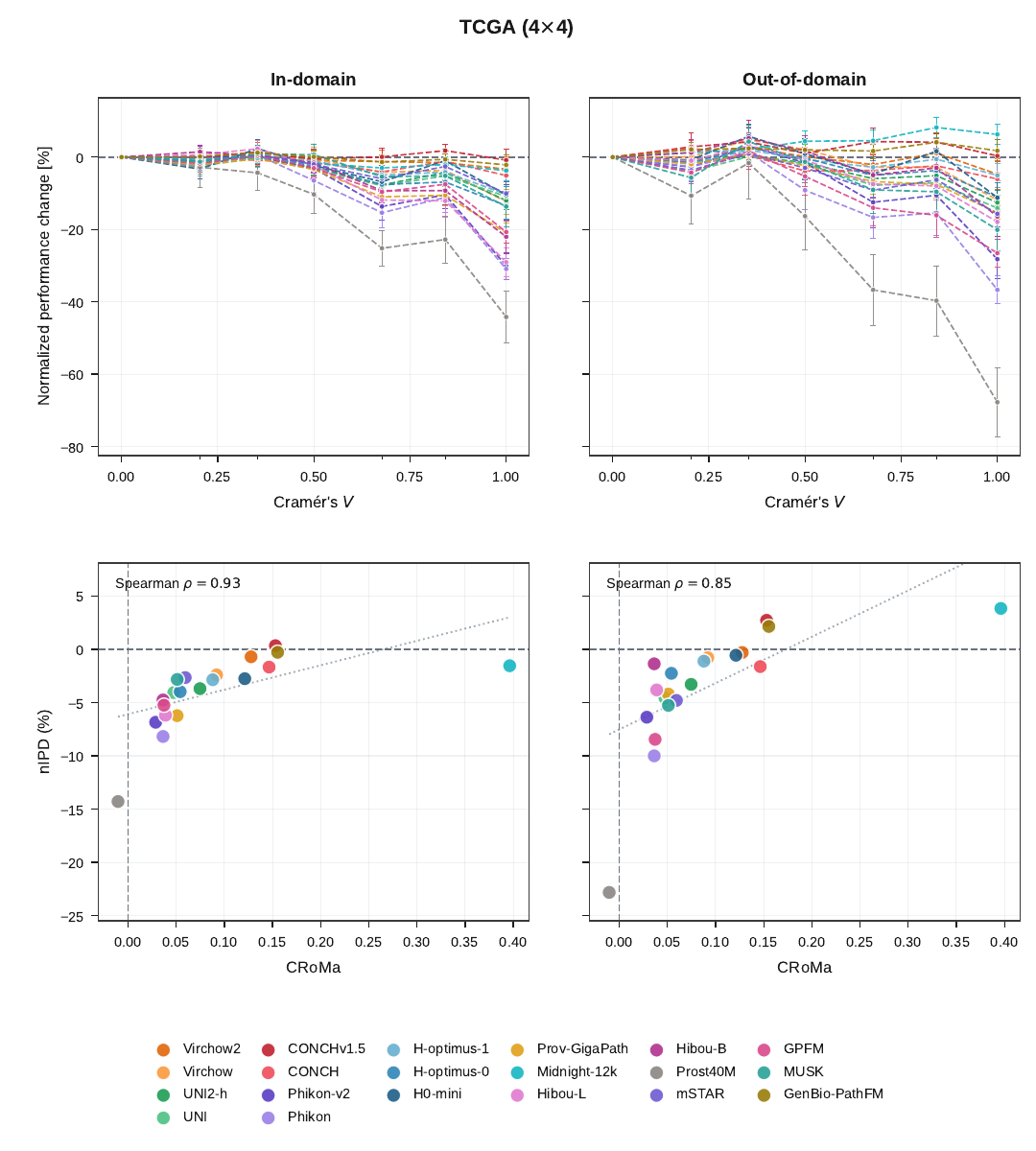}
\caption{\textbf{Representation robustness and downstream shortcut susceptibility on TCGA-4x4.} \code{CRoMa} measures robustness in frozen representation space: higher values indicate more biology-dominant geometry. \code{nIPD} summarises downstream linear-probe degradation as the training-set confounder--biology correlation increases: lower values indicate greater susceptibility. Columns show evaluation on acquisition groups represented during training (left, in domain) and unseen acquisition groups (right, out of domain). \textbf{Top row:} normalised performance change $g(V)$ relative to the above-chance performance at $V=0$ as the training-set confounder--biology correlation increases to $V{=}1$. Curves denote the $20$ tile encoders and error bars show 95\% Student's $t$ confidence intervals for the mean normalised change across 20 resampling runs. \textbf{Bottom row:} median \code{CRoMa}$(m{=}5)$ versus \code{nIPD}, one point per encoder. Dashed lines mark $\mcode{CRoMa}=0$ and $\mcode{nIPD}=0$. Corresponding \code{RI}, \code{MaRI}, and legacy \code{APD} correlations are reported in Table~\ref{tab:apd-correlation}.}
\label{fig:croma-nipd-tcga4x4}
\end{figure}
\begin{figure}[!htbp]
\centering
\includegraphics[width=0.92\linewidth]{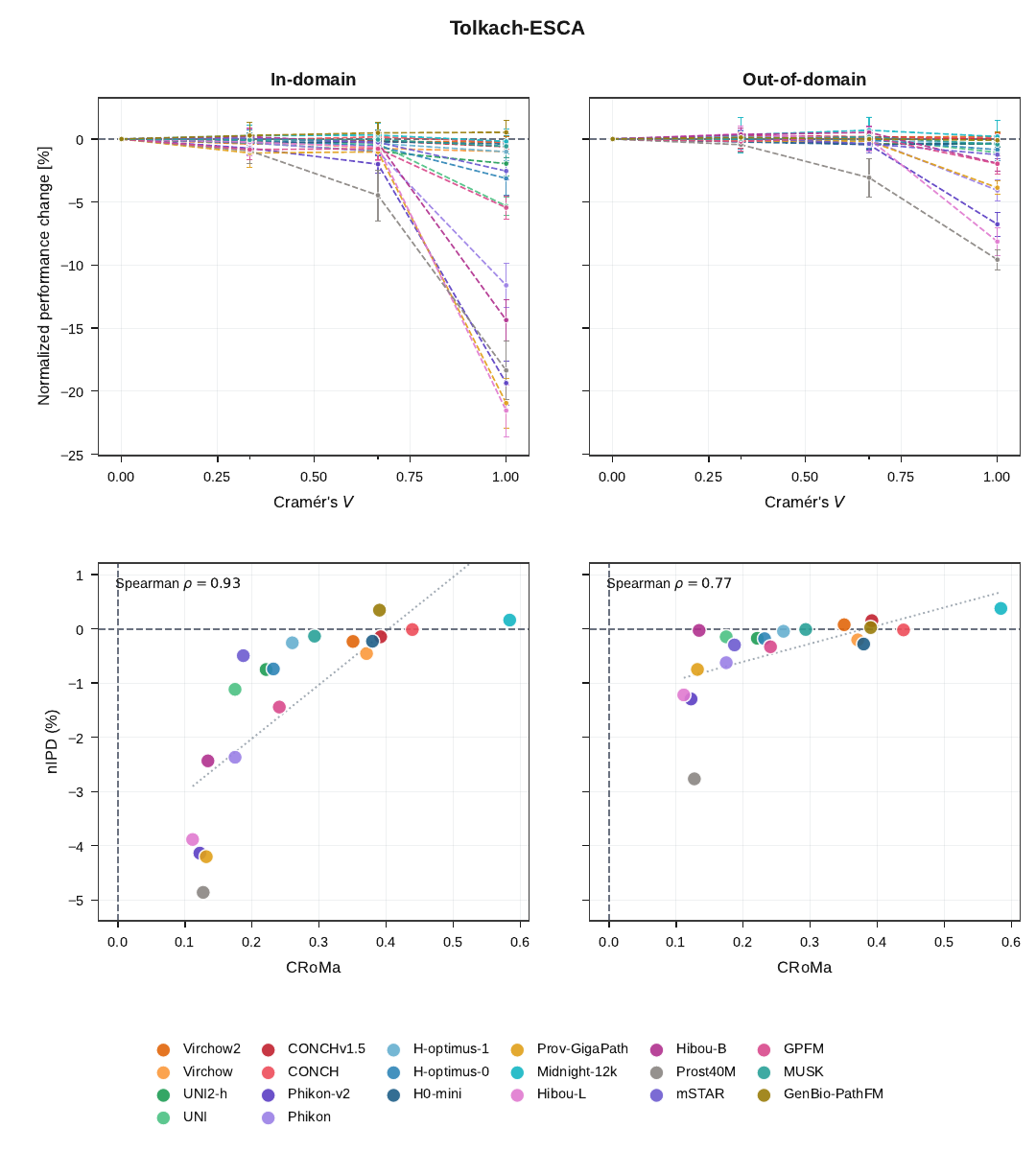}
\caption{\textbf{Representation robustness and downstream shortcut susceptibility on Tolkach-ESCA.} \code{CRoMa} measures robustness in frozen representation space: higher values indicate more biology-dominant geometry. \code{nIPD} summarises downstream linear-probe degradation as the training-set confounder--biology correlation increases: lower values indicate greater susceptibility. Columns show evaluation on acquisition groups represented during training (left, in domain) and unseen acquisition groups (right, out of domain). \textbf{Top row:} normalised performance change $g(V)$ relative to the above-chance performance at $V=0$ as the training-set confounder--biology correlation increases to $V{=}1$. Curves denote the $20$ tile encoders and error bars show 95\% Student's $t$ confidence intervals for the mean normalised change across 20 resampling runs. \textbf{Bottom row:} median \code{CRoMa}$(m{=}5)$ versus \code{nIPD}, one point per encoder. Dashed lines mark $\mcode{CRoMa}=0$ and $\mcode{nIPD}=0$. Corresponding \code{RI}, \code{MaRI}, and legacy \code{APD} correlations are reported in Table~\ref{tab:apd-correlation}.}
\label{fig:croma-nipd-tolkach}
\end{figure}

% AUTO-GENERATED by scripts/repro/generate_apd_floats.py -- do not edit by hand.
\begin{table}[!htbp]
\centering
\small
\begin{tabular}{llccc}
\hline
 & Metric & Camelyon & TCGA-4x4 & Tolkach-ESCA \\
\hline
\multirow{3}{*}{\code{nIPD}$_\mathrm{ID}$} & \textbf{\code{CRoMa}} &  \textbf{\NipdIdCromaCamelyon} &  \textbf{\NipdIdCromaTcga} &  \textbf{\NipdIdCromaTolkach} \\
 & \code{RI} & \NipdIdRiCamelyon & \NipdIdRiTcga & \NipdIdRiTolkach \\
 & \code{MaRI} & \NipdIdMariCamelyon & \NipdIdMariTcga & \NipdIdMariTolkach \\
\hline
\multirow{3}{*}{\code{nIPD}$_\mathrm{OOD}$} & \textbf{\code{CRoMa}} & \NipdOodCromaCamelyon & \NipdOodCromaTcga & \NipdOodCromaTolkach \\
 & \code{RI} & \NipdOodRiCamelyon & \NipdOodRiTcga & \NipdOodRiTolkach \\
 & \code{MaRI} & \NipdOodMariCamelyon & \NipdOodMariTcga & \NipdOodMariTolkach \\
\hline
\code{APD}$_\mathrm{ID}$ & \textbf{\code{CRoMa}} & \ApdIdCromaCamelyon & \ApdIdCromaTcga & \ApdIdCromaTolkach \\
\code{APD}$_\mathrm{OOD}$ & \textbf{\code{CRoMa}} & \ApdOodCromaCamelyon & \ApdOodCromaTcga & \ApdOodCromaTolkach \\
\hline
\end{tabular}
\caption{\textbf{Per-benchmark association between representation robustness and downstream shortcut susceptibility.} Spearman $\rho$ relates each representation metric to normalised integrated performance degradation (\code{nIPD}), the primary downstream reduction. Positive values indicate that more robust representations undergo less degradation. In-domain evaluation uses acquisition groups represented during training and more directly isolates susceptibility to the induced shortcut. Out-of-domain evaluation uses unseen acquisition groups and therefore also reflects distribution shift. Each benchmark contains \ApdNModels{} ranked tile encoders. The final two rows repeat the \code{CRoMa} analysis using PathoROB's original \code{APD} for continuity. PCaBiop contains only four whole-slide encoders, so we excluded it from the correlation table.}
\label{tab:apd-correlation}
\end{table}

\clearpage

\subsection{Median--tail Pareto frontiers}
\label{supp:pareto}

% AUTO-GENERATED by scripts/repro/generate_pareto_float.py -- do not edit by hand.
% The PDF is drawn by scripts/repro/figures/croma_pareto_figure.py, into
% output/metrics/median-k/pathorob-tcga-4x4/studies/plots/pdf/croma_pareto.pdf.
% It is NOT staged: copy that file to paper/figures/results/pathorob-tcga-4x4/pdf/croma_pareto.pdf to resolve the
% \includegraphics below (\graphicspath reaches paper/figures/, not output/).
\begin{figure}[!htbp]
\centering
\includegraphics[width=0.7\linewidth]{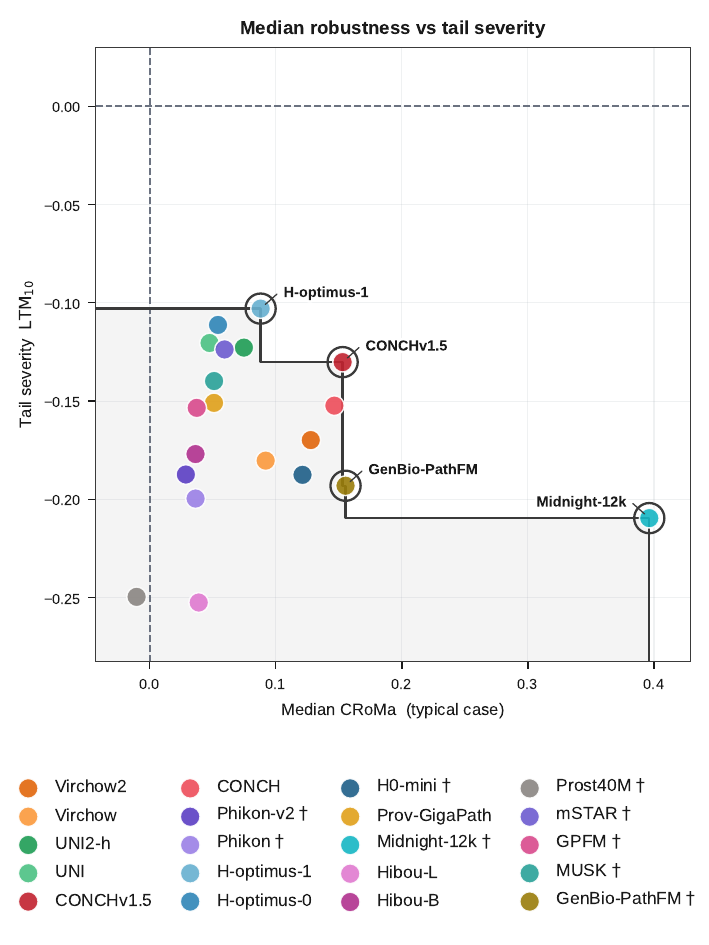}
\caption{\textbf{Median and lower-tail \code{CRoMa} on TCGA-4x4.} Pooled median \code{CRoMa} against worst-decile mean $\mcode{LTM}_{10}$ for the \TcgaFourByFourRankedNModels{} pathology encoders. Higher values are preferable on both axes. Ringed points form the upper-right Pareto frontier, while shaded points are dominated on both axes. Per-model $F(0)$ and $\mcode{LTM}_{10}$ are reported in Table~\ref{tab:main-results-tcga4x4}. $\dagger$ marks the 9 TCGA-exposed encoders.}
\label{fig:croma-pareto-tcga4x4}
\end{figure}
% AUTO-GENERATED by scripts/repro/generate_pareto_float.py -- do not edit by hand.
% The PDF is drawn by scripts/repro/figures/croma_pareto_figure.py, into
% output/metrics/median-k/pathorob-tolkach-esca/studies/plots/pdf/croma_pareto.pdf.
% It is NOT staged: copy that file to paper/figures/results/pathorob-tolkach-esca/pdf/croma_pareto.pdf to resolve the
% \includegraphics below (\graphicspath reaches paper/figures/, not output/).
\begin{figure}[!htbp]
\centering
\includegraphics[width=0.7\linewidth]{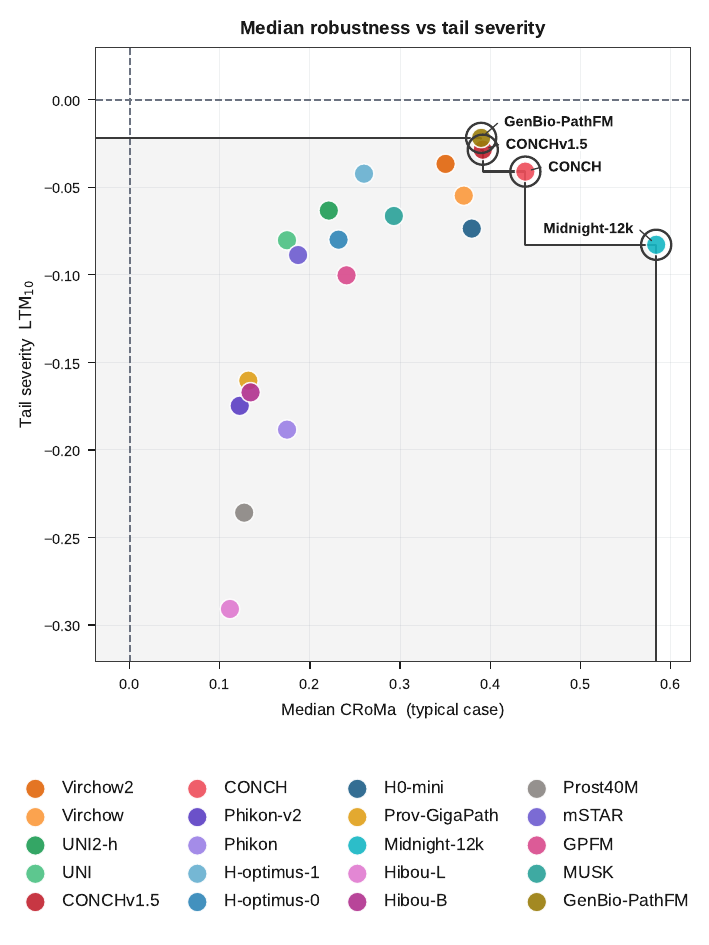}
\caption{\textbf{Median and lower-tail \code{CRoMa} on Tolkach-ESCA.} Pooled median \code{CRoMa} against worst-decile mean $\mcode{LTM}_{10}$ for the \TolkachRankedNModels{} pathology encoders. Higher values are preferable on both axes. Ringed points form the upper-right Pareto frontier, while shaded points are dominated on both axes. Per-model $F(0)$ and $\mcode{LTM}_{10}$ are reported in Table~\ref{tab:main-results-tolkach}.}
\label{fig:croma-pareto-tolkach}
\end{figure}

\clearpage

% Bootstrap-CI supplement: to restore it, uncomment the \input below AND set
% with_ci=True on the entries in scripts/repro/paper_manifest.py -- with_ci makes the
% results captions cite tab:bootstrap-uncertainty, which only this \input defines.
% \input{sections/supp/bootstrap_uncertainty}
\subsection{TCGA: the paired 2x2 configuration}
\label{supp:tcga}

\begin{figure}[!htbp]
\centering
\includegraphics[width=0.6\linewidth]{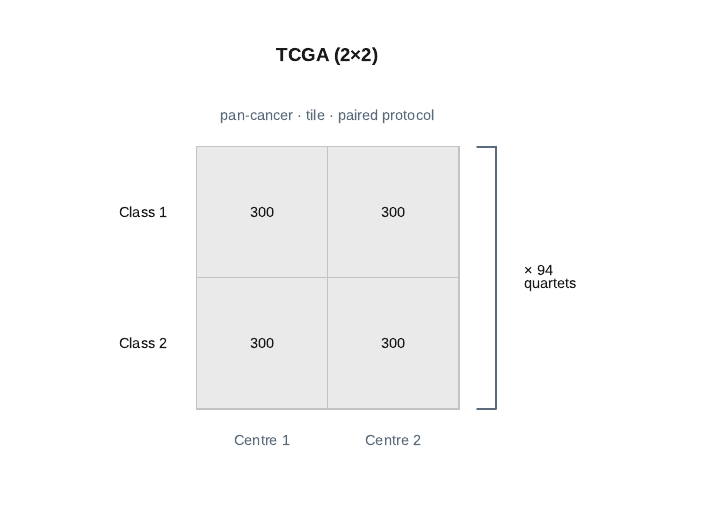}
\caption{\textbf{The paired TCGA protocol comprises 94 balanced 2x2 quartets.} The grid shows one generic quartet: two biological classes by two medical centres, with $300$ tile occurrences per cell. Class and centre identities vary across quartets, spanning $21$ cancer types and $26$ centres. Typed-neighbour counts are pooled across quartets before scoring, for $112{,}800$ tile occurrences in total. The dataset-wide TCGA-4x4 design is shown in Figure~\ref{fig:dataset-cardinality}.}
\label{fig:tcga2x2-cardinality}
\end{figure}

For completeness, we additionally report PathoROB's paired
TCGA-2x2 protocol here (Table~\ref{tab:main-results-tcga}; composition in
Figure~\ref{fig:tcga2x2-cardinality}): $94$ balanced quartets,
each two biological classes $\times$ two centres, with neighbour-type statistics computed
within each quartet and pooled at the level of \textbf{SO}/\textbf{OS} counts before scoring.
Results on this paired configuration align with the observations made on the TCGA-4x4 benchmark: biology-dominant pooled scores and a
confounder-dominant lower tail for every model, with \code{Midnight-12k} leading
(\code{CRoMa}~$0.42$ on TCGA-2x2 and $0.40$ on TCGA-4x4).

\begin{table}[!htbp]
\centering
\small
\setlength{\tabcolsep}{4pt}
\begin{tabular}{lccccccccc}
\hline
Model & bio bacc & conf bacc & \code{RI} & \code{MaRI} & $\Delta$ & \code{CRoMa} & $F(0)$ & $\mcode{LTM}_{10}$ & support \\
\hline
Midnight-12k$^{\dagger}$ & 0.934 & 0.707 & \textbf{0.858} & \textbf{0.890} & $+0.032$ & \textbf{0.42} & \textbf{0.124} & -0.20 & 93.3\% \\
GenBio-PathFM$^{\dagger}$ & 0.943 & 0.761 & 0.840 & 0.859 & $+0.020$ & 0.27 & 0.148 & -0.15 & 92.4\% \\
CONCHv1.5 & 0.921 & 0.692 & 0.832 & 0.855 & $+0.023$ & 0.26 & 0.144 & -0.13 & 98.9\% \\
CONCH & 0.912 & 0.668 & 0.824 & 0.849 & $+0.025$ & 0.25 & 0.154 & -0.15 & \textbf{99.1\%} \\
Virchow2 & 0.928 & 0.715 & 0.822 & 0.841 & $+0.019$ & 0.24 & 0.157 & -0.15 & 97.6\% \\
H0-mini$^{\dagger}$ & 0.925 & 0.741 & 0.792 & 0.810 & $+0.018$ & 0.19 & 0.194 & -0.17 & 95.6\% \\
Virchow & 0.909 & 0.748 & 0.761 & 0.775 & $+0.014$ & 0.17 & 0.235 & -0.26 & 95.1\% \\
H-optimus-1 & \textbf{0.949} & 0.761 & 0.844 & 0.869 & $+0.026$ & 0.17 & 0.144 & \textbf{-0.09} & 95.1\% \\
UNI2-h & 0.943 & 0.815 & 0.796 & 0.821 & $+0.025$ & 0.13 & 0.182 & -0.11 & 91.8\% \\
H-optimus-0 & 0.938 & 0.777 & 0.794 & 0.819 & $+0.026$ & 0.11 & 0.193 & -0.11 & 94.3\% \\
mSTAR$^{\dagger}$ & 0.929 & 0.798 & 0.765 & 0.778 & $+0.013$ & 0.11 & 0.213 & -0.13 & 95.4\% \\
MUSK$^{\dagger}$ & 0.887 & 0.754 & 0.718 & 0.726 & $+0.008$ & 0.10 & 0.248 & -0.17 & 98.2\% \\
Prov-GigaPath & 0.925 & 0.820 & 0.728 & 0.753 & $+0.025$ & 0.10 & 0.249 & -0.14 & 92.6\% \\
UNI & 0.928 & 0.833 & 0.729 & 0.744 & $+0.015$ & 0.09 & 0.243 & -0.12 & 92.4\% \\
Phikon$^{\dagger}$ & 0.913 & 0.884 & 0.600 & 0.604 & $+0.004$ & 0.06 & 0.360 & -0.22 & 81.8\% \\
GPFM$^{\dagger}$ & 0.899 & 0.844 & 0.645 & 0.643 & $-0.002$ & 0.06 & 0.318 & -0.18 & 94.9\% \\
Hibou-B & 0.909 & 0.867 & 0.607 & 0.605 & $-0.002$ & 0.06 & 0.361 & -0.22 & 86.1\% \\
Hibou-L & 0.904 & 0.881 & 0.559 & 0.528 & $-0.031$ & 0.05 & 0.422 & -0.38 & 78.5\% \\
Phikon-v2$^{\dagger}$ & 0.918 & 0.887 & 0.587 & 0.588 & $+0.001$ & 0.05 & 0.382 & -0.22 & 81.6\% \\
Prost40M$^{\dagger}$ & 0.839 & 0.847 & 0.508 & 0.500 & $-0.008$ & 0.01 & 0.463 & -0.34 & 92.5\% \\
\hline
DINOv2-B & 0.802 & 0.696 & 0.612 & 0.604 & $-0.008$ & 0.04 & 0.350 & -0.16 & 99.9\% \\
\hline
\end{tabular}
\caption{\textbf{Representation robustness on TCGA-2x2.} Pooled results for 20 tile-level pathology foundation models, ordered by \code{CRoMa} ($m{=}5$), with the natural-image control \code{DINOv2-B} shown separately beneath. All models are evaluated at the shared operating point $k{=}61$, the dataset median of the per-model biological $k^\star$. Columns: biological and confounder $k$-NN balanced accuracy; pooled \code{RI} and \code{MaRI}; $\Delta{=}\code{MaRI}-\code{RI}$; median \code{CRoMa}; $F(0)$, the fraction with $\mcode{CRoMa}\leq0$; $\mcode{LTM}_{10}$, the mean of the lowest decile; and support, the fraction of samples effectively contributing to \code{RI}/\code{MaRI}. Bold denotes the best value in each score column (conf bacc and $\Delta$ are diagnostics). $\dagger$ marks the $9$ TCGA-exposed encoders (Table~\ref{tab:model-summary}).}
\label{tab:main-results-tcga}
\end{table}

\subsection{Robustness of slide-level representations}
\label{supp:panda}

Unlike the tile-level panels, the four slide-level encoders were evaluated at their model-specific, biologically selected $k^\star$. With only four models, the median (the lower of the two central optima) is highly sensitive to panel composition and collapses to $k=3$ on PCaBiop, reducing mean \code{RI}/\code{MaRI} support across the four encoders from $36.9\%$ at per-model $k^\star$ to $27.0\%$. We therefore treat the slide-level panel as an explicit exception to the shared median-$k$ protocol. This preserves each encoder's biologically selected neighbourhood scale, although comparisons of \code{RI}, \code{MaRI} and support remain conditional on model-specific $k$. \code{CRoMa}, $F(0)$ and $\mcode{LTM}_{10}$ are $k$-independent and are unaffected by this choice.

\begin{figure}[!htbp]
\centering
\includegraphics[width=0.78\linewidth]{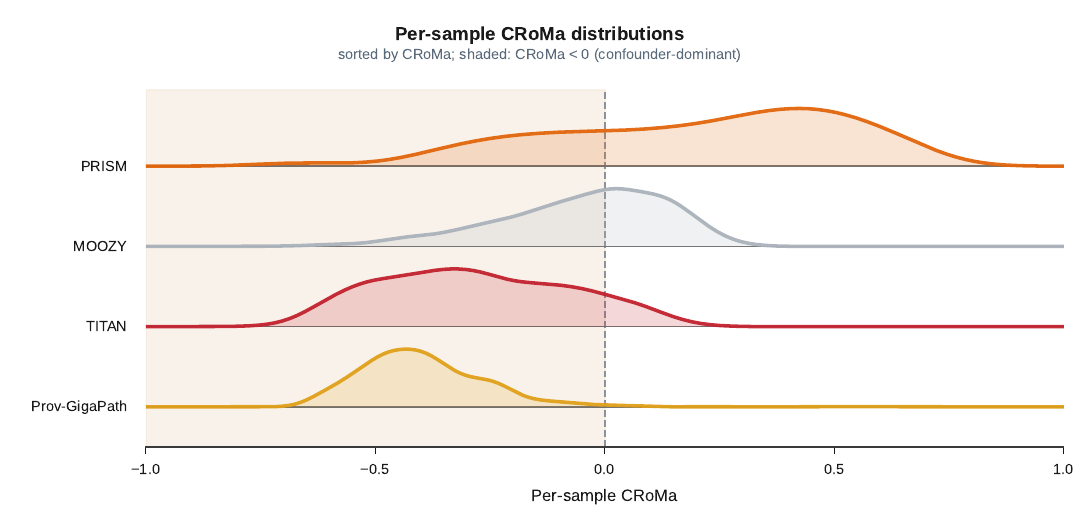}
\caption{\textbf{\code{CRoMa} distributions on PCaBiop.} Ridgeline distributions for the \PandaRankedNModels{} slide-level encoders on PCaBiop, ordered by pooled median. The dashed line denotes $\mcode{CRoMa}=0$. Shading denotes $\mcode{CRoMa}<0$. Per-model $F(0)$ and $\mcode{LTM}_{10}$ are reported in Table~\ref{tab:main-results-panda}.}
\label{fig:croma-distribution-panda}
\end{figure}

\paragraph{Binary cancer detection.} On PCaBiop, the contributing centre was near-perfectly decodable for all four encoders (confounder $k$-NN balanced accuracy, $0.987$--$1.000$), whereas biological balanced accuracy ranged from $0.716$ to $0.971$ (Supplementary Table~\ref{tab:main-results-panda}). Model-level median \code{CRoMa} values spanned \PandaCromaSpan{}, with only \code{PRISM} exhibiting biology-dominant neighbourhood geometry (\PandaBestCroma{}). \code{MOOZY}~\cite{moozy} attained the highest biological balanced accuracy ($0.971$), but its median margin remained slightly negative ($-0.02$), illustrating that high biological discriminability does not imply robustness to confounder. The per-sample distributions provided complementary information (Supplementary Figure~\ref{fig:croma-distribution-panda}). \code{PRISM} and \code{MOOZY} differed in median margin (\PandaBestCroma{} versus $-0.02$) and non-positive fraction ($28.8\%$ versus $53.5\%$), but had similar worst-decile severity ($\mcode{LTM}_{10}=-0.39$ versus $-0.41$). Whereas the Camelyon analysis distinguished encoders with similar medians but different lower tails, this comparison holds lower-tail severity approximately constant while median robustness and the non-positive fraction differ, further illustrating that the three summaries are non-redundant.

% AUTO-GENERATED by scripts/repro/generate_distribution_floats.py -- do not edit by hand.
% The PDF is drawn by scripts/repro/figures/croma_distribution_figure.py, into
% output/metrics/k-star/panda/studies/plots/pdf/croma_distribution.pdf.
% It is NOT staged: copy that file to paper/figures/results/panda/pdf/croma_distribution.pdf to resolve the
% \includegraphics below (\graphicspath reaches paper/figures/, not output/).

\begin{figure}[!htbp]
\centering
\includegraphics[width=0.92\linewidth]{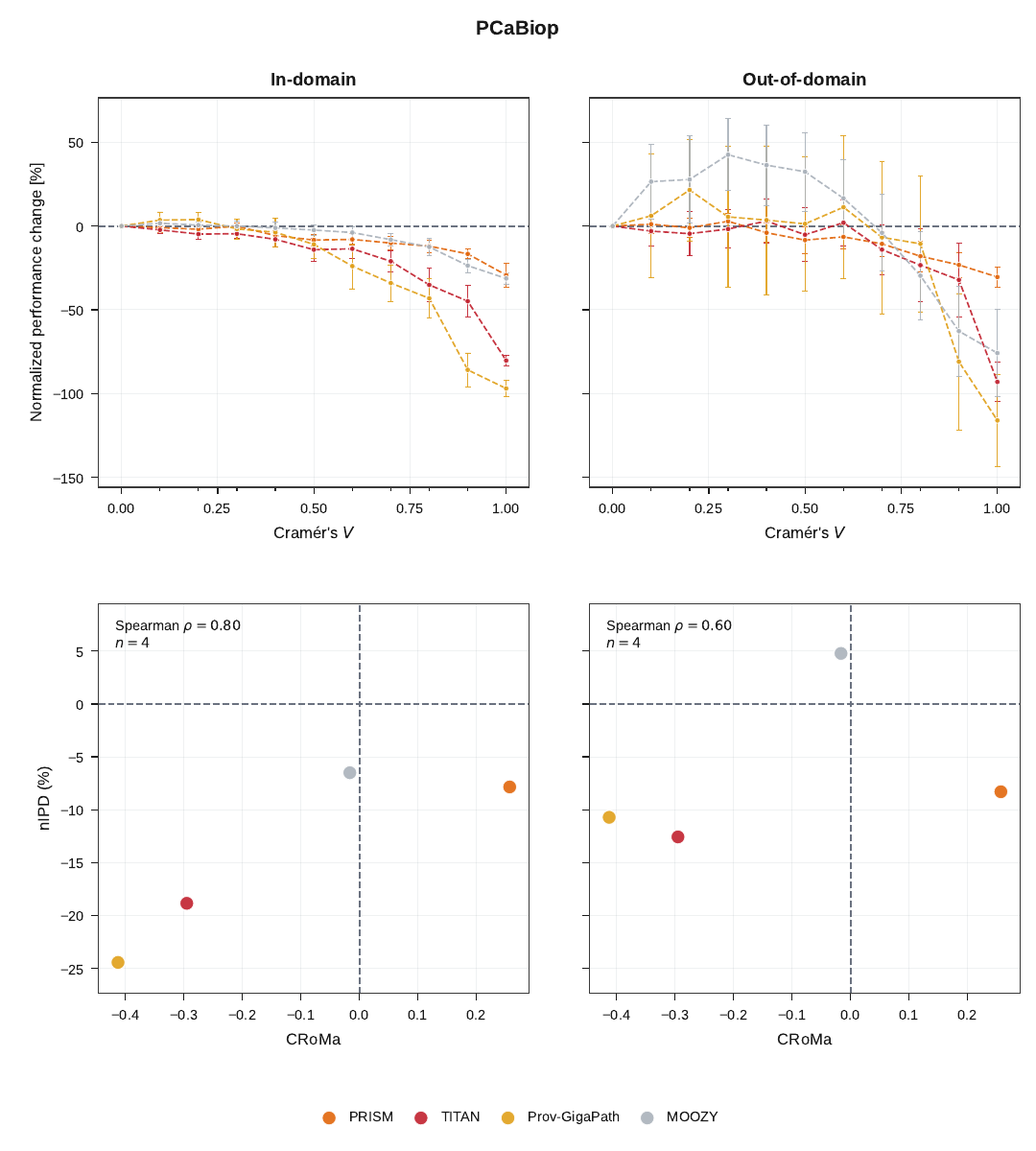}
\caption{\textbf{Representation robustness and downstream shortcut susceptibility on PCaBiop.} \code{CRoMa} measures robustness in frozen representation space: higher values indicate more biology-dominant geometry. \code{nIPD} summarises downstream linear-probe degradation as the training-set confounder--biology correlation increases: lower values indicate greater susceptibility. Columns show in-domain evaluation on PANDA acquisition groups represented during training (left) and out-of-domain evaluation on the held-out PAR cohort (right). \textbf{Top row:} normalised performance change $g(V)$ relative to the above-chance performance at $V=0$ as the training-set confounder--biology correlation increases to $V=1$. Curves denote the four whole-slide encoders and error bars show 95\% Student's $t$ confidence intervals for the mean normalised change across 20 resampling runs. \textbf{Bottom row:} median \code{CRoMa}$(m{=}5)$ versus \code{nIPD}, one point per encoder. Dashed lines mark $\mcode{CRoMa}=0$ and $\mcode{nIPD}=0$. Spearman coefficients are descriptive because the panel contains only $n=4$ encoders.}
\label{fig:croma-nipd-pcabiop}
\end{figure}

\noindent
The representation-level ordering broadly anticipated downstream shortcut susceptibility (Supplementary Figure~\ref{fig:croma-nipd-pcabiop}). Higher \code{CRoMa} tracked higher \code{nIPD}, and thus less degradation, both in domain (Spearman $\rho=\NipdIdCromaPcabiop{}$) and on the held-out PAR cohort ($\rho=\NipdOodCromaPcabiop{}$). Because the slide-level panel contains only four encoders, these associations are descriptive.

% AUTO-GENERATED by scripts/repro/generate_apd_floats.py -- do not edit by hand.

\begin{figure}[!htbp]
\centering
\includegraphics[height=0.42\textheight]{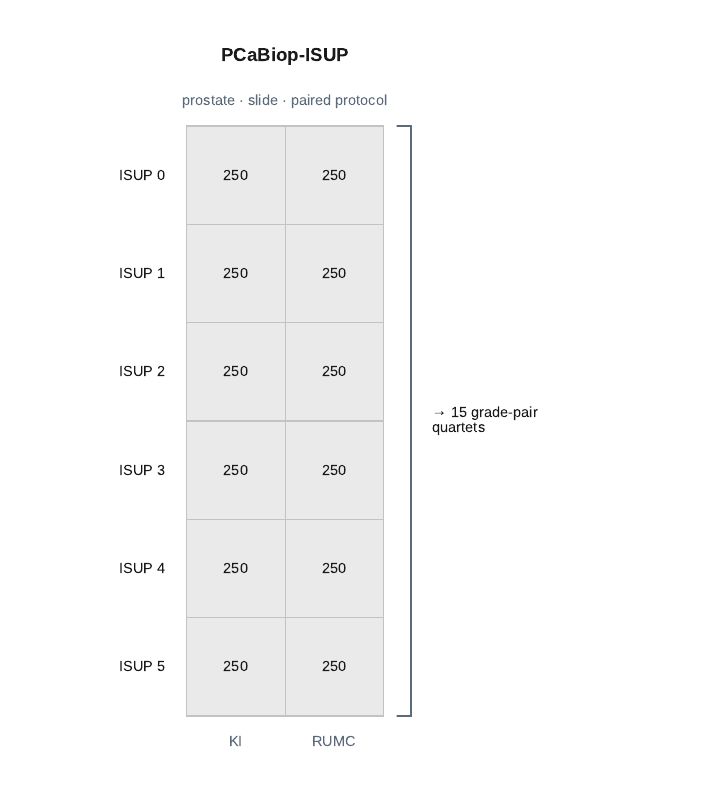}
\caption{\textbf{PCaBiop-ISUP is balanced across ISUP grade and contributing centre.} Rows denote ISUP grade groups (grade $0$ is benign, grades $1$--$5$ increasingly aggressive prostate cancer) and columns the acquisition centre. Cells show the evaluated biopsies per ISUP grade$\times$centre combination: $250$ slides in each of the $6\times2$ cells, for $3{,}000$ slides in total.}
\label{fig:pcabiop-isup-cardinality}
\end{figure}

\paragraph{ISUP grading.} To assess a more granular biological label, we curated PCaBiop-ISUP, which comprises $3{,}000$ prostate biopsies. Each biopsy comes with its ISUP grade group: grade $0$ denotes benign samples and grades $1$--$5$ denote increasingly aggressive prostate cancer. Whole slides are still sampled from PANDA, so the confounder remains the medical centre that provided the biopsy (KI versus RUMC). Each ISUP$\times$centre cell has exactly $250$ biopsies (Supplementary Figure~\ref{fig:pcabiop-isup-cardinality}). Cell balance alone does not, however, balance the typed candidate pools in the full $6\times2$ cohort. For any anchor, the \textbf{SO} pool contains $250$ slides of the same grade from the other centre, whereas the \textbf{OS} pool contains $1{,}250$ slides from the five other grades at the same centre. This asymmetry would structurally favour \textbf{OS} evidence in both fixed-$k$ and nearest-neighbour scores. We therefore decomposed the six grades into all $\binom{6}{2}=15$ grade pairs. Crossing each pair with the two centres yielded a balanced quartet of $1{,}000$ slides, with $250$ \textbf{SO} and $250$ \textbf{OS} candidates available to every anchor. \code{RI} and \code{MaRI} pooled typed-neighbour evidence across occurrences. To weight each grade contrast equally, the headline \code{CRoMa} was the median of the $15$ pair-specific medians. $F(0)$ and $\mcode{LTM}_{10}$ retained their occurrence-level interpretations and were computed over all occurrence-level margins, with $m=5$ throughout. PCaBiop is a matched $1{,}000$-slide subset of the broader $3{,}000$-slide PCaBiop-ISUP cohort. It contains both complete grade-$0$ cells as the benign class and, for the cancer class, $250$ slides total per provider sampled across grades $1$--$5$. The analyses differ in label resolution, sample composition and evaluation design. They therefore provide related, but not independent, evidence.\\
\\
On PCaBiop-ISUP, the medical centre remained near-perfectly decodable across the four encoders (confounder balanced accuracy, $0.989$--$1.000$), and every encoder had a negative median \code{CRoMa} (Table~\ref{tab:main-results-panda-isup}). \code{MOOZY} combined the highest biological balanced accuracy (\PandaIsupBestBioBacc) with the least negative median margin (\PandaIsupBestCroma). \code{PRISM}, however, had a slightly lower non-positive fraction than \code{MOOZY} ($65.1\%$ versus $67.5\%$), again distinguishing median robustness from the prevalence of non-positive occurrences. Because PCaBiop-ISUP also differs from binary PCaBiop in sample composition and evaluation design, the shift towards negative margins cannot be attributed to label granularity alone. Moreover, PANDA contributed to \code{MOOZY}'s pretraining corpus (Table~\ref{tab:model-summary}): its relative advantage should therefore be interpreted in light of potential pretraining--evaluation overlap.

\begin{table}[!htbp]
    \centering
    \small
    \setlength{\tabcolsep}{5pt}
    \begin{tabular}{lccccccccc}
        \hline
        \multicolumn{10}{l}{\emph{(a) Cancer detection}} \\
        \hline
        Model & $k^\star$ & bio bacc & conf bacc & \code{RI} & \code{MaRI} & \code{CRoMa} & $F(0)$ & $\mcode{LTM}_{10}$ & support \\
        \hline
        PRISM & 3 & 0.968 & 0.987 & \textbf{0.281} & \textbf{0.195} & \textbf{0.26} & \textbf{0.288} & \textbf{-0.39} & 10.1\% \\
        MOOZY & 9 & \textbf{0.971} & 0.995 & 0.236 & 0.181 & -0.02 & 0.535 & -0.41 & 24.3\% \\
        TITAN & 9 & 0.915 & 1.000 & 0.015 & 0.001 & -0.30 & 0.895 & -0.60 & 49.1\% \\
        Prov-GigaPath & 3 & 0.716 & 0.995 & 0.014 & 0.003 & -0.41 & 0.990 & -0.59 & \textbf{63.9\%} \\
        \hline
    \end{tabular}

    \vspace{5pt}
    \begin{tabular}{lccccccccc}
        \hline
        \multicolumn{10}{l}{\emph{(b) ISUP grading}} \\
        \hline
        Model & $k^\star$ & bio bacc & conf bacc & \code{RI} & \code{MaRI} & \code{CRoMa} & $F(0)$ & $\mcode{LTM}_{10}$ & support \\
        \hline
        MOOZY & 11 & \textbf{0.934} & 0.999 & \textbf{0.077} & \textbf{0.053} & \textbf{-0.09} & 0.675 & \textbf{-0.45} & \textbf{45.3\%} \\
        PRISM & 3 & 0.858 & 0.989 & 0.063 & 0.039 & -0.14 & \textbf{0.651} & -0.51 & 33.4\% \\
        TITAN & 1 & 0.804 & 1.000 & 0.001 & 0.000 & -0.39 & 0.960 & -0.61 & 19.6\% \\
        Prov-GigaPath & 1 & 0.656 & 0.998 & 0.004 & 0.000 & -0.47 & 0.993 & -0.62 & 34.5\% \\
        \hline
    \end{tabular}
    \caption{\textbf{Robustness of slide-level representations on PCaBiop and PCaBiop-ISUP.} The four slide-level encoders are ordered within each panel by \code{CRoMa} ($m{=}5$). Biological and confounder balanced accuracy (bio bacc and conf bacc) are both reported at the model-specific, biologically selected $k^\star$. \code{RI} and \code{MaRI} are pooled at $k^\star$. Support is the fraction of samples contributing to these fixed-$k$ scores. $F(0)$ is the fraction with $\mcode{CRoMa}\leq0$, and $\mcode{LTM}_{10}$ is the mean of the lowest decile. Bold denotes the most favourable value in each score. \textbf{a}, Binary cancer detection on $1{,}000$ slides. \textbf{b}, Six-class ISUP grading over $3{,}000$ slides.}
    \label{tab:main-results-panda}
    \label{tab:main-results-panda-isup}
\end{table}

\end{document}